\pdfoutput=1
\documentclass[10pt,twocolumn,letterpaper]{article}

\usepackage[pagenumbers]{cvpr} %

\usepackage{cuted} %

\usepackage{currfile} %

\usepackage{caption} %
\usepackage{adjustbox}
\usepackage{enumitem}

\renewcommand{\paragraph}[1]{\par\addvspace{.5em}\noindent\textbf{#1}\hspace{.75em}}

\usepackage{xspace}
\newcommand{\method}{WhereBits\xspace}

\usepackage{amsmath,amssymb,amsthm}
\newtheorem{proposition}{Proposition}
\usepackage{float}
\usepackage{balance}

\definecolor{cvprblue}{rgb}{0.21,0.49,0.74}
\usepackage[pagebackref,breaklinks,colorlinks,allcolors=cvprblue]{hyperref}

\title{Bits, Not Boxes: Label-Free Layer Selection for Visual Localization in VLMs}

\author{%
Haisheng Fu$^{1}$\qquad Gexin Huang$^{1}$\qquad Haikai Zhao$^{2}$\qquad Jie Liang$^{2}$\\
Zhenman Fang$^{3}$\qquad Jingning Han$^{4}$\qquad Xiaoxiao Li$^{1}$\\[0.5em]
{\normalsize $^{1}$University of British Columbia\quad $^{2}$Simon Fraser University\quad $^{3}$University of Minnesota--Twin Cities\quad $^{4}$Google}\\[0.3em]
{\normalsize Code: \url{https://github.com/fengyurenpingsheng/BitsNotBoxes}}%
}
\hypersetup{pdftitle={Bits, Not Boxes: Label-Free Layer Selection for Visual Localization in VLMs},pdfauthor={Haisheng Fu, Gexin Huang, Haikai Zhao, Jie Liang, Zhenman Fang, Jingning Han, Xiaoxiao Li}}

\begin{document}
\maketitle
\begin{strip}
\centering
\includegraphics[width=0.91\linewidth]{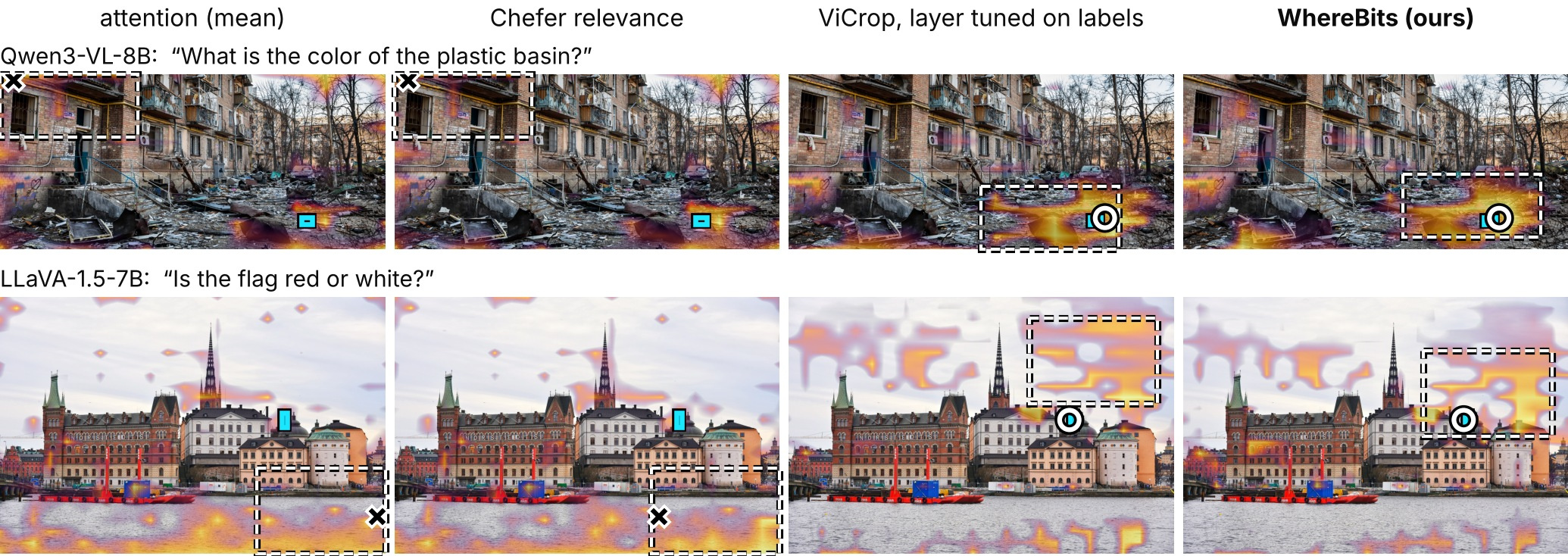}
\captionsetup{hypcap=false}
\captionof{figure}{\textbf{Attention sinks pull zoom-in crops away from the evidence.}
For each model, we show the first V$^*$ question on which \method hits the target (cyan box) while mean attention misses on a sink. Rings mark peaks on the target, crosses mark misses, and dashed boxes show the crops for a one-third window. Over all 191 questions, 2\% of the mean-attention crops of Qwen3-VL-8B contain the target, and its accuracy stays at 47.1\% (47.6\% without a crop). With \method, 54\% contain it and accuracy rises to 64.9\%.}
\captionsetup{hypcap=true}
\label{fig:teaser}
\end{strip}

\begin{abstract}
Attention maps cheaply show where a vision-language model (VLM) looks, but attention sinks make them unreliable for localization. Contrasting the question with a generic prompt suppresses sinks, yet only some layers localize the queried object. Existing methods find these layers with box annotations or fixed rules. We show that attention itself can identify them. A question changes how much a head attends to the image and where this attention goes. The two effects separate exactly, and the second, measured in bits on unlabeled questions, indicates which layers localize. \method turns these where-to-look bits into a layer prior and combines it with per-head contrastive scores. Its three design constants were fixed once on labeled COCO data, and a new model or resolution needs only unlabeled questions. Across eight VLMs, \method matches the pointing accuracy of box-tuned layers on COCO and RefCOCO and ranks target tokens better. This advantage holds on a held-out model family and at higher resolutions. Its COCO pointing accuracy is 0.72, against 0.22 for mean attention. In training-free zoom-in, its crops raise V$^*$ accuracy by 6.0 points under forced-choice evaluation.
\end{abstract}

\section{Introduction}
\label{sec:intro}

Attention maps offer the cheapest view of where a vision-language model (VLM) looks. They need one forward pass and no training, and training-free zoom-in methods already use them to crop the region that a question refers to~\cite{zhang2025vicrop,ge2025carve}. Raw attention, however, is unreliable for localization because of \emph{attention sinks}, tokens that attract attention regardless of the prompt~\cite{xiao2024streamingllm,sun2024massive,kang2025sink} (\cref{fig:teaser}). On held-out COCO questions, the mean attention of Qwen3-VL-8B~\cite{bai2025qwen3vl} peaks at the top-left visual token 89\% of the time. It hits the queried object for only 9.7\% of the questions, less often than a random token (19.3\%). On V$^*$ Bench~\cite{wu2024vstar}, only 2--7\% of its mean-attention crops contain the target, and the crops do not improve accuracy.

A common remedy is to contrast two prompts. Attention that appears under both the question and a generic instruction does not depend on the question and can be divided out~\cite{zhang2025vicrop,ge2025carve}. The contrast removes most of the sink, but it localizes only in some layers, and these layers differ from one model to another (\cref{fig:layers}). A contrastive method therefore has to decide which layers to read.

Existing methods make this decision with annotations or with fixed rules. ViCrop~\cite{zhang2025vicrop} selects one layer per model using box annotations, so each new model needs labeled data. CARVE~\cite{ge2025carve} reads a fixed band of deep layers. LASER~\cite{zhu2026laser} picks a layer for each question from the magnitude of the attention change, and PriorTR~\cite{chen2026priortr} scores tokens at a fixed early layer for token pruning. Such rules are simple to apply, but they are not derived from how well a layer localizes in a given model. We ask instead whether attention itself can indicate which layers localize, without any labels.

Our answer starts from the two ways in which a question can change a head's attention to the image. It can change \emph{whether} the head looks at the image (the total attention on visual tokens) and \emph{where} it looks (how this attention is distributed over them). The divergence between a head's attention under the question and under the generic prompt separates exactly into these two parts (\cref{prop:chain}). Only the second part concerns location, and we find that it carries the signal for layer selection. Averaged over unlabeled questions, where-to-look bits yield layer scores that follow held-out localization accuracy across depth (Spearman $\rho=0.68$--$0.92$), while whether-to-look bits alone lower COCO pointing accuracy by 0.108. Ten unlabeled questions suffice for these scores, so they can take the place of box annotations in layer selection.

\begin{figure*}[t]
\centering
\includegraphics[width=0.93\linewidth]{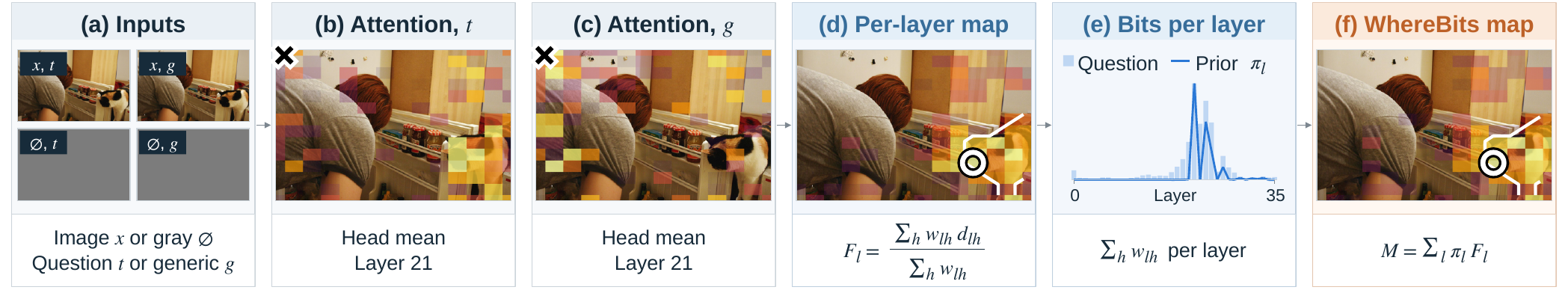}
\caption{\textbf{From attention rows to a question-specific map.}
(a)~The inputs pair the image $x$ or a gray image $\varnothing$ with the question $t$ or a generic prompt $g$.
(b,c)~On a held-out COCO example (supplement), head-mean attention of Qwen3-VL-8B at layer 21 peaks in the top-left corner under both prompts (crosses).
(d)~Weighted per-head log-ratios at this layer peak on the queried cat (circle, target outlined).
(e)~Bits per layer for this question (bars) and the prior (line).
(f)~The final map over all layers. Heatmaps show within-image percentiles.}
\label{fig:method}
\end{figure*}

We turn this observation into \method, a training-free attention map (\cref{fig:method}). \method weights layers by a prior computed from where-to-look bits on unlabeled questions, after subtracting the bits that the same prompts produce on a gray image. Within a layer, each head scores a visual token by the log-ratio of its attention under the two prompts. For a single head, a logit offset shared by both prompts, such as that of a sink, cannot change the token ranking of this score (\cref{prop:cancel}). We call \method \emph{label-free} because, once its three design constants were fixed on labeled COCO data, applying it to a new model or input resolution uses only unlabeled questions. Our contributions are as follows.
\begin{itemize}[leftmargin=1.1em,itemsep=1pt,topsep=2pt,parsep=0pt]
\item We separate the effect of a question on attention into whether-to-look and where-to-look parts, and show that where-to-look bits from unlabeled questions identify the layers whose maps localize.
\item We propose \method, which combines a label-free layer prior with per-head log-ratio scores and needs only two or three gradient-free forward passes.
\item On eight VLMs from five families, \method matches the pointing accuracy of a box-tuned ViCrop layer on COCO and RefCOCO and ranks target tokens better, also in 11 of 14 transfer settings. It ranks first in deletion--insertion faithfulness on COCO, and its crops raise V$^*$ accuracy by 6.0 points under forced-choice evaluation.
\end{itemize}

\section{Related work}
\label{sec:related}

\paragraph{Localization with contrastive attention.}
ViCrop~\cite{zhang2025vicrop} showed that VLM attention locates small visual details once it is divided by the attention under a generic instruction. It crops the image with a single layer of this map, chosen per model with annotated boxes. CARVE~\cite{ge2025carve} contrasts similar prompts, aggregates the ratio over a fixed band of deep layers and over decoding steps, and then masks and crops the image. LASER~\cite{zhu2026laser} and ARA~\cite{chen2026ara} adapt the layers to each question. LASER contrasts the question with a query-free reference and selects the layer whose positive attention differences have the largest norm, and ARA fuses relative-attention maps according to their peaks and entropy. These works establish that contrast makes attention useful for localization. How to choose the layers of a given model without labels remains open, and this is the question that we address. \method estimates layer weights once per model from unlabeled questions. Search-based methods~\cite{wu2024vstar,shen2024zoomeye} and learned zoom policies~\cite{zheng2025deepeyes} study how to acquire higher-resolution evidence and can be combined with any map.

\paragraph{Heads and layers that read the image.}
Bi \etal~\cite{bi2025visualheads} identify visual perception heads, and Kang \etal~\cite{kang2025fewheads} show that a few heads, selected by visual attention mass and spatial entropy, suffice for grounding. Attention knockout traces the flow of visual information across layers~\cite{kaduri2025whats,zhang2025crossmodal}, and labeled probes find that deep attention can localize the evidence even when the answer is wrong~\cite{liu2026seeing}. Song \etal~\cite{song2026fusion} contrast early and final layers, and Liu \etal~\cite{liu2026profiling} profile vision-encoder layers with unlabeled statistics. Token pruning~\cite{chen2024fastv} also scores visual tokens with attention. PriorTR~\cite{chen2026priortr} is closest to our token score. It ranks visual tokens by an attention-weighted log-ratio between the instruction rows and a separator reference at a fixed early layer. MAP~\cite{sun2026map} selects a teacher layer by question contrast. These methods keep informative tokens for efficiency, while \method targets localization. In our ablation, the layer prior accounts for most of the difference from PriorTR's score (\cref{sec:ablation}).

\paragraph{Attention sinks and faithfulness.}
Sinks have been documented in language models~\cite{xiao2024streamingllm,sun2024massive,gu2025sink}, vision transformers~\cite{darcet2024registers} and VLMs~\cite{kang2025sink,luo2026sink,choi2026sinks}. Existing remedies first detect the sink tokens and then redistribute their attention~\cite{kang2025sink} or modify their keys before selecting a crop~\cite{zhang2026borrowed}. \method needs no detection step, because a contribution shared by both prompts cancels in the per-head log-ratio (\cref{prop:cancel}). A map that localizes well need not reflect what the prediction depends on~\cite{jain2019attention,wiegreffe2019attention,jacovi2020faithfulness}. We therefore also measure deletion and insertion~\cite{petsiuk2018rise,rong2022road,song2026heterogeneous}, with a second budget axis from a learned image codec~\cite{minnen2018joint,balle2018hyperprior} that accounts for the uneven coding cost of image regions.

\section{Method}
\label{sec:method}

\method measures how each attention head changes when a generic prompt is replaced by the question (\cref{fig:method}). Per-token log-ratios indicate which visual tokens gain attention (\cref{sec:pmi}). The amount of redistribution, measured in bits, determines how much each head and each layer contributes (\cref{sec:bits,sec:prior}).

\subsection{Four input conditions}
\label{sec:setup}
We read the attention row of the last prompt token, whose representation predicts the first answer token, and call it the \emph{answer-start row}. For layer $l$ and head $h$, let $a_{lh}^{c}(n)$ be the attention of this row to visual token $n$ under input condition $c$. We separate the total visual mass $m_{lh}^c=\sum_{n=1}^{N}a_{lh}^{c}(n)$ from the distribution over the $N$ visual tokens, $P_{lh}^c(n)=a_{lh}^c(n)/m_{lh}^c$. We call $P_{lh}^c$ the \emph{visual routing distribution}. It describes where the head looks within the image, regardless of how much it looks at the image.

An input condition pairs the image $x$ or a uniform gray image $\varnothing$ of the same size with the question $t$ or ViCrop's generic prompt $g$ (\textquotedblleft Write a general description of the image.\textquotedblright)~\cite{zhang2025vicrop}. Such content-free inputs also serve calibration and contrastive decoding~\cite{zhao2021calibrate,leng2024vcd}. This gives four conditions, $xt$, $xg$, $\varnothing t$ and $\varnothing g$. The last one depends on neither the image nor the question, so we cache it per model and input size.

\subsection{Per-head token scores}
\label{sec:pmi}
For each head, we score a visual token by the change in its log routing probability when $g$ is replaced by $t$,
\begin{equation}
d_{lh}(n)=\log_2 P^{xt}_{lh}(n)-\log_2 P^{xg}_{lh}(n).
\label{eq:pmi}
\end{equation}
A positive score means that the question increases the token's share of the head's visual attention, measured in bits. The score is a reference-based analogue of pointwise mutual information~\cite{church1990pmi}, as the supplement shows. We drop the indices $l$ and $h$ when they are clear from context.

\begin{proposition}[Invariance of relative token scores]
\label{prop:cancel}
Suppose that the same additive term $b(n)$ is added to the attention logit of visual token $n$ under both prompts. Then, for a fixed head, the difference $d(n)-d(j)$ between any two visual tokens does not depend on $b$, so the token ranking is unchanged.
\end{proposition}
\begin{proof}
Restricted to the visual tokens, the routing distribution is a softmax of the attention logits $s^c$, so $d(n)=\bigl(s^{xt}(n)-s^{xg}(n)\bigr)/\ln 2+C$ with $C$ constant across tokens. A shared term $b(n)$ cancels in the logit difference and can only change $C$.
\end{proof}

\Cref{prop:cancel} explains why the per-head log-ratio is robust to sinks. In the causal decoders that we study, the image precedes the text, so both prompts see the same visual keys. A key component that both queries read equally then adds the same $b(n)$ under both prompts. We model a sink that attracts attention under either prompt~\cite{xiao2024streamingllm,sun2024massive,kang2025sink} as such a component. The sink then cannot change the token ranking within a head, or in any combination of heads with fixed weights. The guarantee ends there. It does not cover head weights computed from the attention itself, crop rules that use only the positive part of the map, or ratios taken after averaging heads. Such ratios are not invariant when the shared term differs across heads. \Cref{sec:sinks} tests empirically whether the full method suppresses real sinks.

\subsection{Where-to-look bits}
\label{sec:bits}
A prompt change can alter both how much a head attends to the image and how it distributes that attention. To separate the two effects, we merge all non-visual tokens into one outcome. This gives the grouped distribution $\hat a^c=(m^cP^c,\,1-m^c)$.

\begin{proposition}[Chain rule for visual routing~\cite{cover2006elements}]
\label{prop:chain}
The divergence between the grouped attention distributions under the two prompts is
\begin{equation}
\begin{aligned}
\mathrm{KL}(\hat a^{xt}\Vert\hat a^{xg})
={}&\underbrace{\mathrm{KL}_{\mathrm{bin}}(m^{xt}\Vert m^{xg})}_{\text{whether to look}}\\
&+\underbrace{m^{xt}\mathrm{KL}(P^{xt}\Vert P^{xg})}_{\text{where to look}},
\end{aligned}
\end{equation}
where $\mathrm{KL}_{\mathrm{bin}}$ is the KL divergence between Bernoulli distributions and all logarithms are base 2.
\end{proposition}

The first term depends only on the total visual mass. The second term measures how strongly the question redistributes attention within the image, weighted by the mass that the head places on the image. We call it the head's \emph{where-to-look bits}. It is the only term that depends on where the attention goes.

Where-to-look bits are not yet specific to the image content. Even on a uniform image, the two prompts induce different routing, for example because they differ in length. We therefore compute the same bits on the gray image and subtract them. The head weight is
\begin{equation}
w_{lh}=\bigl[B_{lh}^{x}-B_{lh}^{\varnothing}\bigr]_+,
\qquad
B_{lh}^{z}=m_{lh}^{zt}\,\mathrm{KL}(P_{lh}^{zt}\Vert P_{lh}^{zg}),
\label{eq:w}
\end{equation}
where $[v]_+=\max(v,0)$ and $z\in\{x,\varnothing\}$. Because KL is nonlinear, this difference is not itself a divergence. We use it as an empirical weight and ablate it in \cref{sec:ablation}, and the supplement analyzes the gray reference. The map of layer $l$, denoted $F_l(n)$, is the $w_{lh}$-weighted average of $d_{lh}(n)$ over heads, with $F_l=0$ if every head has zero weight.

\subsection{Layer weights from unlabeled questions}
\label{sec:prior}
The same head weights indicate which layers respond to questions by redistributing visual attention. We score each layer by its mean total head weight, $\bar B_l=\mathbb E_{(x,t)}[\sum_h w_{lh}]$, over 385 COCO calibration questions. The layer prior and the final map are
\begin{equation}
\pi_l=\frac{\bar B_l^{\,\gamma}}{\sum_j\bar B_j^{\,\gamma}},
\qquad M(n)=\sum_l\pi_l F_l(n).
\label{eq:map}
\end{equation}
The exponent $\gamma$ concentrates the prior on the layers with the largest scores and still lets several layers contribute. The prior is computed once per model and input size, while the head weights adapt to each image and question. Whether $\bar B_l$ ranks layers by how well they localize is an empirical question, which we answer on held-out data (\cref{sec:loc,sec:transfer}).

We fixed $\gamma=4$, the generic prompt and the gray level once, using the labels of the COCO calibration split and five models. The recipe was then frozen. Calibrating the prior for a new model or input resolution uses only the images and questions of this split and never their localization labels. This is the sense in which \method is label-free.

\subsection{Using the map for zoom-in}
\label{sec:bitswin}
Our primary crop rule selects the fixed-size window with the largest sum of positive map scores. A secondary rule, the \emph{bits window}, scores a window $W$ by how much the question increases the probability of looking inside it. With $\mu=P^{xt}(W)$ and $\nu=P^{xg}(W)$, a head's score is $b(W)=\mathrm{KL}_{\mathrm{bin}}(\mu\Vert\nu)$ if $\mu>\nu$ and zero otherwise. These scores are averaged with the head and layer weights of the map.

\paragraph{Two-pass inference and cost.}
\label{sec:design}
With $\varnothing g$ cached, \method uses three gradient-free forward passes per question ($xt$, $xg$ and $\varnothing t$). The gray pass can be dropped at test time. If the head weights are taken from the visual mass $m^{xt}_{lh}$, inference needs only $xt$ and $xg$, and the gray image is used once per model and resolution to compute the prior. This two-pass variant changes mean pointing and AUROC on COCO and RefCOCO by at most 0.002 (\cref{tab:ablation}). On one L40S GPU, the two-pass map takes 0.15--0.40\,s per question across five models, the same as ViCrop's two passes, and the three-pass map takes 0.23--0.61\,s (CPU timings in the supplement).

\section{Evaluation protocol}
\label{sec:protocol}

\begin{figure*}[t]
\centering
\includegraphics[width=\linewidth]{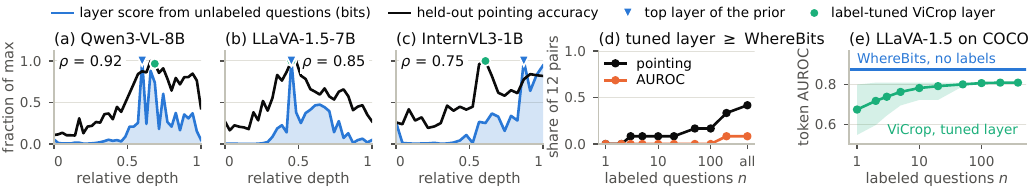}
\caption{\textbf{Unlabeled questions identify the layers whose maps localize.}
(a--c)~On COCO, the layer scores from unlabeled calibration questions and the held-out pointing accuracy of each layer peak at similar depths. Both are normalized by their maximum, and $\rho$ is their Spearman correlation.
(d)~Share of the 12 COCO and RefCOCO pairs in which a ViCrop layer tuned on $n$ labeled questions reaches \method (2{,}000 draws).
(e)~AUROC of the tuned layer for LLaVA-1.5-7B on COCO, with the 10th to 90th percentiles shaded.}
\label{fig:layers}
\end{figure*}

\paragraph{Data.}
Localization uses presence and color questions about annotated objects in COCO val2017~\cite{lin2014coco}, referring expressions from RefCOCO testA/testB~\cite{yu2016refcoco}, and attribute and spatial questions from V$^*$ Bench~\cite{wu2024vstar}. We developed the method on 385 COCO questions from images with odd IDs, the \emph{calibration split}, and we report results on 393 questions from images with even IDs. RefCOCO images are split by ID parity in the same way, with 315 expressions for tuning the label-based baselines and 285 for evaluation. We also use all 191 V$^*$ questions and 500 TextVQA validation questions~\cite{singh2019textvqa}. On every benchmark, the prior of \method comes from the COCO calibration images and questions only. Qwen3-VL-8B, Janus-Pro and the higher-resolution settings were evaluated after the recipe was frozen. The bits-window rule was designed after the primary TextVQA zoom-in results, so its TextVQA numbers are exploratory. Analyses added after the method and the protocol were fixed are marked with $^\S$ in the tables and collected in the supplement.

\paragraph{Localization.}
\emph{Pointing accuracy} is the share of questions for which the peak of the map lies on a target token. \emph{Token AUROC} measures how well the map ranks target tokens above the other tokens, and it is defined for images that contain both classes. A token is a target if the COCO mask or the RefCOCO box covers at least 25\% of its cell. On V$^*$, where targets are small, any overlap with a target box counts. A miss lands on a sink if the peak is among the roughly 5\% of tokens with the highest generic-prompt attention, averaged over heads and layers.

\paragraph{Perturbation faithfulness.}
We follow deletion and insertion~\cite{petsiuk2018rise}. Image cells are replaced by gray or blurred pixels in the order given by a map, and we track the probability of the model's original first answer token. Each map is summarized by its insertion AUC minus its deletion AUC. The same perturbed inputs are scored on two budget axes, the fraction of cells removed or restored and the fraction of the image's coding cost in those cells. The coding cost is estimated with a learned hyperprior codec (CompressAI \texttt{mbt2018-mean}, quality 4)~\cite{minnen2018joint,begaint2020compressai}. As controls, a texture-first order ranks cells by descending coding cost, and a flat-first order reverses it.

\paragraph{Zoom-in.}
Following ViCrop~\cite{zhang2025vicrop}, we search token-aligned windows of about one half or one third of the image side, crop the best window from the original image, and give the model both the image and the crop (the crop alone for LLaVA-1.5). TextVQA answers are generated greedily with at most ten tokens. On V$^*$, generated answers confound crop quality with answer format. Once a crop is added, Qwen3-VL often answers without an option letter. For Qwen3-VL-2B, 44\% of the answers with \method crops lack a letter, and every zoom-in method scores below no zoom. We therefore score V$^*$ by forced choice for all methods and take the option letter with the highest next-token probability in one forward pass. Without a crop, this prediction equals the generated letter on 92--100\% of the questions for four models and on 46\% for SmolVLM-500M, whose answers begin with ``Answer:''. \Cref{sec:robustness} compares the two protocols.

\paragraph{Statistical tests.}
Per-model comparisons use paired two-sided Wilcoxon signed-rank tests with Holm correction within each stated family, and exact McNemar tests for binary pointing give the same corrected counts. Pooled estimates average the paired differences of each question over models, and over window sizes for zoom-in. Their 95\% confidence intervals use a bootstrap over questions, and pooled zoom-in comparisons add a two-sided sign-flip test.

\section{Experiments}
\label{sec:experiments}

\begin{table*}[t]
\begin{minipage}[t]{0.525\textwidth}
\vspace{0pt}
\centering
\footnotesize
\setlength{\tabcolsep}{3.2pt}
\renewcommand{\arraystretch}{1.125}
\begin{adjustbox}{max width=\linewidth}\begin{tabular}{@{}lcccccc@{}}
\toprule
 & \multicolumn{2}{c}{COCO} & \multicolumn{2}{c}{RefCOCO} & \multicolumn{2}{c}{V$^*$} \\
\cmidrule(lr){2-3}\cmidrule(lr){4-5}\cmidrule(lr){6-7}
Method & Pt. & AUROC & Pt. & AUROC & Pt. & AUROC \\
\midrule
\multicolumn{7}{@{}l}{\emph{Uses localization labels for calibration}} \\
ViCrop~\cite{zhang2025vicrop}, tuned layer$^\dagger$ & 0.731 & 0.815 & 0.657 & 0.746 & 0.243 & 0.754 \\
ViCrop, layer tuned for AUROC$^{\dagger\S}$ & 0.690 & 0.822 & 0.670 & 0.750 & 0.206 & 0.725 \\
Three label-tuned heads & 0.617 & 0.814 & 0.609 & 0.785 & -- & -- \\
\addlinespace[2pt]
\multicolumn{7}{@{}l}{\emph{No labeled calibration}} \\
\method (ours) & \textbf{0.723} & \textbf{0.843} & \textbf{0.674} & \textbf{0.755} & 0.210 & 0.789 \\
ViCrop, all layers & 0.664 & 0.788 & 0.593 & 0.710 & 0.204 & 0.739 \\
ViCrop, layer band & 0.664 & 0.793 & 0.626 & 0.718 & 0.196 & 0.739 \\
CARVE~\cite{ge2025carve} (re-impl.) & 0.505 & 0.722 & 0.462 & 0.630 & 0.186 & 0.760 \\
Chefer relevance~\cite{chefer2021generic} & 0.302 & 0.679 & 0.211 & 0.576 & 0.027 & 0.732 \\
Mean attention & 0.221 & 0.670 & 0.184 & 0.573 & 0.002 & 0.715 \\
KL-weighted map & 0.367 & 0.713 & 0.323 & 0.610 & 0.128 & 0.757 \\
Random & 0.196 & 0.500 & 0.234 & 0.502 & 0.013 & 0.494 \\
\addlinespace[2pt]
VAQ~\cite{zhu2026laser}, shared prompts$^\S$ & 0.486 & 0.720 & 0.403 & 0.621 & 0.050 & 0.587 \\
ViCrop, our prior$^\S$ & 0.721 & 0.818 & 0.665 & 0.737 & 0.234 & 0.745 \\
PriorTR~\cite{chen2026priortr}, our prior$^{\ddagger\S}$ & 0.549 & 0.718 & 0.606 & 0.683 & \textbf{0.348} & \textbf{0.799} \\
KL-weighted, sinks masked$^\S$ & 0.548 & 0.732 & 0.407 & 0.631 & 0.253 & 0.771 \\
\bottomrule
\end{tabular}
\end{adjustbox}
\caption{\textbf{Held-out localization.} Pointing accuracy (Pt.) and token AUROC, averaged over six models (five for V$^*$). Bold marks the best label-free result. $^\dagger$Layer tuned with labels (COCO labels for V$^*$). $^\ddagger$Reads the instruction and separator rows. $^\S$Added after the protocol was fixed.}
\label{tab:main}
\end{minipage}\hfill
\begin{minipage}[t]{0.455\textwidth}
\vspace{0pt}
\centering
\includegraphics[width=0.80\linewidth]{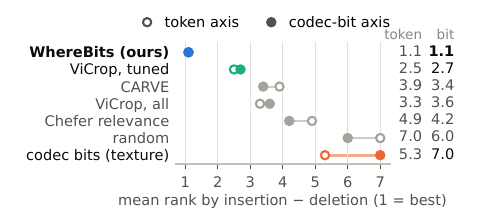}
\captionof{figure}{\textbf{\method ranks first on COCO under both budgets.} Mean rank by insertion minus deletion AUC over ten model--replacement settings per axis (tokens or codec bits).}
\label{fig:faith}
\vspace{6pt}
\footnotesize
\setlength{\tabcolsep}{2.6pt}
\begin{adjustbox}{max width=\linewidth}\begin{tabular}{@{}lcccccc@{}}
\toprule
 & Smol & IVL3 & Qw-2B & Qw-8B & LLaVA & Mean \\
\midrule
No zoom & 35.1 & 41.9 & 49.2 & 47.6 & 47.1 & 44.2 \\
Mean attention & 36.4 & 48.2 & 49.2 & 48.2 & 41.6 & 44.7 \\
ViCrop, tuned$^\dagger$ & 37.2 & 44.8 & 55.8 & 61.5 & 52.9 & 50.4 \\
\method & 37.2 & 41.6 & 54.7 & 63.6 & 53.7 & 50.2 \\
\quad + bits window & 37.4 & 49.0 & 54.2 & 61.8 & 54.5 & \textbf{51.4} \\
\midrule
\method $-$ no zoom & ${+}2.1$ & $-0.3$ & ${+}5.5$ & ${+}16.0^*$ & ${+}6.5^*$ & ${+}6.0^*$ \\
\bottomrule
\end{tabular}
\end{adjustbox}
\captionof{table}{\textbf{V$^*$ zoom-in accuracy per model} (\%, forced choice, mean of the two windows). The columns are SmolVLM-500M, InternVL3-1B, Qwen3-VL-2B and~\mbox{-8B}, and LLaVA-1.5-7B. $^*$The 95\% CI excludes zero. $^\dagger$Tuned with COCO labels.}
\label{tab:vstar-models}
\end{minipage}
\end{table*}

\subsection{Models and baselines}
\label{sec:exp_setup}
We study six VLMs from four families: SmolVLM-256M and~\mbox{-500M}~\cite{marafioti2025smolvlm}, InternVL3-1B~\cite{zhu2025internvl3}, Qwen3-VL-2B and~\mbox{-8B}~\cite{bai2025qwen3vl}, and LLaVA-1.5-7B~\cite{liu2024llava15}. Each model receives one square tile per image (512, 448, 448 and 336\,px by family). V$^*$ is evaluated with all models except SmolVLM-256M, and TextVQA zoom-in with four models.

We compare with 12 label-free maps: mean attention, rollout~\cite{abnar2020rollout}, Chefer relevance~\cite{chefer2021generic}, Grad-CAM~\cite{selvaraju2017gradcam}, TAM~\cite{li2025tam}, ViCrop~\cite{zhang2025vicrop} over all layers or a layer band, CARVE~\cite{ge2025carve}, two variants of the head selection of Kang \etal~\cite{kang2025fewheads}, a DoLa-style layer contrast~\cite{chuang2024dola}, and the KL-weighted map $P^{xt}d$. Two further baselines use the labels of the calibration split, namely a ViCrop layer tuned for pointing and three label-tuned heads. For a controlled comparison, all maps read attention at the answer-start position and share the crop rule. With its original readout, CARVE reaches a pointing accuracy of 0.454, 0.518 and 0.110 on COCO, RefCOCO and V$^*$, compared with 0.505, 0.462 and 0.186 with the shared rows. The original PriorTR score~\cite{chen2026priortr} reaches 0.117 on COCO (supplement).

\subsection{Localization on held-out images}
\label{sec:loc}
On COCO and RefCOCO, \method has the highest pointing accuracy and AUROC of all label-free maps, and it is on par with layers selected using box annotations (\cref{tab:main}).

\paragraph{Comparison with label-free maps.}
\method hits the target 3.3--3.7 times as often as mean attention. The label-free baseline that points best is ViCrop over a layer band. Relative to this baseline, \method raises pointing from 0.664 to 0.723 on COCO and from 0.626 to 0.674 on RefCOCO, and AUROC from 0.793 to 0.843 and from 0.718 to 0.755. Over the 12 baselines, \method wins 252 of 288 model--dataset--metric comparisons after Holm correction and has four losses before correction. On V$^*$, it has higher pointing accuracy and AUROC than the nine baselines evaluated there, with 50 corrected wins and two losses in 90 comparisons.

\paragraph{Comparison with box-tuned layers.}
Without labels, \method matches the pointing accuracy of the label-tuned ViCrop layer. The pooled difference is $-0.008$ on COCO (95\% CI $[-0.023,0.007]$) and $+0.017$ on RefCOCO ($[-0.001,0.035]$). \method also ranks target tokens better. Its AUROC is higher by $0.029$, $0.009$ and $0.035$ on COCO, RefCOCO and V$^*$, and all three pooled intervals exclude zero. Per model and dataset, there are gains in 12 of 17 pairs and no loss after Holm correction, and nine gains and two losses against a layer tuned for AUROC. The exception is pointing on V$^*$, where \method is $0.032$ below the tuned layer ($[-0.054,-0.012]$, \cref{sec:robustness}).

\paragraph{Role of the layer prior.}
The layer prior is the main reason for this parity. Layer scores from unlabeled questions follow held-out pointing across depth, with Spearman $\rho=0.68$--$0.92$ over the 12 COCO and RefCOCO pairs (\cref{fig:layers}a--c). In 11 of these pairs, the prior-weighted map comes within 0.031 pointing of the best single layer selected on the held-out questions. The exception is InternVL3-1B on COCO, with a gap of 0.115. The prior also helps other maps. With our prior instead of its tuned layer, ViCrop's ratio reaches 0.721 pointing on COCO and 0.665 on RefCOCO, against 0.731 and 0.657. The weighted per-head log-ratios then add 0.025 and 0.018 AUROC. More labels do not help the tuned layer much. Even with all calibration labels, it reaches \method in only five of 12 pairs for pointing and in one pair for AUROC (\cref{fig:layers}d).

\paragraph{Per-question layer selection.}
LASER~\cite{zhu2026laser} selects a layer for each question with its Visual Activation by Query (VAQ) rule. Tested alone on our rows and prompts, without LASER's query-free reference and later stages, VAQ points to the target 0.16--0.27 less often than \method (\cref{tab:main}). Combining its layer maps with our prior raises V$^*$ pointing from 0.050 to 0.199.

\subsection{Where localization misses land}
\label{sec:sinks}
\method rarely places its peak on a sink, as expected if sinks act as shared logit terms (\cref{prop:cancel}). On held-out COCO, 90--100\% of the misses of mean attention land on a sink, compared with 71--98\% for Chefer relevance, 59--99\% for the KL-weighted map and only 0--18\% for \method. Under the generic prompt, sink tokens receive 4.5--7.3 bits more attention than the median visual token. Yet their log-ratio scores fall below the median in all six models, while target scores exceed it by 1.4--5.3 bits. Masking the detected sinks also helps and raises mean-attention pointing by 0.13--0.50 on COCO. Even so, \method improves AUROC over the sink-masked baselines in all 12 COCO and RefCOCO pairs after correction. On V$^*$, the sink-masked KL-weighted map points more often than \method but has a lower AUROC (\cref{tab:main}).

\subsection{Faithfulness under image perturbations}
\label{sec:faith}
\method ranks first in deletion--insertion faithfulness on held-out COCO, with a mean rank of 1.1 under both budgets (\cref{fig:faith}). Across the 20 settings, it beats the label-tuned layer in 14 after Holm correction, CARVE in 18, and all-layer ViCrop and Chefer relevance in all 20, with no corrected loss. Texture does not explain this result. The correlation between \method and the codec-bit map is weak ($\rho=-0.01$ to $0.18$), and the texture-first and flat-first controls rank last under the bit budget and the token budget, respectively. Other codec settings change the mean ranks by at most 0.1. The picture differs on TextVQA, where four maps rank ahead of \method (\cref{sec:robustness}).

\subsection{Using the maps to zoom in}
\label{sec:zoom}
\begin{table}[t]
\centering
\footnotesize
\setlength{\tabcolsep}{3pt}
\begin{adjustbox}{max width=\linewidth}\begin{tabular}{@{}lcccc@{}}
\toprule
 & TextVQA & \multicolumn{2}{c}{V$^*$, 448\,px} & V$^*$, high res. \\
\cmidrule(lr){2-2}\cmidrule(lr){3-4}\cmidrule(l){5-5}
Method & Acc. & Acc. & In crop & Acc. \\
\midrule
No zoom & 56.4 & 44.2 & -- & 67.7 \\
\midrule
Mean attention & 63.2 & 44.7 & 17 & 65.5 \\
Chefer relevance~\cite{chefer2021generic} & 63.9 & 46.6 & 22 & 66.6 \\
CARVE~\cite{ge2025carve} & \textbf{64.1} & 48.7 & 39 & 70.2 \\
ViCrop~\cite{zhang2025vicrop}, all layers & 62.3 & 48.1 & 36 & 76.0 \\
ViCrop, tuned layer$^\dagger$ & 61.7 & 50.4 & 43 & 75.3 \\
\addlinespace[2pt]
\textbf{\method}, primary rule & 62.3 & 50.2 & 44 & \textbf{76.2} \\
\textbf{\method}, bits window & 63.4$^\S$ & \textbf{51.4} & 44 & 73.9 \\
\bottomrule
\end{tabular}
\end{adjustbox}
\caption{\textbf{Zoom-in accuracy (\%)}, mean of the two windows. TextVQA uses four models. V$^*$ is scored by forced choice (chance 36.0\%) with five models at 448\,px and with Qwen3-VL-2B and~\mbox{-8B} at higher resolution. ``In crop'' is the share of crops that contain every target box. $^\dagger$Tuned with COCO labels. $^\S$Exploratory.}
\label{tab:zoom}
\end{table}
\paragraph{V$^*$ Bench.}
\method crops improve V$^*$ accuracy and match the crops of the box-tuned layer (\cref{tab:vstar-models,tab:zoom}). Mean attention selects a corner crop for 98--100\% of the questions in four of five models, and its crops do not help ($+0.5$ points over no zoom, 95\% CI $[-1.5,2.6]$). \method lowers the corner rate to 3--14\%. Its crops raise accuracy by 6.0 points over no zoom ($[3.5,8.5]$) and by 5.4 points over mean-attention crops ($[3.2,7.6]$). Without labels, \method is on par with the box-tuned layer ($-0.3$, $[-1.7,1.2]$) and ahead of all-layer ViCrop ($+2.1$, $[0.5,3.7]$). The bits window adds 1.2 points, with an interval that includes zero ($[-0.3,2.7]$). The gain varies across models, from 16.0 points for Qwen3-VL-8B to none for InternVL3-1B (\cref{sec:robustness}), and SmolVLM-500M stays near chance with every map. CARVE's full pipeline, evaluated with generated answers, scores 0.6--2.5 points below \method crops, with intervals that include zero (supplement).

\paragraph{TextVQA.}
On TextVQA, every map helps and the differences between maps are small (\cref{tab:zoom}). The primary rule gains 6.0 points over no zoom (95\% CI $[3.9,8.1]$). This is on par with both ViCrop variants but 1.6 and 1.8 points below Chefer relevance and CARVE (paired $p<0.01$). The exploratory bits window gains 1.7 and 1.1 points over tuned and all-layer ViCrop, and its intervals against CARVE, Chefer relevance and mean attention include zero.

\subsection{Transfer across model families and resolutions}
\label{sec:transfer}
\begin{table*}[t]
\centering
\footnotesize
\setlength{\tabcolsep}{3.1pt}
\begin{adjustbox}{max width=\textwidth}\begin{tabular}{@{}lll@{\hspace{5pt}}cccccc@{\hspace{5pt}}ccccccc@{}}
\toprule
 & & & \multicolumn{6}{c}{COCO (held-out)} & \multicolumn{7}{c}{V$^*$ Bench} \\
\cmidrule(lr){4-9}\cmidrule(l){10-16}
 & & & \multicolumn{2}{c}{\method} & \multicolumn{2}{c}{ViCrop, tuned$^\dagger$} & \multicolumn{2}{c}{ViCrop~\cite{zhang2025vicrop}, all} & \multicolumn{2}{c}{\method} & \multicolumn{2}{c}{ViCrop, tuned$^\dagger$} & \multicolumn{2}{c}{ViCrop, all} & Attn. \\
\cmidrule(lr){4-5}\cmidrule(lr){6-7}\cmidrule(lr){8-9}\cmidrule(lr){10-11}\cmidrule(lr){12-13}\cmidrule(lr){14-15}\cmidrule(l){16-16}
Model & Input & Tokens & Pt. & AUROC & Pt. & AUROC & Pt. & AUROC & Pt. & AUROC & Pt. & AUROC & Pt. & AUROC & Pt. \\
\midrule
Janus-Pro-1B & 384\,px & 576 & \textbf{0.835} & \textbf{0.849} & 0.776$^*$ & 0.800$^*$ & 0.799 & 0.838$^*$ & \textbf{0.277} & \textbf{0.837} & 0.215$^*$ & 0.704$^*$ & 0.267 & 0.749$^*$ & 0.021$^*$ \\
Janus-Pro-7B & 384\,px & 576 & \textbf{0.840} & \textbf{0.855} & 0.819 & 0.835$^*$ & 0.796$^*$ & 0.820$^*$ & 0.257 & \textbf{0.819} & \textbf{0.272} & 0.806$^*$ & 0.267 & 0.774$^*$ & 0.005$^*$ \\
\midrule
Qwen3-VL-2B & 448\,px & 196 & 0.819 & \textbf{0.898} & \textbf{0.827} & 0.890$^*$ & \textbf{0.827} & 0.864$^*$ & 0.387 & \textbf{0.910} & \textbf{0.435} & \textbf{0.910} & 0.351 & 0.893$^*$ & 0.005$^*$ \\
 & 896\,px & 784 & 0.740 & \textbf{0.889} & 0.761 & 0.806$^*$ & \textbf{0.819}$^*$ & 0.868$^*$ & \textbf{0.508} & \textbf{0.948} & 0.225$^*$ & 0.875$^*$ & 0.361$^*$ & 0.931$^*$ & 0.000$^*$ \\
 & native & $\le$1280 & 0.786 & \textbf{0.891} & \textbf{0.789} & 0.885$^*$ & 0.784 & 0.863$^*$ & \textbf{0.524} & \textbf{0.962} & 0.492 & 0.958$^*$ & 0.372$^*$ & 0.945$^*$ & 0.000$^*$ \\
Qwen3-VL-8B & 448\,px & 196 & 0.891 & \textbf{0.913} & 0.885 & 0.897$^*$ & \textbf{0.896} & 0.880$^*$ & 0.372 & 0.900 & \textbf{0.461}$^*$ & \textbf{0.901} & 0.398 & 0.889$^*$ & 0.005$^*$ \\
 & 896\,px & 784 & 0.880 & \textbf{0.908} & \textbf{0.929}$^*$ & 0.892$^*$ & 0.903 & 0.879$^*$ & 0.398 & \textbf{0.931} & \textbf{0.455}$^*$ & 0.925 & 0.429 & 0.930$^*$ & 0.000$^*$ \\
 & native & $\le$1280 & \textbf{0.896} & \textbf{0.912} & \textbf{0.896} & 0.898$^*$ & 0.868 & 0.883$^*$ & 0.450 & 0.950 & 0.482 & \textbf{0.953} & \textbf{0.508}$^*$ & 0.947$^*$ & 0.000$^*$ \\
\bottomrule
\end{tabular}
\end{adjustbox}
\caption{\textbf{Transfer of the frozen method.} Janus-Pro was held out during development. Qwen3-VL runs at 448\,px, at 896\,px and at its native aspect ratio. Bold marks the best contrastive map per metric. $^*$Uncorrected $p<0.05$ against \method. $^\dagger$Re-tuned with COCO labels at each resolution.}
\label{tab:transfer}
\end{table*}
\paragraph{Localization.}
The frozen recipe transfers to a held-out model family and to higher input resolutions. For each new setting, we only re-estimate the prior from unlabeled questions (\cref{tab:transfer}, with RefCOCO in the supplement). On Janus-Pro-1B and~\mbox{-7B}~\cite{chen2025januspro}, a family held out during development, \method improves AUROC over the label-tuned layer in five of six model--dataset pairs. The exception is Janus-Pro-1B on RefCOCO ($0.725$ versus $0.730$). For Qwen3-VL at 896\,px and at native resolution, the baseline layer is re-tuned with labels at each resolution, and \method still has a significantly higher AUROC in six of eight comparisons. Over all 14 transfer settings, Holm correction within each metric leaves 11 AUROC gains and one loss, together with one pointing gain. Two pointing losses for Qwen3-VL-8B at 896\,px are significant only before correction. The gains are largest where the COCO-tuned layer transfers poorly. On V$^*$, AUROC rises from 0.704 to 0.837 for Janus-Pro-1B, and pointing rises from 0.225 to 0.508 for Qwen3-VL-2B at 896\,px. Sinks stay suppressed, with 94--100\% of the misses of mean attention on a sink against 0--3\% for \method. The prior is also stable across resolutions ($\rho=0.98$--$0.99$), and reusing the 448\,px prior changes COCO pointing by at most 0.023.

\paragraph{Zoom-in at higher resolution.}
More input pixels do not remove the need to zoom in. At 896\,px and native resolution, the no-zoom accuracy of Qwen3-VL-2B and~\mbox{-8B} on V$^*$ rises from 48--49\% to 64--72\%, yet mean attention still sends 62--100\% of its crops to a corner. \method crops contain every target box for 57--77\% of the questions. They raise accuracy by 8.5 points over no zoom ($[4.7,12.4]$), by 10.7 points over mean attention and by 6.0 points over CARVE. They are on par with the crops of the re-tuned layer ($+0.9$, $[-0.9,2.6]$, \cref{tab:zoom}). Reusing the 448\,px prior costs 0.3 points ($[-0.2,0.9]$). The bits window is 2.3 points below the primary rule in this setting.

\subsection{What matters in the estimator}
\label{sec:ablation}
\begin{table}[t]
\centering
\footnotesize
\setlength{\tabcolsep}{2.4pt}
\begin{adjustbox}{max width=\linewidth}\begin{tabular}{@{}lccccc@{}}
\toprule
 & \multicolumn{2}{c}{COCO} & \multicolumn{2}{c}{RefCOCO} & \\
\cmidrule(lr){2-3}\cmidrule(lr){4-5}
Variant & $\Delta$Pt. & $\Delta$AUROC & $\Delta$Pt. & $\Delta$AUROC & $-$\,/\,$+$ \\
\midrule
\multicolumn{6}{@{}l}{\emph{Layer choice}} \\
Own map, one layer tuned on labels$^\S$ & ${+}0.019$ & $-0.004$ & ${+}0.004$ & ${+}0.005$ & 3 / 4 \\
Prior from mass $m$ only$^\S$ & $-0.255$ & $-0.150$ & $-0.188$ & $-0.113$ & 19 / 1 \\
Prior without gray correction$^\S$ & $-0.053$ & $-0.045$ & $-0.085$ & $-0.050$ & 8 / 1 \\
No prior (uniform layers) & $-0.068$ & $-0.045$ & $-0.097$ & $-0.045$ & 21 / 0 \\
Whether- instead of where-to-look bits$^\S$ & $-0.108$ & $-0.047$ & $-0.047$ & $-0.026$ & 17 / 0 \\
Total divergence, whether $+$ where$^\S$ & $-0.009$ & $-0.007$ & ${+}0.003$ & $-0.003$ & 7 / 3 \\
\midrule
\multicolumn{6}{@{}l}{\emph{Map and head weights}} \\
Uniform head weights & $-0.020$ & $-0.012$ & $-0.020$ & $-0.009$ & 14 / 1 \\
Head weights from mass $m$ only & $-0.001$ & $-0.002$ & ${+}0.001$ & $-0.001$ & 9 / 2 \\
Log-ratio of head mean (ViCrop~\cite{zhang2025vicrop}) & $-0.012$ & $-0.024$ & $-0.002$ & $-0.015$ & 11 / 0 \\
\bottomrule
\end{tabular}
\end{adjustbox}
\caption{\textbf{Ablation of the estimator.} Each row changes one component and reports the mean change from \method over six models. The last column counts significant losses and gains in 24 model--dataset--metric comparisons ($p<0.05$, uncorrected). $^\S$Added after the protocol was fixed.}
\label{tab:ablation}
\end{table}
The layer prior is the component that matters most (\cref{tab:ablation}). Uniform layer weights or a prior based on visual mass lower pointing by 0.07--0.26 on COCO and RefCOCO. The prior needs the where-to-look term. Using whether-to-look bits in the prior and the head weights costs 0.11 pointing on COCO, in line with \cref{prop:chain}. The total divergence, which contains both terms and is used in the same two places, performs within 0.01 of where-to-look bits. The where-to-look term therefore carries the signal, and adding the whether-to-look term changes little. The gray reference matters for the prior. Removing it there lowers pointing by 0.053 on COCO and 0.085 on RefCOCO, while removing it from the head weights lowers pointing by only 0.010 and 0.001 (supplement). This motivates the two-pass variant (\cref{sec:design}). Per-head log-ratios help ranking, since uniform head weights or a log-ratio taken after head averaging reduce AUROC. Finally, we start from PriorTR's $P\log(P/Q)$ score on our rows and add the prior, the log-ratio and per-head weighting in turn. COCO pointing then rises from $0.109$ to $0.578$, $0.709$ and $0.723$.

\section{Failure modes and limitations}
\label{sec:robustness}

\paragraph{Small targets.}
\method points less often than the box-tuned layer on V$^*$. The deficit is largest for the smallest third of the targets, which may cover a single visual token. Reading PriorTR's instruction rows instead of the answer-start row raises V$^*$ pointing by 0.15 (95\% CI $[0.12,0.18]$) but lowers COCO pointing by 0.05, so no single readout serves both target scales (supplement).

\paragraph{A prior that favors the wrong depth.}
InternVL3-1B localizes in two depth ranges, and the prior favors the later one. This range reaches 0.71--0.72 pointing on RefCOCO but only 0.45--0.49 on COCO, where layer 14 reaches 0.58. On V$^*$, \method crops leave the accuracy of this model unchanged ($-0.3$ points). A strong response to questions thus need not localize on every dataset.

\paragraph{Localization and answer faithfulness.}
A map that localizes the queried object need not rank the pixels that the answer depends on. On TextVQA (200 questions per model, three models), CARVE, mean attention, Chefer relevance and all-layer ViCrop rank ahead of \method under deletion and insertion. CARVE beats it in eight of 12 settings after Holm correction. These maps keep the attention mass in their scores, but the cause of the gap remains open (supplement). Zoom-in is also less consistent on TextVQA, where \method matches ViCrop but trails Chefer relevance and CARVE (\cref{sec:zoom}).

\paragraph{Answer format in zoom-in evaluation.}
The V$^*$ gains reported above hold under forced choice. With generated answers, many responses contain no option letter. The primary rule then gains 2.6 points over no zoom at 448\,px ($[-0.2,5.3]$) and 2.7 points at higher resolution ($[-1.4,6.8]$), and both intervals include zero. When answers without a letter receive chance credit, the gains are 6.7 and 6.6 points (supplement).

\paragraph{Calibration questions.}
Ten unlabeled COCO questions suffice for the prior, but their distribution matters. Calibrating on V$^*$ questions reduces COCO pointing from 0.718 to 0.455 for SmolVLM-500M and from 0.463 to 0.267 for InternVL3-1B. V$^*$ differs from COCO in both target size and question format.

\paragraph{Reference prompt and scope.}
Gray levels of 64 or 192 instead of 124 change COCO pointing by less than 0.011, but the reference prompt matters. A short-caption prompt, under which the model already attends to the object, lowers pointing by 0.018--0.191 in three models and yields fewer corrected bits. Across ten prompt--model variants, the bits track pointing ($\rho=0.76$), so they offer a label-free check of the reference. All models in our study are causal decoders with the image before the text, and the invariance argument of \cref{sec:pmi} relies on this layout.

\section{Conclusion}
\label{sec:conclusion}

We asked whether attention itself can indicate which layers of a VLM localize, and we found that it can. The way a question redistributes a head's attention over the image, measured in bits on a few unlabeled questions, identifies the localizing layers. \method builds a label-free layer prior on this finding and combines it with per-head contrastive scores. Without localization labels for a new model, it matches box-tuned layers in pointing accuracy on COCO and RefCOCO, ranks target tokens better, and improves training-free cropping on V$^*$ under forced-choice evaluation. Its main limitations are small targets, the reference prompt and the distribution of the calibration questions.

\section*{Acknowledgments}
Claude \mbox{(Anthropic)} and Codex \mbox{(OpenAI)} assisted with the literature search, the development of the method and the experiments, code, analyses, and writing. The authors directed the research, verified all results, and take full responsibility for the paper.

\balance
{\small
\bibliographystyle{ieeenat_fullname}
\bibliography{main}
}

\clearpage
\nobalance
\maketitlesupplementary
\setcounter{section}{0}\renewcommand{\thesection}{\Alph{section}}\renewcommand{\theHsection}{supp.\Alph{section}}
\setcounter{table}{0}\renewcommand{\thetable}{S\arabic{table}}\renewcommand{\theHtable}{supp.\arabic{table}}
\setcounter{figure}{0}\renewcommand{\thefigure}{S\arabic{figure}}\renewcommand{\theHfigure}{supp.\arabic{figure}}
\raggedbottom
\renewcommand{\floatpagefraction}{0.67}
\renewcommand{\dblfloatpagefraction}{0.80}
\setcounter{topnumber}{4}
\setcounter{dbltopnumber}{4}
\setcounter{bottomnumber}{3}
\setcounter{totalnumber}{8}
\setlength{\textfloatsep}{11pt plus 3pt minus 3pt}
\setlength{\dbltextfloatsep}{11pt plus 3pt minus 3pt}
\noindent Sections, figures, and tables with a letter or an S prefix are in this supplement. References use the numbering of the main paper.

\section{Proofs and derivations}
\label{sec:s_proofs}

Throughout, $a^c(n)$ is the attention of the answer-start query of one head to visual key $n$ under condition $c$, $m^c=\sum_n a^c(n)$, $P^c=a^c/m^c$, and logarithms are base 2 unless written $\ln$.

\paragraph{Proposition~\ref{prop:cancel} (shared logit components).}
In a decoder with causal attention, the hidden state at position $j$ depends only on the tokens up to $j$. The two inputs share the same system prefix, image, and visual positions. Induction over layers gives identical keys and values at each visual position, including after the key rotation. This argument also applies to multimodal rotary embeddings when the visual positions come from the shared image grid, as in Qwen3-VL. The two prompts can have different lengths, so we absorb the rotation of each answer-start position into its own query $\tilde q^c$.

Let $Z^c$ denote the softmax normalizer over the full row and $Z^c_{\mathcal V}=\sum_{n\in\mathcal V}e^{s^c(n)}$ the corresponding sum over visual keys. Dividing $a^c(n)=e^{s^c(n)}/Z^c$ by $m^c=Z^c_{\mathcal V}/Z^c$ gives $P^c(n)=e^{s^c(n)}/Z^c_{\mathcal V}$. Hence
\begin{align*}
d(n)&=\frac{s^{xt}(n)-s^{xg}(n)}{\ln2}-\log_2\frac{Z^{xt}_{\mathcal V}}{Z^{xg}_{\mathcal V}}\\
&=\frac{\langle\tilde q^{xt}-\tilde q^{xg},\tilde k_n\rangle}{\sqrt{d_k}\ln2}+C.
\end{align*}
If a key component $\kappa_n$ contributes the same logit offset $b(n)$ to both rows, that offset disappears from the first term. It can still change $C=-\log_2(Z^{xt}_{\mathcal V}/Z^{xg}_{\mathcal V})$ uniformly over tokens. Thus $d(n)-d(j)$ is invariant for every pair of visual tokens $n,j$. The routing $P^{xt}(n)$ retains the token-dependent factor $e^{b(n)}$ up to normalization, so $P^{xt}(n)d(n)$ still depends on $b(n)$. Rollout and gradient-weighted relevance also use the attention probabilities, whose response to $b(n)$ depends on the rest of the row and model.

At fixed head and layer weights, the sum of the per-head constants is constant across the final map, preserving its ranking. Recomputing weights can change the ranking because both visual mass and KL can change. The crop score sums the positive part of the map and can respond to a constant shift. A sink that the question and reference read differently contributes that difference to the log-ratio~\cite{xiao2024streamingllm,sun2024massive,kang2025sink}. \hfill$\square$

\paragraph{Relationship to conditional PMI (\cref{sec:pmi}).}
For a distribution over questions about $x$, let $\bar P_x$ be the marginal routing. Then $d(n)=\mathrm{PMI}(t;n\mid x)+\log_2(\bar P_x(n)/P^{xg}(n))$. The correction from using the generic prompt instead of the marginal can vary across tokens, and is shared across questions about $x$.

\paragraph{Proposition~\ref{prop:chain} (chain rule).}
For $\hat a=(mP,1-m)$,
$\mathrm{KL}(\hat a^{xt}\Vert\hat a^{xg})=\sum_n m^{xt}P^{xt}(n)\log\frac{m^{xt}P^{xt}(n)}{m^{xg}P^{xg}(n)}+(1-m^{xt})\log\frac{1-m^{xt}}{1-m^{xg}}$; splitting the first logarithm gives $m^{xt}\log\frac{m^{xt}}{m^{xg}}+m^{xt}\mathrm{KL}(P^{xt}\Vert P^{xg})$, and the two binary terms form $\mathrm{KL}_{\mathrm{bin}}(m^{xt}\Vert m^{xg})$~\cite{cover2006elements}. \hfill$\square$

\paragraph{Gray-image reference (\cref{sec:bits}).}
\begin{proposition}[Gray-image reference]
\label{prop:gray}
Suppose that the logit shift from $g$ to $t$ on image $x$ is $u_x(n)+r(n)$, where $r(n)$ is shared by the real and gray inputs, and that $u_\varnothing(n)$ is constant across visual tokens. Then the log-ratio on the real image minus the log-ratio on the gray image equals $u_x(n)/\ln2$ up to a token-independent constant.
\end{proposition}
\noindent\emph{Proof.} Under the assumed decomposition, the same position- and prompt-dependent term $r(n)$ appears in the real- and gray-image logit differences. If the gray-image content term is constant across visual positions, it can be absorbed into the gray row's additive constant. Then $d(n)-d^{\varnothing}(n)=u_x(n)/\ln2+\mathrm{const}$, and $\mathrm{KL}(P^{\varnothing t}\Vert P^{\varnothing g})=\sum_n P^{\varnothing t}(n)d^{\varnothing}(n)$ measures the routing change on the gray input. Uniform pixels need not imply identical visual embeddings. The implemented discount is a heuristic difference of mass-weighted KL terms, which are nonlinear in the underlying logit changes. \hfill$\square$

\noindent In practice, subtracting the gray-image log-ratio from each token score lowers pointing (\cref{tab:s_ablation_more}), so \method uses the gray image only in the head weights and the prior.

\paragraph{Bits bound the output change (\cref{sec:bits}).}
\begin{proposition}[Bits bound the change of the head output]
\label{prop:pinsker}
Let $o=\sum_n a(n)\,v_n$ be the visual part of the head output, with $\Vert v_n\Vert\le V$.
At fixed visual mass, the question's re-routing changes it by $m^{xt}\Vert\sum_n(P^{xt}(n)-P^{xg}(n))v_n\Vert\le m^{xt}V\sqrt{2\ln2\cdot\mathrm{KL}(P^{xt}\Vert P^{xg})}$.
\end{proposition}
\noindent\emph{Proof.} Values are shared (see above), so at fixed mass $m^{xt}$ the re-routing changes the visual part of the output by $m^{xt}\sum_n(P^{xt}(n)-P^{xg}(n))v_n$, whose norm is at most $m^{xt}V\Vert P^{xt}-P^{xg}\Vert_1\le m^{xt}V\sqrt{2\,\mathrm{KL}_{\mathrm{nats}}}=m^{xt}V\sqrt{2\ln2\cdot\mathrm{KL}}$ by Pinsker's inequality. The bound grows as $m\sqrt{\mathrm{KL}}$ and ignores the output projection. \hfill$\square$

\paragraph{Layer prior (\cref{sec:prior}).}
Fix an image $x$, let $T$ be a random question and $K\mid T{=}t\sim P^{xt}$. With $\bar P_x=\mathbb E_TP^{xT}$ the marginal of $K$,
$\mathbb E_T\mathrm{KL}(P^{xT}\Vert P^{xg})=\mathbb E_T\sum_kP^{xT}(k)\big[\log\frac{P^{xT}(k)}{\bar P_x(k)}+\log\frac{\bar P_x(k)}{P^{xg}(k)}\big]=I(T;K\mid x)+\mathrm{KL}(\bar P_x\Vert P^{xg})$.
The second term measures how the marginal routing differs from the generic-prompt routing. The same decomposition holds with mass weights under the tilted question distribution $\tilde p(t)\propto p(t)m_t$, giving $\mathbb E[m\,\mathrm{KL}]=\mathbb E[m]\big(I_{\tilde p}(T;K\mid x)+\mathrm{KL}(\bar P_{\tilde p}\Vert P^{xg})\big)$. The prior uses the corrected bits $w_{lh}$ as a surrogate layer score. \Cref{tab:ablation,tab:s_layer_choice} compare this score with uncorrected and mass-only alternatives.

\paragraph{Crop rule (\cref{sec:bitswin}).}
Let $P=P^{xt}$, $Q=P^{xg}$, $\mu=P(W)$, and $\nu=Q(W)$. Grouping visual keys into $W$ and its complement $W^c$ gives
\begin{align*}
\mathrm{KL}(P\Vert Q)
={}&\mathrm{KL}_{\mathrm{bin}}(\mu\Vert\nu)\\
 &+\mu\,\mathrm{KL}(P(\cdot\mid W)\Vert Q(\cdot\mid W))\\
 &+(1-\mu)\,\mathrm{KL}(P(\cdot\mid W^c)\Vert Q(\cdot\mid W^c)).
\end{align*}
The first term measures the change in the probability of routing inside the window. The bits-window rule keeps this term only when $\mu>\nu$.

\section{Implementation details}
\label{sec:s_impl}

\paragraph{Inputs and prompts.}
In the main setting, we resize each image to one square tile at the model's default vision input size (\cref{sec:exp_setup}). This gives a fixed $G_h\times G_w$ visual-token grid. COCO val2017 contributes 500 questions about the presence of an annotated object (``Is there a frisbee in the image? Answer yes or no.'') and 278 about its color (``What color is the couch? Answer with one word.''). All queried objects are present. RefCOCO uses ``Where is the \{expression\} in the image? Answer briefly.'' For TextVQA, we append ``Answer the question using a single word or phrase.'' to a sample of 500 questions drawn from the first 1{,}600 validation questions (seed 0). All four models have zoom-in results for all 500 questions. The LLaVA-1.5 questions were processed in two batches with the same code; two questions run in both batches gave identical maps, answers, and crops. V$^*$ uses the benchmark's questions, options, and instruction.

The reference prompt is ``Write a general description of the image.'' The gray image is uniform RGB $(124,124,124)$, which stays nonzero after normalization (SmolVLM treats all-zero tiles as padding). A visual token is a localization target if the COCO mask or RefCOCO box covers at least 25\% of its cell. For V$^*$ we use any positive overlap up to numerical tolerance. Output-targeted methods and faithfulness use the model's first greedy answer token under $xt$, including when the answer is wrong. The role of this token on RefCOCO is analyzed in \cref{tab:s_target,tab:s_chefer_targets}.

\paragraph{Attention rows.}
For each of the four conditions we store one answer-start row over the $N$ visual keys per head, along with the visual mass computed before quantization. The rows use 8-bit codes of $-\log_2a$ in increments of $1/8$ bit, with floor $2^{-31.875}$. Above the floor, rounding changes each log-probability by at most $1/16$ bit. \method reconstructs routing by dividing dequantized attention by the stored visual mass, clamped below at $2^{-40}$. If no entries are floored, the reconstructed row sum lies in $[2^{-1/16},2^{1/16}]$. Flooring can inflate this sum in very low-mass heads. Renormalizing each head preserves its within-head log-ratio ranks but can change the aggregation weights. Across the 12 COCO and RefCOCO settings, recomputing \method with renormalization changes pointing by at most 0.008 and AUROC by at most 0.0001. Bits-window scoring uses normalized rows.

Row-based maps are computed offline from the same stored rows. Rollout, Chefer relevance, and Grad-CAM are computed in the original model run from full attention matrices or their gradients. The original readouts of PriorTR and CARVE use additional rows recorded in a separate run (\cref{tab:s_own}).

\paragraph{\method.}
We compute $d$ and $w$ by \cref{eq:pmi,eq:w} and set the weighted average $F_l$ to zero for layers with $\sum_hw_{lh}=0$. For each model and input setting, we estimate $\pi_l\propto\bar B_l^4$ from the 385 COCO questions with odd image IDs and reuse it across datasets; if every layer score were zero, the prior would be uniform. The per-question variant retains the three layers with the largest $\sum_hw_{lh}$ for that question and weights them in proportion to this sum.

\paragraph{Cost.}
The $\varnothing g$ row can be cached per model and input size. We use three forward passes ($xt$, $xg$, $\varnothing t$) per calibration or test question; using visual mass alone for head weights reduces test cost to two passes (\cref{tab:ablation}). \Cref{tab:s_timing} times each map end to end on one NVIDIA L40S GPU (median over 30 held-out COCO questions) and on 16 CPU threads of a shared cluster node (median over 20), after three warm-up questions. Each time includes the attention-returning forward passes that the map needs (with $\varnothing g$ cached) and the map arithmetic. The question pass is timed without its input preprocessing, which answering needs anyway, whereas the generic and gray passes include theirs; this matters only for SmolVLM on the GPU, whose generic pass takes 0.20\,s against 0.05\,s for the question pass. Chefer relevance is timed as one forward and one backward pass for the first answer token, which a separate, untimed greedy pass provides. The two-pass map costs the same as ViCrop's head-mean ratio, which needs the same two passes; the third pass adds about half (about 80\% for SmolVLM on the GPU). Chefer relevance costs no more than the two-pass map except for LLaVA-1.5-7B on CPU, but its backward pass also needs the activations of the whole forward pass. Calibration takes three passes per unlabeled question, 0.2--0.6\,s on the GPU, so the 385 calibration questions take at most four minutes and the ten that already estimate the prior to within 0.011 pointing take seconds. LLaVA-1.5-7B was timed on CPU only.
\begin{table}[!tbp]
\centering
\footnotesize
\setlength{\tabcolsep}{2.2pt}
\begin{adjustbox}{max width=\linewidth}\begin{tabular}{@{}lccccccc@{}}
\toprule
 & & & & \multicolumn{3}{c}{\method} & Calib. \\
\cmidrule(lr){5-7}
Model & Attention & ViCrop & Chefer & 2-pass & 3-pass & map (ms) & per q. \\
\midrule
\multicolumn{8}{@{}l}{\emph{CPU, 16 threads (s)}} \\
SmolVLM-256M & 1.58 & 3.65 & 1.95 & 3.65 & 5.72 & 2.3 & 5.71 \\
SmolVLM-500M & 1.76 & 4.00 & 2.32 & 4.00 & 6.25 & 3.1 & 6.25 \\
InternVL3-1B & 2.62 & 5.11 & 3.71 & 5.12 & 7.74 & 4.0 & 7.73 \\
Qwen3-VL-2B & 3.17 & 6.33 & 5.63 & 6.33 & 9.51 & 4.5 & 9.51 \\
Qwen3-VL-8B & 9.44 & 17.94 & 17.26 & 17.94 & 26.76 & 5.5 & 26.75 \\
LLaVA-1.5-7B & 5.66 & 11.20 & 13.45 & 11.20 & 16.87 & 3.1 & 16.87 \\
\midrule
\multicolumn{8}{@{}l}{\emph{GPU, one L40S (ms)}} \\
SmolVLM-256M & 50 & 248 & 127 & 248 & 446 & 1.1 & 444 \\
SmolVLM-500M & 53 & 253 & 134 & 253 & 454 & 1.5 & 452 \\
InternVL3-1B & 71 & 149 & 142 & 150 & 229 & 2.3 & 226 \\
Qwen3-VL-2B & 78 & 164 & 165 & 165 & 253 & 2.6 & 250 \\
Qwen3-VL-8B & 194 & 400 & 356 & 400 & 609 & 3.6 & 605 \\
\bottomrule
\end{tabular}\end{adjustbox}
\caption{\textbf{Time per map} (median over held-out COCO questions: 20 on CPU, 30 on the GPU). Each entry includes the forward passes that the map needs and its arithmetic; map (ms): the three-pass \method arithmetic alone. Calib.\ per q.: three passes per unlabeled question, run once per model and input size.}
\label{tab:s_timing}
\end{table}

\paragraph{Design space.}
\Cref{tab:design} places the attention-based maps of the paper in one design space.
\begin{table}[!tbp]
\centering
\footnotesize
\setlength{\tabcolsep}{2.2pt}
\begin{adjustbox}{max width=\linewidth}
\begin{tabular}{@{}lllll@{}}
\toprule
Map & Contrast & Ref. & Heads & Layers \\
\midrule
Attention (mean) & -- & -- & mean & mean \\
Chefer~\cite{chefer2021generic} & -- & -- & gradient & rollout \\
Kang \etal~\cite{kang2025fewheads} & -- & -- & 3 (mass, entropy) & via heads \\
KL-weighted map & $P^{xt}d$ & $g$ & sum & sum \\
\midrule
ViCrop~\cite{zhang2025vicrop} & $\bar a^{xt}/\bar a^{xg}$ & $g$ & mean, then ratio & 1 (labels) or all \\
CARVE~\cite{ge2025carve} & $\check a^{xt}/(\check a^{xg}{+}\epsilon)$ & $g$ & mean, then ratio & deep band \\
PriorTR~\cite{chen2026priortr} & $\bar P\log(\bar P/\bar Q)$ & separator & mean, then score & layer 2 \\
\midrule
\textbf{\method} & $d=\log_2(P^{xt}/P^{xg})$ & $g$ & log-ratio, bits & bits prior \\
\bottomrule
\end{tabular}
\end{adjustbox}
\caption{\textbf{Attention maps by score and aggregation.}
$\bar a$ denotes head-mean attention and $\check a$ its max-normalized version. The per-head log-ratio preserves token ranks under shared logit offsets (\cref{prop:cancel}). ViCrop and CARVE take ratios after averaging heads. ViCrop's single-layer variant uses labeled calibration, and our shared-row CARVE implementation uses relative depths 0.70--0.90.}
\label{tab:design}
\end{table}

\paragraph{Baselines.}
The shared-row comparison reads attention at the answer-start position. Mean attention averages $a^{xt}$ over heads and layers. The KL-weighted map sums $\sum_{l,h}[P^{xt}_{lh}d_{lh}]_+$. Our layer-contrast baseline, inspired by DoLa~\cite{chuang2024dola}, compares the head-mean routing at 87.5\% depth with the shallower candidate at greatest Hellinger distance, using $|P_k-P_j|$.

\emph{ViCrop}~\cite{zhang2025vicrop} divides head-mean attention under $t$ by head-mean attention under $g$ within each layer. The tuned variant uses the layer that maximizes pointing on the labeled odd split. The other variants average the maps over all layers or over relative depths 0.5--0.9.

\emph{CARVE}~\cite{ge2025carve} in the shared-row comparison divides max-normalized head-mean attention under $t$ by its counterpart under $g$ plus $\lambda=0.05$, then averages over relative depths 0.70--0.90. The published method aggregates the ratio over decoding steps with increasing weights, uses layers 20--25, and crops and masks selected regions before answering again. The main zoom-in comparison applies the same window rule to every map. \Cref{tab:s_own,tab:s_carve_zoom} report the additional decoding-step readout and pipeline experiments.

\emph{Kang \etal}~\cite{kang2025fewheads} read the last input token of three heads. We use the last input token of our answer prompt, while their input is the bare expression. We exclude the first two layers, then keep heads whose visual mass exceeds the maximum-curvature point of the sorted mass curve. For each calibration question we identify the ten heads with lowest spatial entropy, computed after binarizing each map at its mean and forming 8-connected components. We select the three heads that appear most often across the odd split. Their maps are Gaussian-smoothed ($\sigma=1$ token) and summed. Their original method selects heads on 1,000 RefCOCO training examples and predicts boxes. We also evaluate a variant restricting heads to depth $\ge0.25$, and a supervised control that chooses the three heads with highest individual pointing on the odd split.

\emph{PriorTR}~\cite{chen2026priortr} scores $P\log_2((P+\epsilon)/(Q+\epsilon))$ with $\epsilon=10^{-6}$ and visual-token-normalized $P,Q$. In the separate readout run, we form $P$ by averaging heads and instruction rows through the answer-start token, and $Q$ by averaging heads at the separator just after the image (\cref{tab:s_own}). In the shared-row comparison, we use head-mean routing under $t$ and $g$ at the answer-start position (\cref{tab:s_decomp}). We apply the score at layer 2, as in PriorTR's token-pruning setup, or aggregate it across layers with our prior.

\emph{Rollout}~\cite{abnar2020rollout} multiplies head-averaged attention matrices with the identity added and rows renormalized. \emph{Chefer relevance}~\cite{chefer2021generic} differentiates the first greedy answer token's log-probability and updates $R\leftarrow R+\mathbb{E}_h[(\nabla A\odot A)^+]R$ from $R=I$. \emph{Grad-CAM}~\cite{selvaraju2017gradcam} weights last-layer attention by its gradient for the same output. We read the answer-start row of each resulting map over the visual keys. \emph{TAM}~\cite{li2025tam} uses its base map, reading the first answer token's logit at each visual position in the last residual stream.

\paragraph{Faithfulness.}
For deletion or insertion, we remove or retain tokens in map order at token-count fractions $\{0,0.05,0.1,0.2,0.3,0.5,0.7,1\}$. Removed cells are painted uniform RGB $(128,128,128)$ or replaced by a Gaussian-blurred copy (PIL radius $\lfloor\min(W,H)/16\rfloor$ pixels). The 200-question COCO runs for LLaVA-1.5 and Qwen3-VL-8B sample without replacement with seed 0. TextVQA uses the first 200 of the randomized question list. We track the target token's probability and integrate by the trapezoidal rule, normalizing by the axis span. On the codec-bit axis, each cell's contribution is its share of the original image's estimated latent bits~\cite{minnen2018joint} (CompressAI \texttt{mbt2018-mean}, quality 4~\cite{begaint2020compressai}). We project the main latent's negative log-likelihoods to image space and sum them within each token cell. The budget excludes the hyperprior side information, which has no assigned image cell. The two axes account for the same perturbed images by cell count or estimated bit budget. Deleting a prefix of one order keeps the same tokens as inserting the complementary prefix of the reverse order. The flat-first analysis uses this identity on the symmetric subset $\{0,0.3,0.5,0.7,1\}$ of the measured fractions, where reversing the order changes the sign of insertion minus deletion AUC.

\paragraph{Zoom-in.}
We evaluate every token-aligned window of $\mathrm{round}(fG_h)\times\mathrm{round}(fG_w)$ tokens for $f\in\{1/2,1/3\}$, then map the selected window to pixels of the original image. Qwen3-VL, InternVL3, and SmolVLM receive the original image and the crop as two images. LLaVA-1.5 receives the crop alone. On $8\times8$, $16\times16$, $14\times14$, and $24\times24$ grids, rounding makes the nominal $1/3$ window 0.375, 0.3125, 0.357, and 0.333 of the side, respectively. On TextVQA, answers are generated greedily for up to 10 tokens. On V$^*$, the primary protocol is forced choice. For every condition (no zoom, and each map, window and model) we run one forward pass on the inputs used for generation, with the benchmark's question, options and instruction, and score each option letter by its probability as the first answer token, summed over the tokenizer's variants of the letter (with and without a leading space). The prediction is the most probable letter; nothing is generated or parsed. For comparison, we also report generated answers on the same crops: answers are generated greedily for up to 10 tokens, the first stand-alone option letter is parsed, and an answer without a letter counts as incorrect or, in a second scoring, at chance ($1/k$ for $k$ options). No-zoom answers are generated with up to four tokens; with ten tokens, every model's no-zoom accuracy is the same. The benchmark has 107 four-option and 84 two-option questions, giving 36.0\% chance accuracy.

\paragraph{Statistics.}
Per-model comparisons use two-sided Wilcoxon signed-rank tests on paired question scores and 2{,}000-draw bootstrap intervals. For binary pointing, exact McNemar tests on discordant questions give the same Holm-corrected counts in the tuned-layer comparison (\cref{tab:s_counts}). The main pooled comparisons cluster by question ID and average the paired differences across available models and, for zoom-in, the two windows. We use a percentile bootstrap and, for zoom-in, a two-sided sign-flip test with 20{,}000 draws each. Holm correction applies within each stated family, for example the 17 model--dataset pairs for a localization metric or 20 faithfulness settings for a baseline comparison.

The question-clustered TextVQA analysis was chosen after the per-setting results were known. For V$^*$ with generated answers, clustering by model--question pairs gives a narrower interval for the primary rule versus no zoom, which excludes zero; with the question clusters that we report, the interval includes zero (\cref{tab:s_zoom_vstar_q}). The forced-choice analysis was specified with question clusters before it was run; its intervals, pooled or per model, use the question bootstrap with 10{,}000 draws, and its per-setting marks use exact McNemar tests.

\paragraph{Qualitative examples.}
For each model in \cref{fig:teaser}, we show the first V$^*$ question by ID on which \method peaks on a target box and mean attention peaks off target at a token in the top 5\% by question-conditioned mean attention. We skip images already shown for another model. The peak criteria yield 70 eligible questions for Qwen3-VL-8B and 40 for LLaVA-1.5-7B. For \cref{fig:method}, we use the first held-out COCO question by ID for Qwen3-VL-8B whose target covers 3--25\% of tokens, whose \method peak hits the target, and whose KL-weighted peak misses at a token in the top 5\% by question-conditioned mean attention. \Cref{fig:s_vstar_gallery,fig:s_coco_gallery} show random examples.

\section{Additional results}
\label{sec:s_results}

\paragraph{Per-model localization.}
\Cref{tab:s_loc_coco,tab:s_loc_refcoco,tab:s_vstar_loc} report each model separately. On COCO and RefCOCO, \method outperforms label-free baselines in 265 of 288 paired comparisons at $p<0.05$ (252 after Holm correction), with four losses before correction. Those losses are in AUROC against ViCrop's band on InternVL3-1B COCO and against variants of Kang \etal on LLaVA-1.5-7B. The latter baseline performs particularly well on RefCOCO with LLaVA-1.5-7B, the setting used by Kang \etal. It gives 0.660 pointing and 0.774 AUROC, compared with 0.649 and 0.754 for \method. On COCO, the selection still includes a head from layers 2--5 in five of six models after excluding the first two layers. Pointing spans 2--36\% across models.

\begin{table*}[!tbp]
\centering
\footnotesize
\setlength{\tabcolsep}{3pt}
\begin{adjustbox}{max width=\textwidth}\begin{tabular}{@{}lcccccccccccc@{}}
\toprule
 & \multicolumn{2}{c}{SmolVLM-256M} & \multicolumn{2}{c}{SmolVLM-500M} & \multicolumn{2}{c}{InternVL3-1B} & \multicolumn{2}{c}{Qwen3-VL-2B} & \multicolumn{2}{c}{Qwen3-VL-8B} & \multicolumn{2}{c}{LLaVA-1.5-7B} \\
\cmidrule(lr){2-3}\cmidrule(lr){4-5}\cmidrule(lr){6-7}\cmidrule(lr){8-9}\cmidrule(lr){10-11}\cmidrule(lr){12-13}
Method & Pt. & AUROC & Pt. & AUROC & Pt. & AUROC & Pt. & AUROC & Pt. & AUROC & Pt. & AUROC \\
\midrule
\multicolumn{13}{@{}l}{\emph{Uses labels}} \\
ViCrop~\cite{zhang2025vicrop}, tuned layer$^\dagger$ & 0.486$^*$ & 0.725$^*$ & 0.687$^*$ & 0.827$^*$ & 0.588$^*$ & 0.738$^*$ & 0.827 & 0.890$^*$ & 0.885 & 0.897$^*$ & 0.911 & 0.810$^*$ \\
3 heads chosen with labels & 0.361$^*$ & 0.709$^*$ & 0.707 & 0.839 & 0.438 & 0.680$^*$ & 0.639$^*$ & 0.864$^*$ & 0.761$^*$ & 0.890$^*$ & 0.794$^*$ & 0.901$^*$ \\
\multicolumn{13}{@{}l}{\emph{Label-free}} \\
\method (ours) & 0.537 & 0.762 & 0.718 & 0.854 & 0.463 & 0.755 & 0.819 & 0.898 & 0.891 & 0.913 & 0.908 & 0.877 \\
\method, per-question layers & 0.534 & 0.761 & 0.702 & 0.854 & 0.486 & 0.729$^*$ & 0.809 & 0.896 & 0.883 & 0.911$^*$ & 0.888$^*$ & 0.865 \\
\bottomrule
\end{tabular}\end{adjustbox}
\caption{\textbf{COCO, held-out (393 questions)}: pointing (Pt.) and token AUROC per model. $^*$Significantly different from \method (paired two-sided Wilcoxon, $p<0.05$). $^\dagger$Layer tuned with labels on the odd ids.}
\label{tab:s_loc_coco}
\end{table*}
\begin{table*}[!tbp]
\centering
\footnotesize
\setlength{\tabcolsep}{3pt}
\begin{adjustbox}{max width=\textwidth}\begin{tabular}{@{}lcccccccccccc@{}}
\toprule
 & \multicolumn{2}{c}{SmolVLM-256M} & \multicolumn{2}{c}{SmolVLM-500M} & \multicolumn{2}{c}{InternVL3-1B} & \multicolumn{2}{c}{Qwen3-VL-2B} & \multicolumn{2}{c}{Qwen3-VL-8B} & \multicolumn{2}{c}{LLaVA-1.5-7B} \\
\cmidrule(lr){2-3}\cmidrule(lr){4-5}\cmidrule(lr){6-7}\cmidrule(lr){8-9}\cmidrule(lr){10-11}\cmidrule(lr){12-13}
Method & Pt. & AUROC & Pt. & AUROC & Pt. & AUROC & Pt. & AUROC & Pt. & AUROC & Pt. & AUROC \\
\midrule
\multicolumn{13}{@{}l}{\emph{Uses labels}} \\
ViCrop~\cite{zhang2025vicrop}, tuned layer$^\dagger$ & 0.435 & 0.649 & 0.460$^*$ & 0.657$^*$ & 0.702 & 0.784$^*$ & 0.818 & 0.806$^*$ & 0.898 & 0.837$^*$ & 0.628 & 0.742$^*$ \\
3 heads chosen with labels & 0.404 & 0.680$^*$ & 0.463 & 0.724$^*$ & 0.505$^*$ & 0.754$^*$ & 0.712$^*$ & 0.813 & 0.863 & 0.900$^*$ & 0.705 & 0.840$^*$ \\
\multicolumn{13}{@{}l}{\emph{Label-free}} \\
\method (ours) & 0.460 & 0.638 & 0.533 & 0.680 & 0.726 & 0.793 & 0.786 & 0.812 & 0.888 & 0.852 & 0.649 & 0.754 \\
\method, per-question layers & 0.460 & 0.632 & 0.481 & 0.642$^*$ & 0.625$^*$ & 0.764$^*$ & 0.723$^*$ & 0.777$^*$ & 0.874 & 0.829$^*$ & 0.596 & 0.732$^*$ \\
\bottomrule
\end{tabular}\end{adjustbox}
\caption{\textbf{RefCOCO testA/testB, held-out (285 expressions)}, as in \cref{tab:s_loc_coco}. The layer prior is the one estimated on COCO.}
\label{tab:s_loc_refcoco}
\end{table*}
\begin{table*}[!tbp]
\centering
\footnotesize
\setlength{\tabcolsep}{3pt}
\begin{adjustbox}{max width=\textwidth}\begin{tabular}{@{}lcccccccccc@{}}
\toprule
Model & Chance & \method (ours) & \method, per-question & ViCrop~\cite{zhang2025vicrop}, tuned layer$^\dagger$ & ViCrop, layer band & ViCrop, all layers & CARVE~\cite{ge2025carve} & Chefer relevance~\cite{chefer2021generic} & Attention (mean) & KL-weighted map \\
\midrule
SmolVLM-500M & 0.029 & 0.042 / 0.629 & 0.063 / 0.650 & 0.026 / 0.577$^*$ & 0.084$^*$ / 0.637 & \textbf{0.094}$^*$ / 0.620 & 0.089$^*$ / \textbf{0.690}$^*$ & 0.000$^*$ / 0.622 & 0.000$^*$ / 0.644 & 0.026 / 0.623 \\
InternVL3-1B & 0.011 & 0.042 / 0.724 & 0.031 / 0.722 & \textbf{0.063} / 0.754$^*$ & 0.037 / 0.731 & 0.047 / 0.728 & 0.042 / \textbf{0.769}$^*$ & 0.005$^*$ / 0.725 & 0.000$^*$ / 0.668$^*$ & 0.005$^*$ / 0.736 \\
Qwen3-VL-2B & 0.013 & 0.387 / \textbf{0.910} & 0.382 / 0.905 & \textbf{0.435} / 0.910 & 0.351 / 0.894$^*$ & 0.351 / 0.893$^*$ & 0.429 / 0.844$^*$ & 0.126$^*$ / 0.797$^*$ & 0.005$^*$ / 0.781$^*$ & 0.340 / 0.833$^*$ \\
Qwen3-VL-8B & 0.013 & 0.372 / 0.900 & 0.398 / 0.885 & \textbf{0.461}$^*$ / \textbf{0.901} & 0.398 / 0.885$^*$ & 0.398 / 0.889$^*$ & 0.335 / 0.795$^*$ & 0.005$^*$ / 0.781$^*$ & 0.005$^*$ / 0.759$^*$ & 0.225$^*$ / 0.816$^*$ \\
LLaVA-1.5-7B & 0.007 & 0.209 / \textbf{0.783} & 0.194 / 0.745$^*$ & \textbf{0.230} / 0.628$^*$ & 0.110$^*$ / 0.550$^*$ & 0.131$^*$ / 0.568$^*$ & 0.037$^*$ / 0.701$^*$ & 0.000$^*$ / 0.734$^*$ & 0.000$^*$ / 0.723$^*$ & 0.042$^*$ / 0.776 \\
\bottomrule
\end{tabular}
\end{adjustbox}
\caption{\textbf{V$^*$ localization} (191 questions): pointing / token AUROC per model; chance = expected pointing of a random token. Bold: best per model and metric. $^\dagger$Layer tuned on COCO labels.}
\label{tab:s_vstar_loc}
\end{table*}

\begin{figure*}[!tbp]
\centering
\includegraphics[width=\linewidth]{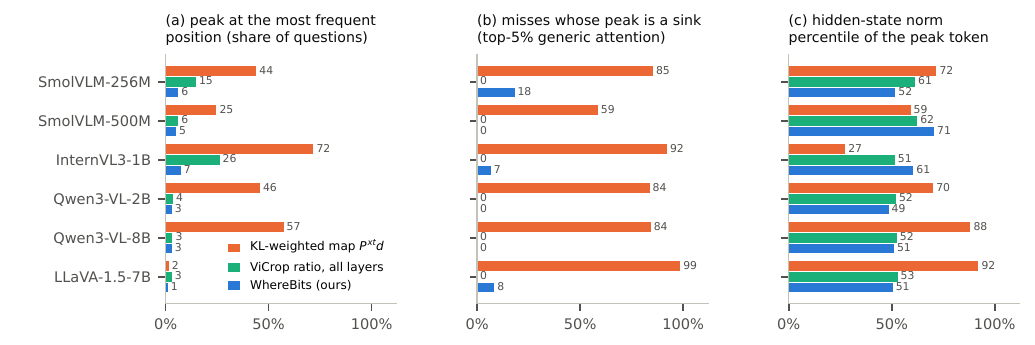}
\caption{\textbf{Where the peaks go} (held-out COCO). (a)~Share of questions whose peak lies at the map's most frequent position. (b)~Share of misses whose peak is a sink. (c)~Hidden-state norm percentile of the peak token at 60\% depth.}
\label{fig:s_mechanism}
\end{figure*}

\paragraph{Sinks.}
For five models, a single corner contains the mean-attention peak on 44--89\% of held-out COCO questions (\cref{tab:s_sinks,fig:s_mechanism}). LLaVA-1.5 peaks vary more across images, but its mean-attention peak still falls on a generic-prompt sink on every V$^*$ miss. ViCrop with all layers produces no COCO misses at generic-prompt sinks.

KL-weighted peaks have above-median hidden-state norms in five of six models. The norm percentiles span 27--92 across models, and \method has higher peak norm percentiles in two models. The results give limited support to hidden-state norm alone as a detector of these sinks~\cite{sun2024massive}.

If sinks are defined from the gray image, 89--97\% of COCO misses from mean attention land on them in five models, and 7\% in LLaVA-1.5. The corresponding \method range is 0--15\%. If sinks are defined by question-conditioned attention, 92--100\% of KL-weighted misses and 17--30\% of \method misses land on them.

\begin{table*}[!tbp]
\centering
\footnotesize
\begin{adjustbox}{max width=\textwidth}\begin{tabular}{@{}lcccccccccccc@{}}
\toprule
 & \multicolumn{7}{c}{COCO, held-out} & \multicolumn{4}{c}{V$^*$} \\
\cmidrule(lr){2-8}\cmidrule(lr){9-12}
 & \multicolumn{2}{c}{attention peak} & \multicolumn{5}{c}{misses on a sink (\%)} & \multicolumn{2}{c}{one position (\%)} & \multicolumn{2}{c}{misses on a sink (\%)} \\
\cmidrule(lr){2-3}\cmidrule(lr){4-8}\cmidrule(lr){9-10}\cmidrule(lr){11-12}
Model & token & share & attn. & Chefer~\cite{chefer2021generic} & KL-w. & ViCrop~\cite{zhang2025vicrop} & ours & attn. & ours & attn. & ours \\
\midrule
SmolVLM-256M & (7,7) & 44 & 95 & 89 & 85 & 0 & 18 & -- & -- & -- & -- \\
SmolVLM-500M & (7,7) & 68 & 93 & 71 & 59 & 0 & 0 & 100 & 17 & 100 & 0.5 \\
InternVL3-1B & (15,15) & 80 & 97 & 88 & 92 & 0 & 7 & 100 & 4 & 100 & 1.1 \\
Qwen3-VL-2B & (0,13) & 45 & 90 & 91 & 84 & 0 & 0 & 46 & 4 & 100 & 0.0 \\
Qwen3-VL-8B & (0,0) & 89 & 98 & 98 & 84 & 0 & 0 & 92 & 4 & 100 & 0.0 \\
LLaVA-1.5-7B & (19,22) & 2 & 100 & 98 & 99 & 0 & 8 & 2 & 2 & 100 & 1.3 \\
\bottomrule
\end{tabular}\end{adjustbox}
\caption{\textbf{Sinks.} A sink is one of the 5\% of visual tokens with the highest attention under the generic prompt. COCO: grid position (row, column) of the most frequent mean-attention peak and the share (\%) of held-out questions on which it holds the peak; share of misses whose peak is a sink for mean attention (attn.), Chefer relevance, the KL-weighted map (KL-w.), ViCrop over all layers and \method (ours). V$^*$: share of questions whose peak lies at the map's single most frequent position, and share of misses on a sink.}
\label{tab:s_sinks}
\end{table*}

\begin{figure*}[!tbp]
\centering
\includegraphics[width=\linewidth]{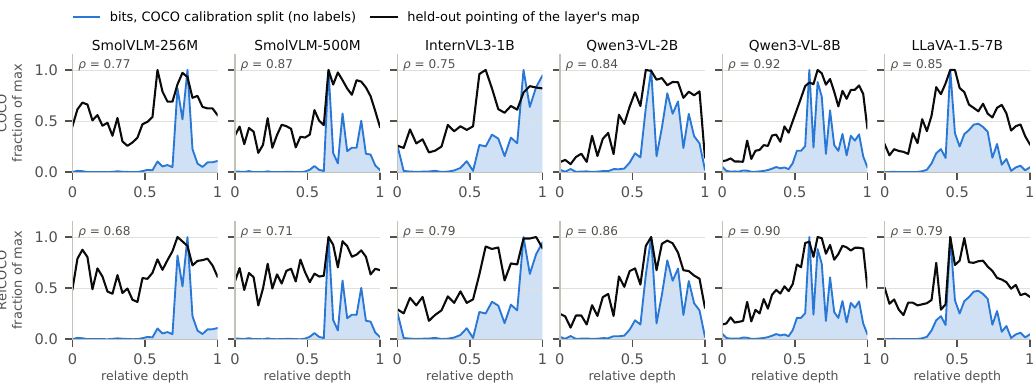}
\caption{\textbf{Bits and per-layer pointing in every model.} Blue: where-to-look bits per layer, $\sum_hw_{lh}$ averaged over the COCO calibration split (no labels; the only input of the prior). Black: pointing of each layer's map $F_l$ on held-out COCO (top) and RefCOCO (bottom). $\rho$: Spearman correlation over layers.}
\label{fig:s_layers}
\end{figure*}

\begin{figure*}[!tbp]
\centering
\includegraphics[width=\linewidth]{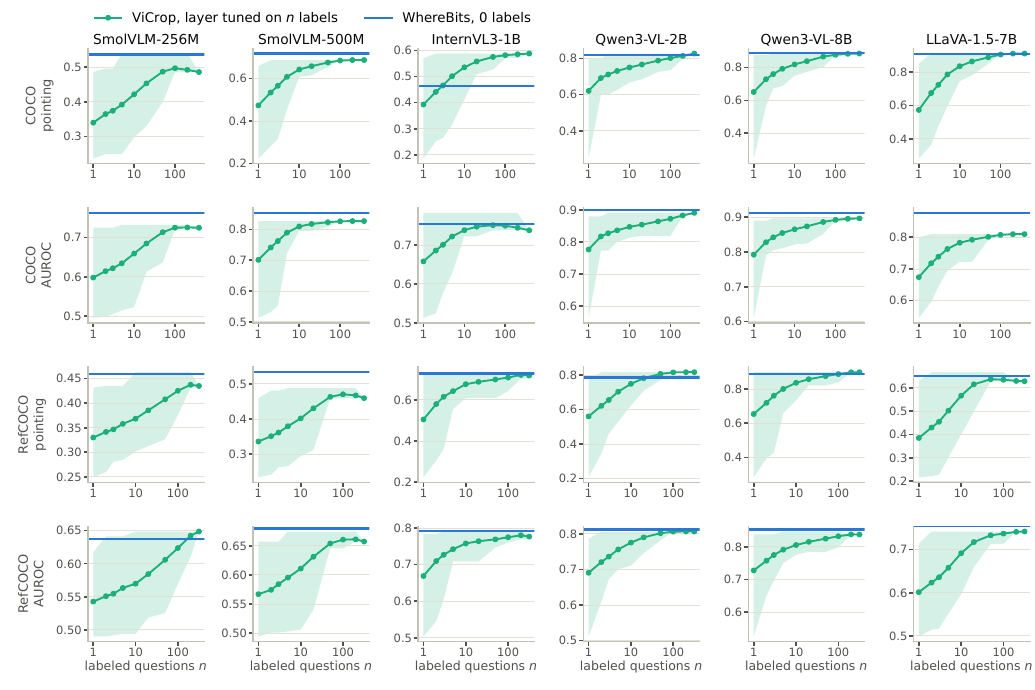}
\caption{\textbf{Label efficiency of layer tuning.} A ViCrop layer is chosen on $n$ random labeled questions of the calibration split and evaluated on held-out data (line: mean over 2000 draws; band: 10--90th percentile). Blue: \method with no labels.}
\label{fig:s_label_eff}
\end{figure*}

\paragraph{Layers and labels.}
\Cref{fig:s_layers} extends \cref{fig:layers}a--c to all models and to RefCOCO; \cref{fig:s_label_eff} extends \cref{fig:layers}e to all 12 model--dataset pairs and both metrics.


\paragraph{Further ablations.}
\Cref{tab:s_ablation_more} lists five single-factor ablations that complement \cref{tab:ablation}. Removing the gray-image correction from the head weights alone lowers pointing by 0.010 on COCO and 0.001 on RefCOCO, and removing it from both the head weights and the prior costs about as much as removing it from the prior alone. Head weights from the KL term without the visual mass lose about 0.01 on each metric. A prior exponent of $\gamma=1$ instead of 4 lowers pointing by 0.014 and 0.023. Using the double difference $d-d^{\varnothing}$ as the map, that is, subtracting the gray-image log-ratio from each token score, lowers pointing by 0.059 and 0.084.

\begin{table}[!tbp]
\centering
\footnotesize
\setlength{\tabcolsep}{2.4pt}
\begin{adjustbox}{max width=\linewidth}\begin{tabular}{@{}lccccc@{}}
\toprule
 & \multicolumn{2}{c}{COCO} & \multicolumn{2}{c}{RefCOCO} & \\
\cmidrule(lr){2-3}\cmidrule(lr){4-5}
Variant & $\Delta$Pt. & $\Delta$AUROC & $\Delta$Pt. & $\Delta$AUROC & $-$\,/\,$+$ \\
\midrule
No gray correction in head weights & $-0.010$ & $-0.007$ & $-0.001$ & $-0.003$ & 10 / 1 \\
No gray correction in head weights or prior & $-0.058$ & $-0.043$ & $-0.087$ & $-0.045$ & 11 / 1 \\
Head weights from KL only, no mass $m$ & $-0.010$ & $-0.009$ & $-0.012$ & $-0.006$ & 10 / 2 \\
Double difference $d-d^{\varnothing}$ as the map & $-0.059$ & $-0.050$ & $-0.084$ & $-0.073$ & 22 / 1 \\
Prior exponent $\gamma=1$ instead of 4 & $-0.014$ & $-0.008$ & $-0.023$ & $-0.012$ & 10 / 2 \\
\bottomrule
\end{tabular}\end{adjustbox}
\caption{\textbf{Further single-factor ablations.} Mean change from \method over six models, as in \cref{tab:ablation}; last column: significant losses / gains in 24 model--dataset--metric comparisons ($p<0.05$, uncorrected).}
\label{tab:s_ablation_more}
\end{table}

\paragraph{Faithfulness.}
\Cref{tab:s_faith_token,tab:s_faith_bit} give insertion$-$deletion AUC per model and setting.

\begin{table*}[!tbp]
\centering
\footnotesize
\setlength{\tabcolsep}{3.5pt}
\begin{adjustbox}{max width=\textwidth}\begin{tabular}{@{}lcccccccccc@{}}
\toprule
 & \multicolumn{6}{c}{gray} & \multicolumn{4}{c}{blur} \\
\cmidrule(lr){2-7}\cmidrule(lr){8-11}
Map & \shortstack{SmolVLM\\256M} & \shortstack{SmolVLM\\500M} & \shortstack{InternVL3\\1B} & \shortstack{Qwen3-VL\\2B} & \shortstack{Qwen3-VL\\8B} & \shortstack{LLaVA-1.5\\7B} & \shortstack{SmolVLM\\256M} & \shortstack{SmolVLM\\500M} & \shortstack{InternVL3\\1B} & \shortstack{Qwen3-VL\\2B} \\
\midrule
\method (ours) & 23.5 & 36.5 & 21.4 & 39.5 & 39.7 & 35.6 & 23.2 & 36.0 & 21.2 & 38.7 \\
\method, per-question & 23.2 & 37.2 & 20.5 & 39.3 & 40.0 & 34.5 & 23.2 & 36.1 & 20.6 & 39.3 \\
ViCrop~\cite{zhang2025vicrop}, tuned$^\dagger$ & 17.9$^*$ & 34.2$^*$ & 16.7$^*$ & 38.0$^*$ & 40.0 & 28.1$^*$ & 18.8$^*$ & 33.6$^*$ & 17.7$^*$ & 37.9$^*$ \\
ViCrop, all layers & 16.4$^*$ & 34.9$^*$ & 19.7$^*$ & 35.7$^*$ & 36.1$^*$ & 17.3$^*$ & 16.5$^*$ & 34.0$^*$ & 19.4$^*$ & 35.3$^*$ \\
CARVE~\cite{ge2025carve} (re-impl.) & 19.2$^*$ & 33.7$^*$ & 9.2$^*$ & 23.1$^*$ & 24.5$^*$ & 20.1$^*$ & 18.6$^*$ & 33.6$^*$ & 9.5$^*$ & 23.8$^*$ \\
Chefer relevance~\cite{chefer2021generic} & 17.3$^*$ & 29.2$^*$ & 9.4$^*$ & 16.4$^*$ & 14.9$^*$ & 27.2$^*$ & 16.5$^*$ & 27.8$^*$ & 9.1$^*$ & 16.2$^*$ \\
codec bits (texture) & 11.6$^*$ & 13.1$^*$ & 15.9$^*$ & 18.4$^*$ & 20.5$^*$ & 19.4$^*$ & 12.8$^*$ & 13.7$^*$ & 13.9$^*$ & 15.8$^*$ \\
Random & -1.9$^*$ & 1.6$^*$ & 0.8$^*$ & 3.9$^*$ & 3.5$^*$ & 3.5$^*$ & -0.5$^*$ & 1.3$^*$ & -0.6$^*$ & 2.2$^*$ \\
\bottomrule
\end{tabular}\end{adjustbox}
\caption{\textbf{Faithfulness, token axis}: insertion$-$deletion AUC ($\times100$, higher is better) on held-out COCO. $^*$Significantly different from \method (paired two-sided Wilcoxon, $p<0.05$, uncorrected).}
\label{tab:s_faith_token}
\end{table*}
\begin{table*}[!tbp]
\centering
\footnotesize
\setlength{\tabcolsep}{3.5pt}
\begin{adjustbox}{max width=\textwidth}\begin{tabular}{@{}lcccccccccc@{}}
\toprule
 & \multicolumn{6}{c}{gray} & \multicolumn{4}{c}{blur} \\
\cmidrule(lr){2-7}\cmidrule(lr){8-11}
Map & \shortstack{SmolVLM\\256M} & \shortstack{SmolVLM\\500M} & \shortstack{InternVL3\\1B} & \shortstack{Qwen3-VL\\2B} & \shortstack{Qwen3-VL\\8B} & \shortstack{LLaVA-1.5\\7B} & \shortstack{SmolVLM\\256M} & \shortstack{SmolVLM\\500M} & \shortstack{InternVL3\\1B} & \shortstack{Qwen3-VL\\2B} \\
\midrule
\method (ours) & 22.0 & 34.6 & 19.0 & 35.1 & 33.9 & 33.9 & 21.8 & 33.9 & 19.0 & 34.7 \\
\method, per-question & 21.4 & 35.5 & 19.3 & 35.1 & 34.4 & 33.1 & 21.5 & 34.3 & 19.4 & 35.4 \\
ViCrop~\cite{zhang2025vicrop}, tuned$^\dagger$ & 16.5$^*$ & 31.6$^*$ & 16.1$^*$ & 34.5 & 33.3 & 27.2$^*$ & 17.3$^*$ & 31.0$^*$ & 17.1 & 34.7 \\
ViCrop, all layers & 12.0$^*$ & 32.1$^*$ & 16.0$^*$ & 30.6$^*$ & 29.4$^*$ & 19.2$^*$ & 12.3$^*$ & 31.2$^*$ & 16.1$^*$ & 30.7$^*$ \\
CARVE~\cite{ge2025carve} (re-impl.) & 18.4$^*$ & 33.1$^*$ & 13.0$^*$ & 25.0$^*$ & 26.7$^*$ & 19.7$^*$ & 17.9$^*$ & 32.9 & 13.0$^*$ & 25.1$^*$ \\
Chefer relevance~\cite{chefer2021generic} & 18.3$^*$ & 30.0$^*$ & 14.0$^*$ & 21.0$^*$ & 18.0$^*$ & 26.1$^*$ & 17.6$^*$ & 28.6$^*$ & 13.3$^*$ & 19.9$^*$ \\
codec bits (texture) & -12.7$^*$ & -11.0$^*$ & -10.9$^*$ & -11.0$^*$ & -9.6$^*$ & -6.8$^*$ & -11.3$^*$ & -9.5$^*$ & -9.9$^*$ & -9.9$^*$ \\
Random & -2.0$^*$ & 1.4$^*$ & 0.8$^*$ & 4.1$^*$ & 3.6$^*$ & 3.6$^*$ & -0.5$^*$ & 1.2$^*$ & -0.5$^*$ & 2.3$^*$ \\
\bottomrule
\end{tabular}\end{adjustbox}
\caption{\textbf{Faithfulness, codec-bit axis}, as in \cref{tab:s_faith_token}. The texture-only codec-bit map falls below random in every setting.}
\label{tab:s_faith_bit}
\end{table*}

\paragraph{Zoom-in.}
\Cref{tab:s_vstar_fc,tab:s_textvqa_zoom} give per-model zoom-in accuracy under the primary protocols, and \cref{tab:s_fc_checks} checks the forced choice. Without zoom, the forced-choice letter equals the letter that the model generates on 92--100\% of questions for InternVL3-1B, Qwen3-VL and LLaVA-1.5-7B. SmolVLM-500M starts every answer with ``Answer:'', puts a median 0.4\% of its next-token probability on the option letters, and agrees on 46\% of questions; it is near chance under both protocols. Given two images, Qwen3-VL rarely starts with a letter (the top token is a letter for 6\% of Qwen3-VL-2B's crop conditions), and its generated answers often contain none: 31--59\% of the zoom-in answers of Qwen3-VL-2B and 10--24\% of those of Qwen3-VL-8B (1\% or less with one image), which put every generated zoom-in condition below no zoom for Qwen3-VL-2B (\cref{tab:s_vstar_zoom}). Under forced choice, \method' crops gain 3.7 and 7.3 points over no zoom for this model at the two windows.
On V$^*$ the peak of mean attention lies on the target for 0--1 of the 191 questions per model, and the layer that ViCrop tunes on COCO is also the best single ViCrop layer on V$^*$ for Qwen3-VL-2B (L17), Qwen3-VL-8B (L24, tied with L23) and LLaVA-1.5-7B (L14), but not for SmolVLM-500M (L20 vs.\ L28) or InternVL3-1B (L14 vs.\ L15).

\begin{table*}[!tbp]
\centering
\footnotesize
\setlength{\tabcolsep}{3.5pt}
\begin{adjustbox}{max width=\textwidth}\begin{tabular}{@{}llccccccccc@{}}
\toprule
Win. & Model & No zoom & Center & Attention & Chefer~\cite{chefer2021generic} & CARVE~\cite{ge2025carve} & ViCrop-all~\cite{zhang2025vicrop} & ViCrop$^\dagger$ & \method & + bits window \\
\midrule
1/2 & SmolVLM-500M & 35.1 & 37.2 & 35.6 & 35.6 & 35.6 & 37.2 & 38.2 & \textbf{39.3} & 37.7 \\
1/2 & InternVL3-1B & 41.9 & 43.5 & 48.7 & 48.7 & 43.5 & 45.0 & 46.1 & 42.4 & \textbf{49.7} \\
1/2 & Qwen3-VL-2B & 49.2 & 48.7 & 52.4 & 53.9 & \textbf{55.5} & 52.9 & \textbf{55.5} & 52.9 & 51.8 \\
1/2 & Qwen3-VL-8B & 47.6 & 53.9 & 49.2 & 49.7 & 61.3 & 60.7 & 59.7 & \textbf{62.3} & 58.6 \\
1/2 & LLaVA-1.5-7B & 47.1 & 42.4 & 42.9 & 47.1 & 47.1 & 43.5 & 51.3 & 50.8 & \textbf{51.8} \\
\midrule
1/3 & SmolVLM-500M & 35.1 & \textbf{37.7} & 37.2 & 37.2 & 36.6 & 35.1 & 36.1 & 35.1 & 37.2 \\
1/3 & InternVL3-1B & 41.9 & 42.4 & 47.6 & \textbf{48.7} & 42.4 & 41.9 & 43.5 & 40.8 & 48.2 \\
1/3 & Qwen3-VL-2B & 49.2 & 49.2 & 46.1 & 50.3 & 56.5 & \textbf{58.1} & 56.0 & 56.5 & 56.5 \\
1/3 & Qwen3-VL-8B & 47.6 & 53.9 & 47.1 & 48.2 & 60.7 & 64.4 & 63.4 & \textbf{64.9} & \textbf{64.9} \\
1/3 & LLaVA-1.5-7B & 47.1 & 39.3 & 40.3 & 47.1 & 47.6 & 41.9 & 54.5 & 56.5 & \textbf{57.1} \\
\midrule
\multicolumn{2}{l}{\method $-$ x (pooled)} & +6.0 & +5.3 & +5.4 & +3.5 & +1.5 & +2.1 & -0.3 & -- & -- \\
\multicolumn{2}{l}{bits window $-$ x (pooled)} & +7.2 & +6.5 & +6.6 & +4.7 & +2.7 & +3.3 & +0.9 & +1.2 & -- \\
\bottomrule
\end{tabular}\end{adjustbox}
\caption{\textbf{V$^*$ zoom-in accuracy under forced choice} (\%, 191 questions; chance 36.0\%). ``\method'': primary rule; ``+ bits window'': the crop rule of \cref{sec:bitswin}. Win.: window side (fraction of the image side). Pooled rows: mean paired difference over five models and both windows (intervals in \cref{tab:s_vstar_fc_pooled}). $^\dagger$Layer tuned with COCO labels.}
\label{tab:s_vstar_fc}
\end{table*}
\begin{table}[!tbp]
\centering
\footnotesize
\setlength{\tabcolsep}{2.4pt}
\begin{adjustbox}{max width=\linewidth}\begin{tabular}{@{}lcccccc@{}}
\toprule
& \multicolumn{2}{c}{No zoom: agreement} & \multicolumn{2}{c}{Letter mass (median)} & \multicolumn{2}{c}{Top token is a letter (\%)} \\
\cmidrule(lr){2-3}\cmidrule(lr){4-5}\cmidrule(l){6-7}
Model & \% & $n$ & no zoom & crops & no zoom & crops \\
\midrule
SmolVLM-500M & 46 & 191 & 0.00 & 0.01 & 0 & 0 \\
InternVL3-1B & 100 & 191 & 1.00 & 1.00 & 100 & 100 \\
Qwen3-VL-2B & 92 & 190 & 0.22 & 0.13 & 42 & 6 \\
Qwen3-VL-8B & 99 & 189 & 0.80 & 0.37 & 81 & 42 \\
LLaVA-1.5-7B & 100 & 191 & 0.93 & 0.93 & 100 & 100 \\
\midrule
Qwen3-VL-2B, 896\,px & 94 & 191 & 0.10 & 0.11 & 39 & 0 \\
Qwen3-VL-2B, native & 96 & 189 & 0.07 & 0.10 & 36 & 1 \\
Qwen3-VL-8B, 896\,px & 99 & 188 & 0.77 & 0.41 & 80 & 45 \\
Qwen3-VL-8B, native & 100 & 190 & 0.76 & 0.44 & 72 & 48 \\
\bottomrule
\end{tabular}\end{adjustbox}
\caption{\textbf{Checks of the forced choice on V$^*$.} Agreement: share of no-zoom questions on which the forced-choice letter equals the first option letter of the generated answer ($n$: questions whose generated answer has a letter). Letter mass: next-token probability of all option letters together (median over questions, or over all crop conditions). Top token: share of conditions whose most likely next token is an option letter.}
\label{tab:s_fc_checks}
\end{table}
\begin{table*}[!tbp]
\centering
\footnotesize
\setlength{\tabcolsep}{3.5pt}
\begin{adjustbox}{max width=\textwidth}\begin{tabular}{@{}llccccccccc@{}}
\toprule
Win. & Model & No zoom & Center & Attention & Chefer~\cite{chefer2021generic} & CARVE~\cite{ge2025carve} & ViCrop-all~\cite{zhang2025vicrop} & ViCrop$^\dagger$ & \method & + bits window \\
\midrule
1/2 & SmolVLM-500M & 29.8 & 28.8 & 30.4 & 30.4 & 33.5 & 34.0 & \textbf{36.6} & 36.1 & 31.9 \\
1/2 & InternVL3-1B & 41.9 & 43.5 & 48.7 & 48.7 & 43.5 & 45.0 & 46.1 & 42.4 & \textbf{49.7} \\
1/2 & Qwen3-VL-2B & 47.6 & 25.7 & 31.4 & 33.0 & \textbf{38.7} & 34.0 & 32.5 & 33.5 & 34.0 \\
1/2 & Qwen3-VL-8B & 47.1 & 50.3 & 43.5 & 42.9 & 57.6 & 57.1 & 57.6 & \textbf{58.1} & 57.1 \\
1/2 & LLaVA-1.5-7B & 47.1 & 42.4 & 42.9 & 47.1 & 47.1 & 43.5 & 51.3 & 50.8 & \textbf{51.8} \\
\midrule
1/3 & SmolVLM-500M & 29.8 & 30.9 & 32.5 & 32.5 & 32.5 & 31.4 & \textbf{33.5} & \textbf{33.5} & 31.9 \\
1/3 & InternVL3-1B & 41.9 & 42.4 & 47.6 & \textbf{48.7} & 42.4 & 41.9 & 43.5 & 40.8 & 48.2 \\
1/3 & Qwen3-VL-2B & 47.6 & 29.3 & 29.3 & 34.0 & 45.5 & 44.0 & 45.0 & 41.9 & \textbf{46.6} \\
1/3 & Qwen3-VL-8B & 47.1 & 46.6 & 38.7 & 39.8 & 55.0 & 59.2 & 59.2 & 59.2 & \textbf{61.8} \\
1/3 & LLaVA-1.5-7B & 47.1 & 39.3 & 40.3 & 47.1 & 47.6 & 41.9 & 54.5 & 56.5 & \textbf{57.1} \\
\midrule
\multicolumn{2}{l}{\method $-$ x (pooled)} & +2.6 & +7.4 & +6.8 & +4.9 & +0.9 & +2.1 & -0.7 & -- & -- \\
\multicolumn{2}{l}{bits window $-$ x (pooled)} & +4.3 & +9.1 & +8.5 & +6.6 & +2.7 & +3.8 & +1.0 & +1.7 & -- \\
\bottomrule
\end{tabular}
\end{adjustbox}
\caption{\textbf{V$^*$ zoom-in accuracy with generated answers} (\%; first option letter parsed from up to ten generated tokens, answers without a letter counted as wrong). Conventions as in \cref{tab:s_vstar_fc}.}
\label{tab:s_vstar_zoom}
\end{table*}
\begin{table*}[!tbp]
\centering
\footnotesize
\setlength{\tabcolsep}{3pt}
\begin{adjustbox}{max width=\textwidth}\begin{tabular}{@{}llccccccccc@{}}
\toprule
Window & Model & No zoom & Center crop & ViCrop~\cite{zhang2025vicrop}, tuned layer$^\dagger$ & ViCrop, all layers & CARVE~\cite{ge2025carve} & Chefer relevance~\cite{chefer2021generic} & Attention (mean) & \method, primary rule & \method, bits window \\
\midrule
0.5 & SmolVLM-500M & 41.7 & 46.7 & 52.5 & 52.3 & \textbf{54.1} & 53.0 & 53.5 & 52.1 & 52.5 \\
0.5 & InternVL3-1B & 62.4 & 66.2 & 67.9 & 69.6 & 71.2 & 71.8 & \textbf{72.0} & 68.6 & 70.1 \\
0.5 & Qwen3-VL-2B & 72.0 & 72.2 & 73.5 & 73.7 & \textbf{76.1} & 73.8 & 75.6 & 75.2 & 75.2 \\
0.5 & LLaVA-1.5-7B & 49.2 & 37.6 & 56.3 & 56.4 & 58.4 & \textbf{59.0} & 54.3 & 57.3 & 57.3 \\
\midrule
0.33 & SmolVLM-500M & 41.7 & 46.2 & 47.9 & 49.0 & 50.4 & 50.2 & \textbf{50.7} & 48.6 & 49.4 \\
0.33 & InternVL3-1B & 62.4 & 64.1 & 67.4 & 67.4 & 70.1 & 71.7 & \textbf{71.7} & 67.8 & 70.3 \\
0.33 & Qwen3-VL-2B & 72.0 & 70.4 & 75.5 & 75.3 & 75.9 & 75.7 & 76.3 & 76.2 & \textbf{77.6} \\
0.33 & LLaVA-1.5-7B & 49.2 & 28.3 & 52.4 & 54.6 & \textbf{57.0} & 56.2 & 51.7 & 52.7 & 55.0 \\
\midrule
\multicolumn{2}{l}{bits window $-$ x, pooled} & +7.1 &  & +1.7 & +1.1 & -0.7 & -0.5 & +0.2 & +1.1 & -- \\
\bottomrule
\end{tabular}
\end{adjustbox}
\caption{\textbf{TextVQA zoom-in accuracy} (\%, VQA accuracy; 500 questions; the no-zoom score of SmolVLM-500M is over 490). $^\dagger$Layer tuned with COCO labels.}
\label{tab:s_textvqa_zoom}
\end{table*}

\begin{figure*}[p]
\centering
\includegraphics[width=\linewidth]{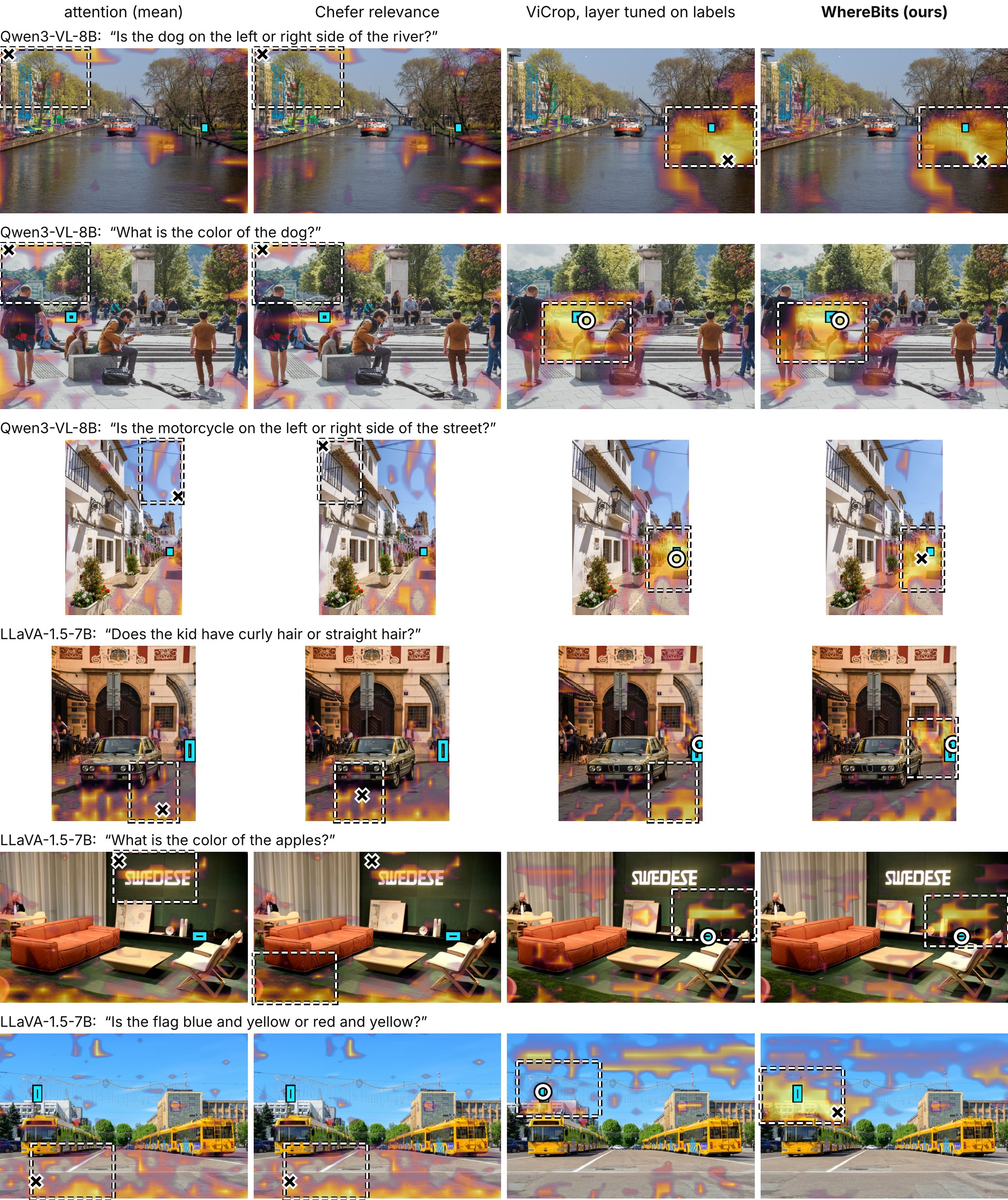}
\caption{\textbf{Random V$^*$ questions} (seed 0). Cyan box: target; white circle / black cross: map peak on / off the target; dashed: the $1/3$ crop of the primary rule (not drawn where it coincides with the panel border).}
\label{fig:s_vstar_gallery}
\end{figure*}

\begin{figure*}[p]
\centering
\includegraphics[width=0.80\textwidth]{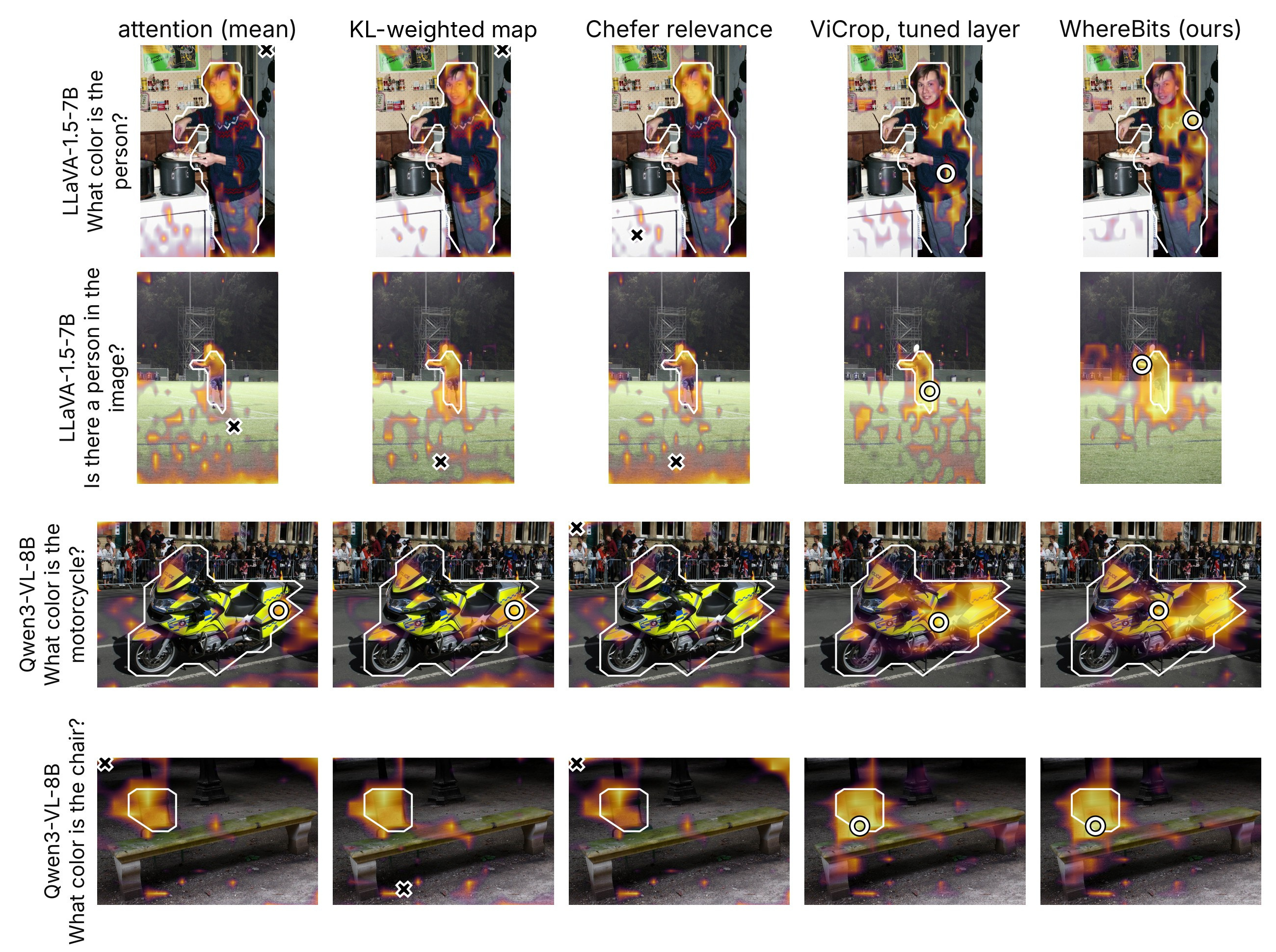}\\[2pt]
\includegraphics[width=0.80\textwidth]{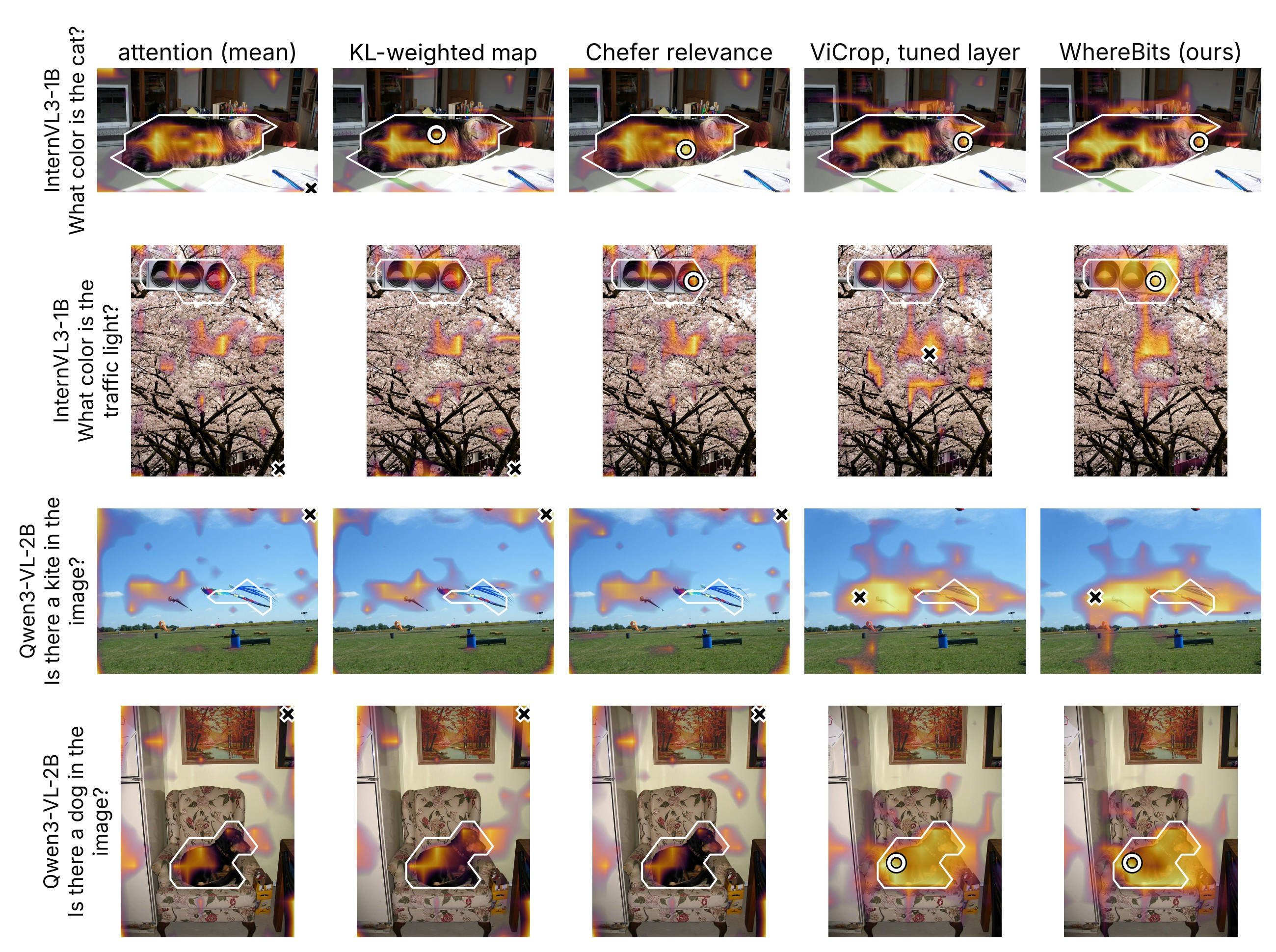}
\caption{\textbf{Random held-out COCO questions} (seed 0; two per model for LLaVA-1.5-7B, Qwen3-VL-8B, InternVL3-1B and Qwen3-VL-2B). White outline: object mask; white circle / black cross: map peak on / off the object. Maps are shown as within-image percentiles.}
\label{fig:s_coco_gallery}
\end{figure*}

\paragraph{Qualitative results.}
\Cref{fig:s_vstar_gallery,fig:s_coco_gallery} show V$^*$ and held-out COCO questions drawn at random (seed 0, no selection on outcomes), including misses of \method.

\section{Analyses added after the protocol was fixed}
\label{sec:s_added}
The analyses in this section were added after the method and the evaluation protocol were fixed; the main paper marks their results with $^\S$. Some use additional forward passes or interventions.

\paragraph{Visual Activation by Query on shared attention rows.}
We implement LASER's VAQ localization rule~\cite{zhu2026laser} on the stored answer-start rows for the question and ViCrop's generic prompt. LASER's published reference omits the query. This experiment compares spatial maps with identical prompts and excludes LASER's crop and decoding stages. For each head, we take $D_{lh}=[a_{lh}^{xt}-a_{lh}^{xg}]_+$ over the visual keys. We select the $K$ heads with largest $\ell_2$ norm of $D_{lh}$ in each layer, and use their mean norm as the layer score. The map averages $D_{lh}$ over those heads at the highest-scoring layer. We use $K=3$ in \cref{tab:main} and vary $K$ and row normalization in \cref{tab:s_vaq}. The prior-swap variant keeps each VAQ layer map and combines them using the \method prior.

For raw VAQ with $K=3$, the paired improvements in pointing are 0.237 (95\% CI $[0.211,0.263]$), 0.271 ($[0.240,0.301]$), and 0.160 ($[0.132,0.188]$) on COCO, RefCOCO, and V$^*$, respectively. The corresponding AUROC improvements are 0.123 ($[0.116,0.130]$), 0.133 ($[0.125,0.141]$), and 0.202 ($[0.183,0.221]$). These intervals resample questions after averaging paired differences across models.

\begin{table}[!tbp]
\centering
\footnotesize
\setlength{\tabcolsep}{3pt}
\begin{adjustbox}{max width=\linewidth}\begin{tabular}{@{}lccc@{}}
\toprule
Map & COCO & RefCOCO & V$^*$ \\
\midrule
VAQ~\cite{zhu2026laser}, raw $K=1$ & 0.458 / 0.686 & 0.382 / 0.610 & 0.036 / 0.566 \\
VAQ, raw $K=3$ & 0.486 / 0.720 & 0.403 / 0.621 & 0.050 / 0.587 \\
VAQ, raw $K=5$ & 0.492 / 0.726 & 0.409 / 0.624 & 0.051 / 0.621 \\
VAQ, raw all & 0.510 / 0.745 & 0.399 / 0.611 & 0.047 / 0.681 \\
\addlinespace[2pt]
VAQ, normalized $K=1$ & 0.280 / 0.618 & 0.209 / 0.537 & 0.116 / 0.615 \\
VAQ, normalized $K=3$ & 0.327 / 0.679 & 0.230 / 0.552 & 0.131 / 0.675 \\
VAQ, normalized $K=5$ & 0.345 / 0.699 & 0.260 / 0.563 & 0.136 / 0.698 \\
VAQ, normalized all & 0.364 / 0.705 & 0.320 / 0.592 & 0.127 / 0.729 \\
\addlinespace[2pt]
Raw $K=3$ with our prior & 0.556 / 0.795 & 0.503 / 0.695 & 0.199 / 0.768 \\
\method & 0.723 / 0.843 & 0.674 / 0.755 & 0.210 / 0.789 \\
\bottomrule
\end{tabular}\end{adjustbox}
\caption{\textbf{VAQ sensitivity on shared prompts.} Pointing / AUROC averaged over models. Raw uses the visual subset of the full attention row. Normalized sums to one over visual tokens.}
\label{tab:s_vaq}
\end{table}

\paragraph{Image-cluster sensitivity.}
Some images contribute multiple COCO questions or RefCOCO expressions. We repeat the pooled comparison to the pointing-tuned ViCrop layer, resampling images and retaining all their questions. We average paired differences over models for each question and retain question weighting in the pooled mean. Across 20{,}000 bootstrap draws, image-level intervals give the same interpretation as question-level intervals (\cref{tab:s_image_ci}).

\begin{table}[!tbp]
\centering
\footnotesize
\setlength{\tabcolsep}{3pt}
\begin{adjustbox}{max width=\linewidth}\begin{tabular}{@{}lccc@{}}
\toprule
Dataset & Images & Pointing & AUROC \\
\midrule
COCO & 253 & $-0.008\;[-0.024,0.008]$ & $+0.029\;[0.026,0.032]$ \\
RefCOCO & 239 & $+0.017\;[-0.001,0.035]$ & $+0.009\;[0.006,0.012]$ \\
V$^*$ & 191 & $-0.032\;[-0.053,-0.012]$ & $+0.035\;[0.028,0.042]$ \\
\bottomrule
\end{tabular}
\end{adjustbox}
\caption{\textbf{Image-clustered localization intervals.} Paired mean differences (\method minus pointing-tuned ViCrop~\cite{zhang2025vicrop}) and 95\% bootstrap intervals.}
\label{tab:s_image_ci}
\end{table}

\begin{table*}[!t]
\centering
\footnotesize
\setlength{\tabcolsep}{3.5pt}
\begin{adjustbox}{max width=\textwidth}\begin{tabular}{@{}lccccccccc@{}}
\toprule
 & tuned & \multicolumn{2}{c}{\method (prior)} & \multicolumn{2}{c}{own map, tuned layer} & \multicolumn{2}{c}{prior from mass only} & \multicolumn{2}{c}{prior w/o gray corr.} \\
\cmidrule(lr){3-4}\cmidrule(lr){5-6}\cmidrule(lr){7-8}\cmidrule(lr){9-10}
Model & layer & Pt. & AUROC & Pt. & AUROC & Pt. & AUROC & Pt. & AUROC \\
\midrule
\multicolumn{10}{@{}l}{\emph{COCO, held-out}} \\
SmolVLM-256M & 17 & 0.537 & 0.762 & 0.542 & 0.767 & 0.506 & 0.763 & 0.529 & 0.762 \\
SmolVLM-500M & 20 & 0.718 & 0.854 & 0.712 & 0.849$^*$ & 0.374$^*$ & 0.636$^*$ & 0.718 & 0.854 \\
InternVL3-1B & 14 & 0.463 & 0.755 & 0.578$^*$ & 0.731$^*$ & 0.226$^*$ & 0.594$^*$ & 0.153$^*$ & 0.489$^*$ \\
Qwen3-VL-2B & 16 & 0.819 & 0.898 & 0.802 & 0.897 & 0.583$^*$ & 0.739$^*$ & 0.817 & 0.897$^*$ \\
Qwen3-VL-8B & 23 & 0.891 & 0.913 & 0.921$^*$ & 0.911 & 0.842$^*$ & 0.903$^*$ & 0.888 & 0.912$^*$ \\
LLaVA-1.5-7B & 14 & 0.908 & 0.877 & 0.893 & 0.879$^*$ & 0.272$^*$ & 0.524$^*$ & 0.911 & 0.877$^*$ \\
\multicolumn{10}{@{}l}{\emph{RefCOCO, held-out}} \\
SmolVLM-256M & 21 & 0.460 & 0.638 & 0.481 & 0.673$^*$ & 0.484 & 0.646$^*$ & 0.463 & 0.636$^*$ \\
SmolVLM-500M & 20 & 0.533 & 0.680 & 0.509 & 0.678 & 0.354$^*$ & 0.559$^*$ & 0.533 & 0.680 \\
InternVL3-1B & 21 & 0.726 & 0.793 & 0.709 & 0.791 & 0.319$^*$ & 0.626$^*$ & 0.214$^*$ & 0.498$^*$ \\
Qwen3-VL-2B & 17 & 0.786 & 0.812 & 0.800 & 0.806$^*$ & 0.540$^*$ & 0.662$^*$ & 0.782 & 0.812 \\
Qwen3-VL-8B & 23 & 0.888 & 0.852 & 0.902 & 0.855 & 0.874 & 0.838$^*$ & 0.891 & 0.851$^*$ \\
LLaVA-1.5-7B & 14 & 0.649 & 0.754 & 0.667 & 0.754 & 0.340$^*$ & 0.516$^*$ & 0.649 & 0.754 \\
\bottomrule
\end{tabular}\end{adjustbox}
\caption{\textbf{Layer choice.} Held-out pointing and AUROC of \method with its label-free prior, of the single layer of the same per-layer map $F_l$ chosen by labeled pointing on the calibration split (``tuned layer''), and of the prior estimated from visual mass alone or from bits without the gray-image correction. $^*$Significantly different from \method (paired two-sided Wilcoxon, $p<0.05$).}
\label{tab:s_layer_choice}
\end{table*}

\paragraph{Layer choice.}
\Cref{tab:s_layer_choice} compares the label-free prior with the best single layer of our own map chosen with all labels of the calibration split, and with priors from visual mass alone or from bits without the gray-image correction. The prior is within 0.02 of the label-tuned layer in mean pointing and AUROC on both datasets. Visual mass ranks layer 0 highest in five models; uncorrected bits rank layer 0 highest for InternVL3-1B, where the prior then fails.

\begin{table*}[!t]
\centering
\footnotesize
\setlength{\tabcolsep}{3pt}
\begin{adjustbox}{max width=\textwidth}\begin{tabular}{@{}lcccc@{}}
\toprule
 & \multicolumn{2}{c}{COCO} & \multicolumn{2}{c}{RefCOCO} \\
\cmidrule(lr){2-3}\cmidrule(l){4-5}
Variant & Pt. & AUROC & Pt. & AUROC \\
\midrule
ViCrop~\cite{zhang2025vicrop}, label-tuned layer$^\dagger$ (reference) & 0.731 & 0.815 & 0.657 & 0.746 \\
(a) ViCrop's ratio $+$ our layer prior & 0.721 & 0.818 & 0.665 & 0.737 \\
(b) \method (per-head log-ratio, bit-weighted heads, prior) & 0.723 & 0.843 & 0.674 & 0.755 \\
(c) \method with uniform head weights & 0.703 & 0.831 & 0.654 & 0.746 \\
(d) \method, prior without gray correction & 0.669 & 0.799 & 0.589 & 0.705 \\
(e) \method, no prior (uniform layers) & 0.655 & 0.798 & 0.577 & 0.709 \\
\bottomrule
\end{tabular}\end{adjustbox}
\caption{\textbf{Isolating the two changes.} Held-out pointing and AUROC, means over six VLMs; (a)--(e) share the prior's calibration questions. $^\dagger$Layer tuned with labels.}
\label{tab:s_components}
\end{table*}

\paragraph{Components relative to ViCrop.}
\Cref{tab:s_components} isolates layer and head aggregation. Replacing ViCrop's tuned layer with our prior preserves mean pointing. Taking the log-ratio before head aggregation and weighting heads by bits adds 0.018--0.025 AUROC, with little change in pointing. Uniform head weights cost about 0.02 in pointing. Omitting the gray correction or the prior costs 0.05--0.10 in pointing.

\begin{table*}[!t]
\centering
\footnotesize
\setlength{\tabcolsep}{2.6pt}
\begin{adjustbox}{max width=\textwidth}\begin{tabular}{@{}lcccccc@{}}
\toprule
 & \multicolumn{2}{c}{COCO} & \multicolumn{2}{c}{RefCOCO} & \multicolumn{2}{c}{V$^*$} \\
\cmidrule(lr){2-3}\cmidrule(lr){4-5}\cmidrule(lr){6-7}
Score and reading & Pt. & AUROC & Pt. & AUROC & Pt. & AUROC \\
\midrule
\multicolumn{7}{@{}l}{\emph{$P\log(P/Q)$, heads averaged (PriorTR's score~\cite{chen2026priortr})}} \\
\quad layer 2 (PriorTR's layer) & 0.109 & 0.494 & 0.151 & 0.495 & 0.002 & 0.509 \\
\quad layer tuned with labels$^\dagger$ & 0.603 & 0.789 & 0.605 & 0.721 & 0.229 & 0.759 \\
\quad our layer prior & 0.578 & 0.794 & 0.538 & 0.717 & 0.207 & 0.757 \\
\multicolumn{7}{@{}l}{\emph{$\log(P/Q)$, heads averaged}} \\
\quad layer 2 & 0.157 & 0.498 & 0.216 & 0.494 & 0.013 & 0.558 \\
\quad our layer prior & 0.709 & 0.819 & 0.671 & 0.739 & 0.227 & 0.748 \\
\multicolumn{7}{@{}l}{\emph{per head, bit-weighted heads, our layer prior}} \\
\quad $P\log(P/Q)$ & 0.587 & 0.804 & 0.552 & 0.721 & \textbf{0.237} & \textbf{0.793} \\
\quad $\log(P/Q)$ = \textbf{\method} & \textbf{0.723} & \textbf{0.843} & \textbf{0.674} & \textbf{0.755} & 0.210 & 0.789 \\
\bottomrule
\end{tabular}\end{adjustbox}
\caption{\textbf{From PriorTR's score to \method.} Held-out pointing and AUROC, means over six VLMs (V$^*$: five); $P$, $Q$: routing of the answer-start query under the question and the generic prompt. Bold: best per column. $^\dagger$Layer tuned with labels on the odd ids (V$^*$: COCO).}
\label{tab:s_decomp}
\end{table*}
\begin{table*}[!t]
\centering
\footnotesize
\setlength{\tabcolsep}{4pt}
\begin{adjustbox}{max width=\textwidth}\begin{tabular}{@{}lcccccc@{}}
\toprule
 & \multicolumn{2}{c}{COCO} & \multicolumn{2}{c}{RefCOCO} & \multicolumn{2}{c}{V$^*$} \\
\cmidrule(lr){2-3}\cmidrule(lr){4-5}\cmidrule(lr){6-7}
Step & Pt. & AUROC & Pt. & AUROC & Pt. & AUROC \\
\midrule
layer 2 $\to$ our prior & $+0.469$ (6/0) & $+0.300$ (6/0) & $+0.387$ (6/0) & $+0.223$ (6/0) & $+0.205$ (4/0) & $+0.248$ (5/0) \\
$P\log(P/Q)\to\log(P/Q)$ & $+0.131$ (5/0) & $+0.026$ (5/0) & $+0.133$ (5/0) & $+0.022$ (4/0) & $+0.020$ (1/0) & $-0.009$ (1/3) \\
heads averaged $\to$ per head, bit weights & $+0.014$ (1/0) & $+0.024$ (5/0) & $+0.002$ (0/0) & $+0.015$ (5/0) & $-0.017$ (0/0) & $+0.041$ (3/0) \\
labeled layer, $P\log(P/Q)$ $\to$ \method & $+0.120$ (4/0) & $+0.054$ (6/0) & $+0.068$ (4/0) & $+0.033$ (5/0) & $-0.019$ (0/1) & $+0.030$ (3/1) \\
\bottomrule
\end{tabular}\end{adjustbox}
\caption{\textbf{From PriorTR's score to \method, step by step}: mean paired change of pointing and AUROC, pooled over models, and in parentheses the number of models in which the change is significantly positive / negative (Wilcoxon, $p<0.05$). The first three rows change one thing at a time; the last compares PriorTR's score at its label-tuned layer with \method.}
\label{tab:s_decomp_steps}
\end{table*}

\paragraph{From PriorTR's score to \method.}
PriorTR uses $P\log(P/Q)$ at layer 2 for token pruning, following FastV~\cite{chen2024fastv}, and finds this score better than the log-ratio for that task. Here we hold the question and reference rows fixed at the answer-start position to separate changes in score, heads, and layers (\cref{tab:s_decomp,tab:s_decomp_steps}).

Replacing layer 2 with our layer prior produces the largest pointing gains, $+0.47$ on COCO and $+0.39$ on RefCOCO. Dropping the factor $P$ adds $+0.13$ on each benchmark. Taking the log-ratio per head and then applying bit weights increases AUROC by 0.024, 0.015, and 0.041 on COCO, RefCOCO, and V$^*$, with little change in pointing.

Attention weighting can help for the much smaller targets in V$^*$. With the same head weights and layer prior, retaining $P$ raises V$^*$ pointing from 0.210 to 0.237 (paired difference 0.026, 95\% CI $[0.003,0.049]$). The corresponding AUROCs are 0.789 and 0.793. If heads are averaged before computing the score, retaining $P$ lowers V$^*$ pointing from 0.227 to 0.207.

\begin{table*}[!t]
\centering
\footnotesize
\setlength{\tabcolsep}{4pt}
\begin{adjustbox}{max width=\textwidth}\begin{tabular}{@{}lcccccc@{}}
\toprule
 & \multicolumn{2}{c}{COCO} & \multicolumn{2}{c}{RefCOCO} & \multicolumn{2}{c}{V$^*$} \\
\cmidrule(lr){2-3}\cmidrule(lr){4-5}\cmidrule(lr){6-7}
Map & Pt. & AUROC & Pt. & AUROC & Pt. & AUROC \\
\midrule
\multicolumn{7}{@{}l}{\emph{$P\log(P/Q)$, heads averaged (PriorTR's score~\cite{chen2026priortr})}} \\
\quad PriorTR's readout, layer 2 (as published) & 0.117 & 0.522 & 0.150 & 0.496 & 0.017 & 0.544 \\
\quad PriorTR's readout, our prior (\cref{tab:main}) & 0.549 & 0.718 & 0.606 & 0.683 & 0.348 & 0.799 \\
\quad instruction rows, generic-prompt reference, our prior & 0.531 & 0.731 & 0.596 & 0.702 & \textbf{0.360} & 0.813 \\
\quad answer-start row, separator reference, our prior & 0.565 & 0.787 & 0.505 & 0.687 & 0.200 & 0.750 \\
\quad our rows, our prior (\cref{tab:s_decomp}) & 0.578 & 0.794 & 0.538 & 0.717 & 0.207 & 0.757 \\
\multicolumn{7}{@{}l}{\emph{$P\log(P/Q)$ per head, bit-weighted heads, our prior}} \\
\quad PriorTR's readout & 0.606 & 0.742 & 0.646 & 0.701 & 0.327 & \textbf{0.815} \\
\quad our rows (\cref{tab:s_decomp}) & 0.587 & 0.804 & 0.552 & 0.721 & 0.237 & 0.793 \\
\multicolumn{7}{@{}l}{\emph{$\log(P/Q)$, our prior}} \\
\quad PriorTR's readout, heads averaged & 0.606 & 0.738 & 0.650 & 0.699 & 0.263 & 0.774 \\
\quad PriorTR's readout, per head, bit-weighted heads & 0.623 & 0.760 & 0.630 & 0.714 & 0.201 & 0.776 \\
\quad our rows, per head, bit-weighted heads = \textbf{\method} & \textbf{0.723} & \textbf{0.843} & \textbf{0.674} & \textbf{0.755} & 0.210 & 0.789 \\
\multicolumn{7}{@{}l}{\emph{CARVE's map~\cite{ge2025carve}}} \\
\quad answer-start step, depth 0.70--0.90 (re-impl., \cref{tab:main}) & 0.505 & 0.722 & 0.462 & 0.630 & 0.186 & 0.760 \\
\quad answer-start step, layers 20--25 & 0.520 & 0.737 & 0.487 & 0.640 & 0.192 & 0.766 \\
\quad all decoding steps, layers 20--25 (CARVE's readout) & 0.454 & 0.717 & 0.518 & 0.631 & 0.110 & 0.731 \\
\bottomrule
\end{tabular}\end{adjustbox}
\caption{\textbf{PriorTR and CARVE with their original attention readouts.} Held-out pointing and AUROC, means over six VLMs (V$^*$: five). PriorTR's readout: $P$ from the rows of all instruction tokens, $Q$ from the separator row after the image; our rows: the answer-start query under the question ($P$) and the generic prompt ($Q$). CARVE's layers 20--25 (layers 17--21 for InternVL3-1B, which has 24). Bold: best per column.}
\label{tab:s_own}
\end{table*}

\begin{table*}[!t]
\centering
\footnotesize
\setlength{\tabcolsep}{4pt}
\begin{adjustbox}{max width=\textwidth}\begin{tabular}{@{}lcccccc@{}}
\toprule
 & \multicolumn{2}{c}{COCO} & \multicolumn{2}{c}{RefCOCO} & \multicolumn{2}{c}{V$^*$} \\
\cmidrule(lr){2-3}\cmidrule(lr){4-5}\cmidrule(lr){6-7}
Model & Pt. & AUROC & Pt. & AUROC & Pt. & AUROC \\
\midrule
SmolVLM-256M & 0.504 / 0.537 & 0.673 / 0.762$^*$ & 0.421 / 0.460 & 0.599 / 0.638$^*$ & -- & -- \\
SmolVLM-500M & 0.656 / 0.718$^*$ & 0.723 / 0.854$^*$ & 0.530 / 0.533 & 0.628 / 0.680$^*$ & 0.141 / 0.042$^*$ & 0.621 / 0.629 \\
InternVL3-1B & 0.448 / 0.463 & 0.617 / 0.755$^*$ & 0.519 / 0.726$^*$ & 0.648 / 0.793$^*$ & 0.277 / 0.042$^*$ & 0.734 / 0.724$^*$ \\
Qwen3-VL-2B & 0.494 / 0.819$^*$ & 0.719 / 0.898$^*$ & 0.758 / 0.786 & 0.728 / 0.812$^*$ & 0.304 / 0.387$^*$ & 0.858 / 0.910$^*$ \\
Qwen3-VL-8B & 0.646 / 0.891$^*$ & 0.784 / 0.913$^*$ & 0.814 / 0.888$^*$ & 0.774 / 0.852$^*$ & 0.508 / 0.372$^*$ & 0.881 / 0.900 \\
LLaVA-1.5-7B & 0.547 / 0.908$^*$ & 0.794 / 0.877$^*$ & 0.593 / 0.649 & 0.720 / 0.754$^*$ & 0.508 / 0.209$^*$ & 0.901 / 0.783$^*$ \\
\bottomrule
\end{tabular}\end{adjustbox}
\caption{\textbf{PriorTR's own readout with our prior, per model}: PriorTR / \method, held-out pointing and AUROC. $^*$: $p<0.05$ (Wilcoxon, uncorrected).}
\label{tab:s_own_models}
\end{table*}

\begin{table*}[!t]
\centering
\footnotesize
\setlength{\tabcolsep}{4pt}
\begin{adjustbox}{max width=\textwidth}\begin{tabular}{@{}lccccccc@{}}
\toprule
 & SmolVLM-500M & InternVL3-1B & Qwen3-VL-2B & Qwen3-VL-8B & LLaVA-1.5-7B & Mean & Mean (adj.) \\
\midrule
No zoom & 29.8 & 41.9 & 47.6 & 47.1 & 47.1 & 42.7 & 42.8 \\
CARVE's pipeline~\cite{ge2025carve}, $(p,K)=(0.3,1)$ (CARVE's unified setting) & 34.6 & 44.5 & 48.2 & 53.4 & 42.4 & 44.6 & 45.3 \\
CARVE's pipeline, $(0.2,1)$ & 30.9 & 40.8 & 49.2 & 56.0 & 39.8 & 43.4 & 44.5 \\
CARVE's pipeline, $(0.4,1)$ & 31.9 & 43.5 & 48.7 & 53.9 & 45.0 & 44.6 & 45.2 \\
CARVE's pipeline, $(0.4,2)$ & 29.3 & 41.9 & 46.6 & 53.4 & 45.0 & 43.2 & 43.4 \\
CARVE's pipeline, $(0.2,3)$ & 33.5 & 45.0 & 46.1 & 55.5 & 43.5 & 44.7 & 45.0 \\
CARVE's pipeline, $(0.6,2)$ & 33.0 & 39.8 & 45.5 & 47.6 & 48.2 & 42.8 & 42.9 \\
CARVE's pipeline, $(0.3,1)$, original + refined$^\ddagger$ & 33.0 & 44.5 & 27.2 & 49.7 & -- & 38.6$^\ddagger$ & 46.0$^\ddagger$ \\
CARVE's pipeline, $(0.4,2)$, original + refined$^\ddagger$ & 31.9 & 42.4 & 23.6 & 42.9 & -- & 35.2$^\ddagger$ & 42.6$^\ddagger$ \\
CARVE's map, our crop rule & 33.0 & 42.9 & 42.1 & 56.3 & 47.4 & 44.3 & 48.1 \\
\method, our crop rule & 34.8 & 41.6 & 37.7 & 58.6 & 53.7 & 45.3 & 49.5 \\
\midrule
CARVE crop area, $(0.3,1)$ (\% of image) & 30 & 24 & 29 & 31 & 27 & 28 & -- \\
CARVE crop contains the target, $(0.3,1)$ (\%) & 39 & 38 & 56 & 67 & 22 & 45 & -- \\
CARVE crop area, $(0.4,2)$ (\% of image) & 87 & 84 & 88 & 89 & 95 & 88 & -- \\
CARVE crop contains the target, $(0.4,2)$ (\%) & 93 & 90 & 95 & 98 & 98 & 95 & -- \\
\bottomrule
\end{tabular}\end{adjustbox}
\caption{\textbf{V$^*$ zoom-in with CARVE's full pipeline} (accuracy in \%, 191 questions). CARVE's pipeline: its own map, the top-$p$ share of tokens, the $K$ regions with the largest summed score, cropped and masked, answered from the refined image alone; $(0.3,1)$ is the unified setting of CARVE's third version. $^\ddagger$With the original image as well (mean over four models, whose no-zoom means are 41.6 and 41.7 adj.; LLaVA-1.5 takes one image). Our crop rule: mean of the two windows (\cref{tab:s_vstar_zoom}). Generated answers; mean (adj.): answers without an option letter counted at chance.}
\label{tab:s_carve_zoom}
\end{table*}

\paragraph{PriorTR and CARVE with their original attention readouts.}
\label{sec:s_own}
These readouts need attention rows that the main runs do not store, so we ran the 393 COCO, 285 RefCOCO, and 191 V$^*$ questions again with the same models, inputs, and prompts. For PriorTR we recorded instruction and separator rows in each layer. For CARVE we recorded every decoding step up to four answer tokens under the question and its generic prompt. For RefCOCO we also computed Chefer relevance for each answer token. The CARVE answer-start maps from these runs correlate with the stored maps at $r\ge0.9994$ for every question. First-token Chefer maps correlate at $r\ge0.99997$ and produce identical pointing decisions.

\emph{PriorTR.} The original layer-2 readout gives pointing of 0.117, 0.150, and 0.017 on COCO, RefCOCO, and V$^*$, respectively. Applying our layer prior raises these scores to 0.549, 0.606, and 0.348 (\cref{tab:s_own}). Compared with \method, pointing changes by $-0.17$ (95\% CI $[-0.20,-0.15]$), $-0.07$ ($[-0.09,-0.04]$), and $+0.14$ ($[0.10,0.17]$), respectively. V$^*$ AUROC changes by $+0.010$ ($[-0.002,0.022]$). \Cref{tab:s_own_models} gives results by model.

With the generic-prompt reference held fixed, replacing the answer-start row by instruction rows raises V$^*$ pointing by 0.15 ($[0.12,0.18]$) and RefCOCO pointing by 0.06, while lowering COCO pointing by 0.05. Replacing the generic reference with the separator row under the instruction readout changes V$^*$ pointing by $-0.01$, with an interval including zero. Applying a plain log-ratio to PriorTR's head-averaged readout reduces V$^*$ pointing to 0.263, still $0.05$ above \method ($[0.02,0.08]$). Reading that log-ratio per head and applying our weights reduces pointing to 0.201. The V$^*$ advantage in this comparison is strongest with instruction rows and attention weighting together. For LLaVA-1.5, the separator after the image is a whitespace token that precedes the question.

\emph{CARVE's map.} At the answer-start step, using the published layer range improves pointing by up to 0.025 and AUROC by up to 0.015 over the shared relative-depth band. Aggregating all decoding steps then yields pointing of 0.454 on COCO, 0.518 on RefCOCO, and 0.110 on V$^*$ (\cref{tab:s_own}). This comparison uses max-normalized rows. With raw probabilities, the stabilizer $\lambda=0.05$ dominates most denominator entries, and pointing falls to 0.349, 0.381, and 0.018. Compared with the max-normalized all-step readout, \method gains 0.27, 0.16, and 0.10 in pointing and 0.13, 0.12, and 0.06 in AUROC.

\emph{CARVE's zoom-in pipeline.} We take the top $p$ fraction of map tokens, group them into 8-connected regions, and retain the $K$ regions with highest summed score. We crop their joint bounding box, fill other cells with gray, and answer from this single refined image at the model's input size. We use CARVE's $\lambda=0.05$ and evaluate six $(p,K)$ settings, including $(0.3,1)$, the unified setting of the CARVE paper. The single-region crop at $(0.3,1)$ covers 24--31\% of the image and contains the target in 22--67\% of questions. The \method half-side window covers 25\% and contains the target in 29--70\%. With $K>1$, the selected regions contain 9--58\% of tokens, while their bounding boxes cover 64--100\% of the image.

These runs used generated answers (\cref{sec:s_impl}). In question-clustered comparisons over five models, CARVE's gains over no zoom range from 0.1 to 2.0 points, with intervals including zero (\cref{tab:s_carve_zoom}). The primary \method crops exceed CARVE by 0.6--2.5 points as scored, with intervals including zero. Crediting answers without a letter at chance raises the gaps to 4.2--6.6 points ($p<0.008$). CARVE's single-image inputs produce at most 12\% of such responses. For the four models that accept both images, adding the original image at $(0.3,1)$ gives $-3.0$ points versus no zoom as scored and $+4.3$ with chance credit ($p=0.007$). Under chance credit, \method is 2.4 points ahead of this two-image variant ($p=0.055$).

\begin{table*}[!t]
\centering
\footnotesize
\setlength{\tabcolsep}{4pt}
\begin{adjustbox}{max width=\textwidth}\begin{tabular}{@{}lcccccc@{}}
\toprule
 & & & \multicolumn{2}{c}{\method $-$ map, Pt.} & \multicolumn{2}{c}{\method $-$ map, AUROC} \\
\cmidrule(lr){4-5}\cmidrule(lr){6-7}
Chefer relevance~\cite{chefer2021generic} explains & Pt. & AUROC & mean [95\% CI] & models $+/-$ & mean & models $+/-$ \\
\midrule
first answer token (as in the paper) & 0.211 & 0.576 & $+0.463$ $[+0.431,+0.495]$ & 6/0 & $+0.179$ & 6/0 \\
first answer token that is not a function word & 0.268 & 0.575 & $+0.406$ $[+0.374,+0.437]$ & 6/0 & $+0.180$ & 6/0 \\
all answer tokens (normalized sum) & 0.235 & 0.580 & $+0.439$ $[+0.408,+0.470]$ & 6/0 & $+0.175$ & 6/0 \\
best answer token, chosen with labels (oracle) & 0.412 & 0.614 & $+0.262$ $[+0.230,+0.293]$ & 4/0 & $+0.141$ & 6/0 \\
\midrule
\method & 0.674 & 0.755 & & & & \\
\bottomrule
\end{tabular}\end{adjustbox}
\caption{\textbf{Chefer relevance explaining other answer tokens} (RefCOCO; means over six VLMs). Models $+/-$: VLMs in which \method is significantly better / worse (Wilcoxon, $p<0.05$). The oracle picks, per expression, the answer token with the best score, separately for each metric.}
\label{tab:s_chefer_targets}
\end{table*}

\begin{table*}[!t]
\centering
\footnotesize
\setlength{\tabcolsep}{4pt}
\begin{adjustbox}{max width=\textwidth}\begin{tabular}{@{}lcccccc@{}}
\toprule
First answer token & $n$ & \method & Chefer~\cite{chefer2021generic} & Grad-CAM~\cite{selvaraju2017gradcam} & TAM~\cite{li2025tam} & \method $-$ Chefer, Pt. [95\% CI] \\
\midrule
function word (``The'', ``On'') & 805 & 0.680 / 0.763 & 0.169 / 0.569 & 0.160 / 0.490 & 0.248 / 0.500 & $+0.511$ $[+0.467,+0.553]$ \\
spatial word (``left'', ``right'') & 455 & 0.703 / 0.780 & 0.237 / 0.603 & 0.123 / 0.480 & 0.288 / 0.528 & $+0.466$ $[+0.411,+0.521]$ \\
other word & 450 & 0.633 / 0.714 & 0.258 / 0.559 & 0.184 / 0.491 & 0.202 / 0.511 & $+0.376$ $[+0.316,+0.434]$ \\
not a function word & 905 & 0.669 / 0.747 & 0.248 / 0.581 & 0.154 / 0.486 & 0.245 / 0.519 & $+0.421$ $[+0.380,+0.460]$ \\
all & 1710 & 0.674 / 0.755 & 0.211 / 0.576 & 0.157 / 0.487 & 0.247 / 0.510 & $+0.463$ $[+0.431,+0.495]$ \\
\bottomrule
\end{tabular}\end{adjustbox}
\caption{\textbf{RefCOCO split by the model's first answer token}, the target of Chefer relevance, Grad-CAM and TAM. Pointing / AUROC pooled over six VLMs ($n$: model--expression pairs); last column: mean paired difference in pointing with a bootstrap over expressions.}
\label{tab:s_target}
\end{table*}

\paragraph{Sensitivity to the target answer token.}
Chefer relevance, Grad-CAM, and TAM use the model's first greedy answer token. This token is the yes/no response or a color token on COCO. On RefCOCO, it is a function word in 47\% of model--expression pairs, a spatial word in 27\%, and another word in 26\%. Function words include articles, pronouns, auxiliaries, and prepositions. Spatial words include ``left'', ``right'', and ``middle''. The first token is ``The'' for all InternVL3-1B RefCOCO responses.

Chefer pointing improves from 0.169 to 0.248 on the subset without an initial function word (\cref{tab:s_target}). In that subset, \method is 0.42 higher in pointing (95\% CI $[0.38,0.46]$ over expressions) and 0.17 higher in AUROC. We also compute Chefer relevance for other target tokens (\cref{tab:s_chefer_targets}). Targeting the first non-function token raises pointing from 0.211 to 0.268. Targeting all generated tokens jointly gives 0.235. A labeled oracle that picks the best token per example attains 0.412 pointing and 0.614 AUROC. The oracle still trails \method by 0.26 in pointing ($[0.23,0.29]$) and 0.14 in AUROC.

\begin{table*}[!t]
\centering
\footnotesize
\setlength{\tabcolsep}{3pt}
\begin{adjustbox}{max width=\textwidth}\begin{tabular}{@{}llcccccccc@{}}
\toprule
 & & \multicolumn{5}{c}{$\Delta\log p$ of the first answer token} & & & \\
\cmidrule(lr){3-7}
Model & $K$ heads & bits & mass & KL & entropy & random & bits $-$ random [95\% CI] & $p$ & token changed (\%): bits / random \\
\midrule
InternVL3-1B & 7 (2\%) & $-$0.117 & $-$0.060 & $-$0.065 & $+$0.004 & $-$0.003 & $-$0.114 [$-$0.185,$-$0.058] & $<$0.001 & 8 / 2 \\
InternVL3-1B & 17 (5\%) & $-$0.227 & $-$0.395 & $-$0.369 & $-$0.012 & $+$0.003 & $-$0.230 [$-$0.334,$-$0.144] & $<$0.001 & 11 / 2 \\
InternVL3-1B & 34 (10\%) & $-$0.440 & $-$0.537 & $-$0.683 & $-$0.028 & $-$0.040 & $-$0.400 [$-$0.570,$-$0.263] & $<$0.001 & 15 / 4 \\
LLaVA-1.5-7B & 20 (2\%) & $-$0.030 & $-$0.045 & $-$0.041 & $-$0.004 & $+$0.000 & $-$0.030 [$-$0.043,$-$0.019] & $<$0.001 & 1 / 0 \\
LLaVA-1.5-7B & 51 (5\%) & $-$0.078 & $-$0.133 & $-$0.110 & $-$0.006 & $-$0.005 & $-$0.073 [$-$0.103,$-$0.047] & $<$0.001 & 2 / 0 \\
LLaVA-1.5-7B & 102 (10\%) & $-$0.144 & $-$0.155 & $-$0.178 & $-$0.012 & $-$0.008 & $-$0.136 [$-$0.184,$-$0.091] & $<$0.001 & 4 / 1 \\
\bottomrule
\end{tabular}\end{adjustbox}
\caption{\textbf{Blocking the visual input of selected heads at the answer-start position} (120 held-out COCO questions per model). Mean change of the log-probability of the model's own first answer token when the top $K$ heads of each ranking are blocked; random: mean over five seeds with the same $K$. bits $-$ random: paired difference with a 95\% bootstrap interval and a two-sided Wilcoxon $p$. Last column: share of questions whose greedy first token changes.}
\label{tab:s_intervene}
\end{table*}

\paragraph{Intervening on selected heads.}
For 120 held-out COCO questions per model (60 presence and 60 color), we rank heads by their map weights $\pi_lw_{lh}/\sum_{h'}w_{lh'}$, visual mass $m^{xt}$, uncorrected where-to-look bits $m^{xt}\mathrm{KL}(P^{xt}\Vert P^{xg})$, routing entropy (lowest first), or random order (five seeds). At the answer-start position we set post-softmax visual-key attention to zero in the top 2\%, 5\%, or 10\% of heads and renormalize the text-key weights. We record the change in log-probability of the original first answer token and whether the greedy token changes.

Blocking 5\% of the heads ranked highest by \method lowers the first-token log-probability by 0.07--0.65 nats across models, compared with at most 0.03 for random heads. The greedy token changes in 2--22\% of questions, compared with 0.3--2.5\% for random heads. Across 18 model--fraction comparisons, blocking \method heads has a larger effect than random heads in 17 at uncorrected $p<0.05$ (\cref{tab:s_intervene}). Visual-mass rankings yield a larger drop than \method in 12 settings, and uncorrected bits in eight. Thus rankings that best localize and rankings with the largest effect under this intervention can differ.

Blocking all heads in the three layers with highest prior weights lowers the first-token log-probability more than blocking three random layers in five of six models. Effects are larger on less confident answers. When the initial probability is below 0.9 (25--43 questions per model), blocking the top 5\% of \method heads lowers log-probability by 0.15--1.05 nats.

\begin{table*}[!t]
\centering
\footnotesize
\setlength{\tabcolsep}{2.4pt}
\begin{adjustbox}{max width=\textwidth}\begin{tabular}{@{}lcccccc@{}}
\toprule
 & \multicolumn{3}{c}{Primary rule} & \multicolumn{3}{c}{Bits window (exploratory)} \\
\cmidrule(lr){2-4}\cmidrule(l){5-7}
\method $-$ & diff. & 95\% CI & $p$ & diff. & 95\% CI & $p$ \\
\midrule
ViCrop~\cite{zhang2025vicrop}, tuned layer$^\dagger$ & ${+}0.6$ & $[-0.2,{+}1.4]$ & 0.143 & ${+}1.7$ & $[{+}0.8,{+}2.6]$ & $<$0.001 \\
ViCrop, all layers & ${+}0.0$ & $[-0.8,{+}0.9]$ & 0.959 & ${+}1.1$ & $[{+}0.2,{+}2.0]$ & 0.018 \\
CARVE~\cite{ge2025carve} (re-impl.) & $-1.8$ & $[-2.7,-1.0]$ & $<$0.001 & $-0.7$ & $[-1.5,{+}0.1]$ & 0.080 \\
Chefer relevance~\cite{chefer2021generic} & $-1.6$ & $[-2.7,-0.6]$ & 0.003 & $-0.5$ & $[-1.4,{+}0.4]$ & 0.263 \\
Attention (mean) & $-0.9$ & $[-1.9,{+}0.1]$ & 0.080 & ${+}0.2$ & $[-0.7,{+}1.1]$ & 0.686 \\
Center crop & ${+}8.3$ & $[{+}6.7,{+}10.0]$ & $<$0.001 & ${+}9.4$ & $[{+}7.8,{+}11.1]$ & $<$0.001 \\
No zoom & ${+}6.0$ & $[{+}3.9,{+}8.1]$ & $<$0.001 & ${+}7.1$ & $[{+}5.1,{+}9.1]$ & $<$0.001 \\
\method, primary rule & \multicolumn{3}{c}{--} & ${+}1.1$ & $[{+}0.3,{+}1.9]$ & 0.009 \\
\bottomrule
\end{tabular}\end{adjustbox}
\caption{\textbf{TextVQA zoom-in, pooled with question clusters} (500 questions). Mean paired accuracy differences in points, averaged within each question across the two windows and available models, with 95\% bootstrap intervals and sign-flip $p$-values (20{,}000 draws each). The no-zoom comparison uses 1{,}990 available model--question pairs. $^\dagger$Layer tuned with COCO labels.}
\label{tab:s_zoom_tvqa_q}
\end{table*}
\begin{table*}[!t]
\centering
\footnotesize
\setlength{\tabcolsep}{2.6pt}
\begin{adjustbox}{max width=\textwidth}\begin{tabular}{@{}lcccccccccccc@{}}
\toprule
 & \multicolumn{6}{c}{Primary rule} & \multicolumn{6}{c}{Bits window} \\
\cmidrule(lr){2-7}\cmidrule(l){8-13}
 & \multicolumn{3}{c}{as scored} & \multicolumn{3}{c}{no letter at chance} & \multicolumn{3}{c}{as scored} & \multicolumn{3}{c}{no letter at chance} \\
\cmidrule(lr){2-4}\cmidrule(lr){5-7}\cmidrule(lr){8-10}\cmidrule(l){11-13}
\method $-$ & diff. & 95\% CI & $p$ & diff. & 95\% CI & $p$ & diff. & 95\% CI & $p$ & diff. & 95\% CI & $p$ \\
\midrule
ViCrop~\cite{zhang2025vicrop}, tuned layer$^\dagger$ & $-0.7$ & $[-2.0,{+}0.6]$ & 0.356 & $-0.3$ & $[-1.6,{+}0.9]$ & 0.653 & ${+}1.0$ & $[-0.5,{+}2.6]$ & 0.220 & ${+}0.9$ & $[-0.5,{+}2.3]$ & 0.240 \\
ViCrop, all layers & ${+}2.1$ & $[{+}0.6,{+}3.6]$ & 0.009 & ${+}2.4$ & $[{+}1.0,{+}3.8]$ & 0.001 & ${+}3.8$ & $[{+}2.1,{+}5.6]$ & $<$0.001 & ${+}3.6$ & $[{+}1.9,{+}5.3]$ & $<$0.001 \\
CARVE~\cite{ge2025carve} (re-impl.) & ${+}0.9$ & $[-1.0,{+}2.8]$ & 0.364 & ${+}1.4$ & $[-0.4,{+}3.2]$ & 0.134 & ${+}2.7$ & $[{+}1.2,{+}4.2]$ & $<$0.001 & ${+}2.6$ & $[{+}1.2,{+}4.0]$ & $<$0.001 \\
Chefer relevance~\cite{chefer2021generic} & ${+}4.9$ & $[{+}2.6,{+}7.1]$ & $<$0.001 & ${+}4.0$ & $[{+}1.9,{+}6.1]$ & $<$0.001 & ${+}6.6$ & $[{+}4.7,{+}8.5]$ & $<$0.001 & ${+}5.2$ & $[{+}3.5,{+}6.8]$ & $<$0.001 \\
Attention (mean) & ${+}6.8$ & $[{+}4.3,{+}9.2]$ & $<$0.001 & ${+}5.7$ & $[{+}3.5,{+}8.0]$ & $<$0.001 & ${+}8.5$ & $[{+}6.5,{+}10.5]$ & $<$0.001 & ${+}6.9$ & $[{+}5.1,{+}8.8]$ & $<$0.001 \\
\bottomrule
\end{tabular}\end{adjustbox}
\caption{\textbf{V$^*$ zoom-in with generated answers, pooled with question clusters} (191 questions, averaging five models and two windows): as scored (an answer without an option letter is wrong) and with such answers counted at chance ($1/k$). Conventions as in \cref{tab:s_zoom_tvqa_q}. $^\dagger$Layer tuned with COCO labels.}
\label{tab:s_zoom_vstar_q}
\end{table*}
\begin{table}[!t]
\centering
\footnotesize
\setlength{\tabcolsep}{2pt}
\begin{adjustbox}{max width=\linewidth}\begin{tabular}{@{}lcccccc@{}}
\toprule
 & \multicolumn{3}{c}{Primary rule} & \multicolumn{3}{c}{Bits window} \\
\cmidrule(lr){2-4}\cmidrule(l){5-7}
\method $-$ & diff. & 95\% CI & $p$ & diff. & 95\% CI & $p$ \\
\midrule
ViCrop~\cite{zhang2025vicrop}, tuned layer$^\dagger$ & $-0.3$ & $[-1.7,{+}1.2]$ & 0.771 & ${+}0.9$ & $[-0.6,{+}2.4]$ & 0.253 \\
ViCrop, all layers & ${+}2.1$ & $[{+}0.5,{+}3.7]$ & 0.014 & ${+}3.3$ & $[{+}1.5,{+}5.2]$ & $<$0.001 \\
CARVE~\cite{ge2025carve} (re-impl.) & ${+}1.5$ & $[-0.3,{+}3.2]$ & 0.111 & ${+}2.7$ & $[{+}1.0,{+}4.2]$ & 0.002 \\
Chefer relevance~\cite{chefer2021generic} & ${+}3.5$ & $[{+}1.4,{+}5.6]$ & 0.002 & ${+}4.7$ & $[{+}2.9,{+}6.5]$ & $<$0.001 \\
Attention (mean) & ${+}5.4$ & $[{+}3.2,{+}7.6]$ & $<$0.001 & ${+}6.6$ & $[{+}4.7,{+}8.6]$ & $<$0.001 \\
Center crop & ${+}5.3$ & $[{+}2.8,{+}7.8]$ & $<$0.001 & ${+}6.5$ & $[{+}4.0,{+}9.1]$ & $<$0.001 \\
No zoom & ${+}6.0$ & $[{+}3.5,{+}8.5]$ & $<$0.001 & ${+}7.2$ & $[{+}4.7,{+}9.6]$ & $<$0.001 \\
\method, primary rule & \multicolumn{3}{c}{--} & ${+}1.2$ & $[-0.3,{+}2.7]$ & 0.139 \\
\bottomrule
\end{tabular}\end{adjustbox}
\caption{\textbf{V$^*$ zoom-in under forced choice, pooled with question clusters} (191 questions, five models, two windows). Conventions as in \cref{tab:s_zoom_tvqa_q}. $^\dagger$Layer tuned with COCO labels.}
\label{tab:s_vstar_fc_pooled}
\end{table}

\paragraph{Zoom-in with question clusters.}
On TextVQA the primary rule gains 6.0 accuracy points over no zoom, while the exploratory bits-window rule gains 7.1 (\cref{tab:s_zoom_tvqa_q}). The primary rule trails CARVE and Chefer relevance and has intervals spanning zero against mean attention and both ViCrop variants. Bits-window gains 1.1 points over the primary rule and 1.1--1.7 points over ViCrop. Its intervals versus CARVE, Chefer, and mean attention include zero. On V$^*$ under forced choice (\cref{tab:s_vstar_fc_pooled}), the primary rule gains 6.0 points over no zoom ($[3.5,8.5]$) and 5.4 over mean attention, leads all-layer ViCrop and Chefer relevance, and is within $[-1.7,1.2]$ of the label-tuned layer; its interval against CARVE's map with our crop rule includes zero ($+1.5$, $[-0.3,3.2]$). The bits window adds 1.2 points ($[-0.3,2.7]$) and leads CARVE by 2.7 ($[1.0,4.2]$). With generated answers (\cref{tab:s_zoom_vstar_q}), the primary rule's gain over no zoom is 2.6 points with an interval including zero, and 6.7 points when answers without a letter receive chance credit.

\begin{table*}[!t]
\centering
\footnotesize
\setlength{\tabcolsep}{3pt}
\begin{adjustbox}{max width=\textwidth}\begin{tabular}{@{}llccccccccccccccc@{}}
\toprule
 & & \multicolumn{3}{c}{Attention (mean)} & \multicolumn{3}{c}{CARVE~\cite{ge2025carve}} & \multicolumn{3}{c}{ViCrop~\cite{zhang2025vicrop}$^\dagger$} & \multicolumn{3}{c}{\method} & \multicolumn{3}{c}{+ bits window} \\
\cmidrule(lr){3-5}\cmidrule(lr){6-8}\cmidrule(lr){9-11}\cmidrule(lr){12-14}\cmidrule(lr){15-17}
Win. & Model & corner & in & acc. in / out & corner & in & acc. in / out & corner & in & acc. in / out & corner & in & acc. in / out & corner & in & acc. in / out \\
\midrule
1/2 & SmolVLM-500M & 100 & 30 & 33 / 37 & 60 & 37 & 39 / 33 & 11 & 27 & 37 / 39 & 11 & 29 & 36 / 40 & 46 & 30 & 37 / 38 \\
1/2 & InternVL3-1B & 100 & 30 & 51 / 48 & 49 & 36 & 54 / 37 & 7 & 42 & 52 / 41 & 9 & 31 & 48 / 40 & 88 & 34 & 51 / 49 \\
1/2 & Qwen3-VL-2B & 99 & 35 & 57 / 50 & 63 & 68 & 65 / 35 & 15 & 68 & 62 / 41 & 14 & 66 & 65 / 29 & 15 & 67 & 60 / 35 \\
1/2 & Qwen3-VL-8B & 100 & 7 & 79 / 47 & 62 & 55 & 76 / 42 & 18 & 69 & 72 / 33 & 13 & 70 & 73 / 38 & 8 & 66 & 69 / 38 \\
1/2 & LLaVA-1.5-7B & 26 & 23 & 53 / 40 & 31 & 35 & 59 / 41 & 36 & 51 & 60 / 43 & 32 & 64 & 54 / 44 & 26 & 58 & 57 / 45 \\
\midrule
1/3 & SmolVLM-500M & 100 & 16 & 39 / 37 & 43 & 29 & 42 / 35 & 2 & 13 & 40 / 36 & 3 & 19 & 47 / 32 & 44 & 21 & 45 / 35 \\
1/3 & InternVL3-1B & 100 & 7 & 57 / 47 & 37 & 15 & 68 / 38 & 1 & 18 & 62 / 39 & 4 & 10 & 63 / 38 & 83 & 9 & 72 / 46 \\
1/3 & Qwen3-VL-2B & 100 & 11 & 81 / 42 & 30 & 54 & 73 / 38 & 3 & 54 & 75 / 34 & 4 & 49 & 76 / 38 & 6 & 55 & 75 / 34 \\
1/3 & Qwen3-VL-8B & 98 & 2 & 75 / 47 & 35 & 47 & 82 / 42 & 5 & 59 & 83 / 35 & 4 & 54 & 86 / 40 & 5 & 59 & 84 / 38 \\
1/3 & LLaVA-1.5-7B & 19 & 6 & 83 / 37 & 15 & 15 & 68 / 44 & 16 & 32 & 65 / 50 & 10 & 46 & 67 / 48 & 14 & 45 & 67 / 49 \\
\bottomrule
\end{tabular}\end{adjustbox}
\caption{\textbf{Where the V$^*$ crops land.} Per model and window (1/2 or 1/3 of the side): share of crops (\%) that start at an image corner, share (\%) that contain every target box (``in''), and forced-choice accuracy (\%) on the questions whose crop contains the target / on the others. $^\dagger$Layer tuned with COCO labels.}
\label{tab:s_crops}
\end{table*}

\paragraph{Crop locations and target coverage.}
\Cref{tab:s_crops} reports, for every model and window on V$^*$, the share of crops that start at an image corner, the share that contain every target box, and the forced-choice accuracy on the questions whose crop contains the target and on the others. Mean attention's crops start at a corner for 98--100\% of the questions in four of the five models, CARVE's for 30--63\% (15--31\% on LLaVA-1.5-7B), and those of \method and the label-tuned layer for 1--18\% (LLaVA-1.5-7B, whose sink is weaker, 10--36\%). Except for SmolVLM-500M, which is near chance, questions whose crop contains the target are answered more often than the others in every model, window and map. The bits-window rule selects a corner window for 83--88\% of the questions on InternVL3-1B (44--46\% on SmolVLM-500M; 5--26\% elsewhere); its pooled gain over the primary rule in \cref{sec:zoom} comes from InternVL3-1B ($+7.3$ points, $[3.7,11.3]$), which answers 46--49\% of the questions correctly even when the crop misses the target (42\% with no crop), and disappears without this model ($-0.3$ points, 95\% CI $[-2.0,1.4]$, question clusters). On TextVQA, where the bits-window rule gains 0--3 points on every model, its crops start at a corner for 1--7\% of the questions, and mean attention's for 5--46\%, so there the sinks do not dominate the crop.

\paragraph{Calibration set size and domain.}
We re-estimate the prior from $n$ randomly sampled COCO calibration questions (200 draws per $n$) and from questions on other datasets (\cref{tab:s_prior_input}). Averaged over draws, ten questions yield held-out COCO pointing within 0.011 of using all 385 questions on every model. One hundred yield pointing within 0.005.

Using RefCOCO or TextVQA questions for calibration gives COCO pointing within 0.08 of the COCO-calibrated prior. A V$^*$-calibrated prior has weaker transfer on SmolVLM-500M (0.455 instead of 0.718) and InternVL3-1B (0.267 instead of 0.463). V$^*$ differs in target size and multiple-choice format, and this comparison varies both factors. On RefCOCO, calibrating on RefCOCO gives lower mean pointing than calibrating on COCO in all six models, with significant decreases on SmolVLM-500M and Qwen3-VL-2B.

\begin{table*}[!tbp]
\centering
\footnotesize
\setlength{\tabcolsep}{3pt}
\begin{adjustbox}{max width=\textwidth}\begin{tabular}{@{}lccccccccccc@{}}
\toprule
 & \multicolumn{6}{c}{COCO held-out; prior from $n$ unlabeled COCO questions} & \multicolumn{3}{c}{COCO held-out; prior from} & \multicolumn{2}{c}{RefCOCO held-out; prior from} \\
\cmidrule(lr){2-7}\cmidrule(lr){8-10}\cmidrule(lr){11-12}
Model & $n{=}1$ & 3 & 10 & 30 & 100 & all & RefCOCO & TextVQA & V$^*$ & COCO & RefCOCO \\
\midrule
SmolVLM-256M & 0.511 / 0.754 & 0.521 / 0.760 & 0.531 / 0.762 & 0.534 / 0.762 & 0.535 / 0.763 & 0.537 / 0.762 & 0.534 / 0.759 & -- & -- & 0.460 / 0.638 & 0.456 / 0.624$^*$ \\
SmolVLM-500M & 0.693 / 0.849 & 0.700 / 0.852 & 0.710 / 0.854 & 0.713 / 0.854 & 0.715 / 0.854 & 0.718 / 0.854 & 0.646 / 0.843 & 0.677 / 0.853 & 0.455 / 0.717 & 0.533 / 0.680 & 0.467$^*$ / 0.638$^*$ \\
InternVL3-1B & 0.430 / 0.723 & 0.462 / 0.747 & 0.468 / 0.752 & 0.469 / 0.755 & 0.467 / 0.755 & 0.463 / 0.755 & 0.455 / 0.753 & 0.491 / 0.760 & 0.267 / 0.638 & 0.726 / 0.793 & 0.702 / 0.793 \\
Qwen3-VL-2B & 0.789 / 0.886 & 0.796 / 0.890 & 0.810 / 0.895 & 0.815 / 0.897 & 0.818 / 0.898 & 0.819 / 0.898 & 0.753 / 0.861 & 0.776 / 0.866 & 0.796 / 0.891 & 0.786 / 0.812 & 0.716$^*$ / 0.775$^*$ \\
Qwen3-VL-8B & 0.856 / 0.906 & 0.872 / 0.908 & 0.882 / 0.911 & 0.883 / 0.912 & 0.887 / 0.913 & 0.891 / 0.913 & 0.896 / 0.897 & -- & 0.863 / 0.911 & 0.888 / 0.852 & 0.877 / 0.830$^*$ \\
LLaVA-1.5-7B & 0.848 / 0.853 & 0.872 / 0.862 & 0.898 / 0.873 & 0.904 / 0.876 & 0.906 / 0.877 & 0.908 / 0.877 & 0.860 / 0.845 & 0.896 / 0.864 & 0.896 / 0.878 & 0.649 / 0.754 & 0.635 / 0.745$^*$ \\
\bottomrule
\end{tabular}\end{adjustbox}
\caption{\textbf{Input of the prior.} Held-out pointing / token AUROC of \method with the prior estimated from $n$ random unlabeled COCO questions (mean over 200 draws, or all 385 calibration questions), from the unlabeled questions of RefCOCO, TextVQA or V$^*$ (``--'': model not run on that dataset), and, on RefCOCO, from COCO (the paper's prior) or from RefCOCO's own odd-ID expressions. $^*$Significantly different from the COCO-estimated prior (paired two-sided Wilcoxon, $p<0.05$).}
\label{tab:s_prior_input}
\end{table*}

\paragraph{Pooled localization differences and the tuning metric.}
\Cref{tab:s_noninf} pools the paired per-question differences between \method and ViCrop's label-tuned layer over the models, with one bootstrap cluster per question. In pointing the intervals contain zero on COCO and RefCOCO and lie below zero on V$^*$; in AUROC they lie above zero on all three. ViCrop's layer is tuned for pointing in the main paper; \cref{tab:s_counts} also tunes it for AUROC, which lowers its pointing on COCO (0.690 against 0.731, \cref{tab:main}) and raises its AUROC there only to 0.822, and repeats the pointing tests with exact McNemar tests.

\begin{table}[!tbp]
\centering
\footnotesize
\setlength{\tabcolsep}{2.6pt}
\begin{adjustbox}{max width=\linewidth}\begin{tabular}{@{}lccc@{}}
\toprule
Dataset & Questions & $\Delta$ pointing [95\% CI] & $\Delta$ AUROC [95\% CI] \\
\midrule
COCO & 393 & $-0.008$ $[-0.023,{+}0.007]$ & ${+}0.029$ $[{+}0.026,{+}0.031]$ \\
RefCOCO & 285 & ${+}0.017$ $[-0.001,{+}0.035]$ & ${+}0.009$ $[{+}0.006,{+}0.012]$ \\
V$^*$ & 191 & $-0.032$ $[-0.054,-0.012]$ & ${+}0.035$ $[{+}0.028,{+}0.042]$ \\
\bottomrule
\end{tabular}
\end{adjustbox}
\caption{\textbf{\method minus ViCrop's label-tuned layer, pooled over models} (six VLMs; V$^*$: five), with 95\% bootstrap intervals over questions (20{,}000 draws).}
\label{tab:s_noninf}
\end{table}
\begin{table}[!tbp]
\centering
\footnotesize
\setlength{\tabcolsep}{2.6pt}
\begin{adjustbox}{max width=\linewidth}\begin{tabular}{@{}lcc@{}}
\toprule
\method against & Pointing: better / worse & AUROC: better / worse \\
\midrule
Layer tuned for pointing$^\dagger$ & 3 (0) / 2 (2) & 13 (12) / 1 (0) \\
Layer tuned for AUROC$^\dagger$ & 4 (3) / 2 (2) & 10 (9) / 2 (2) \\
Pointing-tuned layer, McNemar exact & 2 (0) / 2 (2) & -- \\
\bottomrule
\end{tabular}
\end{adjustbox}
\caption{\textbf{Per-pair tests against label-tuned layers} over the 17 model--dataset pairs: in how many \method is significantly better / worse ($p<0.05$; in parentheses after Holm correction over the 17). Wilcoxon unless marked. $^\dagger$Tuned on the labeled calibration split (V$^*$: on COCO).}
\label{tab:s_counts}
\end{table}

\paragraph{Detect-and-mask baselines and ViCrop with our prior.}
\Cref{tab:s_locextras} masks the sinks (top 5\% of generic-prompt attention) before taking the peak of mean attention, of the KL-weighted map and of Chefer relevance, masks instead the top 5\% of tokens by hidden-state norm at 60\% depth for the first two, and replaces ViCrop's labeled layer by our prior. Masking sinks raises mean attention's COCO pointing by 0.13--0.50 but leaves it below \method in every model; masking by norm lowers it in four of the six models and raises it only on Qwen3-VL-8B and LLaVA-1.5-7B, so the norm is a poor detector of these sinks. On V$^*$, the KL-weighted map with sinks masked points better than \method on SmolVLM-500M and LLaVA-1.5-7B, and Chefer relevance with sinks masked on InternVL3-1B (0.131 against 0.042). With our prior, ViCrop's ratio is never significantly different from \method in pointing after Holm correction over the 17 pairs and is below it in AUROC in 12.

\begin{table*}[!tbp]
\centering
\footnotesize
\setlength{\tabcolsep}{2.4pt}
\begin{adjustbox}{max width=\textwidth}\begin{tabular}{@{}lcccccccc@{}}
\toprule
Model & \method & ViCrop~\cite{zhang2025vicrop}, AUROC-tuned$^\dagger$ & ViCrop, our prior & Attn., sinks masked & Attn., norm masked & KL-w., sinks masked & KL-w., norm masked & Chefer~\cite{chefer2021generic}, sinks masked \\
\midrule
\multicolumn{9}{@{}l}{\emph{COCO}} \\
SmolVLM-256M & 0.537 / 0.762 & 0.486$^*$ / 0.725$^*$ & 0.511 / 0.734$^*$ & 0.440$^*$ / 0.668$^*$ & 0.272$^*$ / 0.626$^*$ & 0.557 / 0.688$^*$ & 0.366$^*$ / 0.648$^*$ & 0.468$^*$ / 0.687$^*$ \\
SmolVLM-500M & 0.718 / 0.854 & 0.626$^*$ / 0.834$^*$ & 0.692 / 0.832$^*$ & 0.677 / 0.768$^*$ & 0.295$^*$ / 0.699$^*$ & 0.664$^*$ / 0.779$^*$ & 0.517$^*$ / 0.720$^*$ & 0.662$^*$ / 0.781$^*$ \\
InternVL3-1B & 0.463 / 0.755 & 0.570$^*$ / 0.783$^*$ & 0.458 / 0.754 & 0.356$^*$ / 0.602$^*$ & 0.140$^*$ / 0.544$^*$ & 0.392$^*$ / 0.626$^*$ & 0.201$^*$ / 0.572$^*$ & 0.402$^*$ / 0.646$^*$ \\
Qwen3-VL-2B & 0.819 / 0.898 & 0.827 / 0.890$^*$ & 0.852 / 0.891$^*$ & 0.420$^*$ / 0.696$^*$ & 0.282$^*$ / 0.643$^*$ & 0.468$^*$ / 0.730$^*$ & 0.402$^*$ / 0.690$^*$ & 0.397$^*$ / 0.685$^*$ \\
Qwen3-VL-8B & 0.891 / 0.913 & 0.885 / 0.897$^*$ & 0.903 / 0.898$^*$ & 0.483$^*$ / 0.712$^*$ & 0.356$^*$ / 0.699$^*$ & 0.529$^*$ / 0.776$^*$ & 0.506$^*$ / 0.770$^*$ & 0.369$^*$ / 0.667$^*$ \\
LLaVA-1.5-7B & 0.908 / 0.877 & 0.746$^*$ / 0.800$^*$ & 0.906 / 0.799$^*$ & 0.606$^*$ / 0.752$^*$ & 0.425$^*$ / 0.745$^*$ & 0.679$^*$ / 0.795$^*$ & 0.550$^*$ / 0.792$^*$ & 0.774$^*$ / 0.766$^*$ \\
\bottomrule
\end{tabular}\end{adjustbox}
\caption{\textbf{Detect-and-mask baselines, an AUROC-tuned layer and ViCrop with our prior}: pointing / token AUROC on held-out data. Attn.: mean attention; KL-w.: KL-weighted map. Sinks masked: the 5\% of tokens with the highest generic-prompt attention are excluded from the peak and ranked last; norm masked: the 5\% with the highest hidden-state norm at 60\% depth. $^*$Significantly different from \method (paired two-sided Wilcoxon, $p<0.05$, uncorrected). $^\dagger$Tuned with labels (V$^*$: on COCO).}
\label{tab:s_locextras}
\end{table*}

\paragraph{Log-ratios at sink tokens.}
For each head and prompt, we measure the sinks' log-routing advantage as the mean log-routing of sink tokens minus the median over visual tokens. We also compute the mean sink log-ratio $d$ minus the median of $d$. This centered score measures how the question changes sink preference relative to other tokens and need not equal the difference between the two log-routing advantages. We average across heads with the \method weights and report medians across questions (\cref{tab:s_sinklogit}).

Selecting sinks on the generic-prompt pass can bias their measured change downward. Gray-image sinks provide a second definition using neither real-image pass. With both definitions the median centered log-ratio at sinks is negative or at most 0.3 bits, while the centered log-ratio at target tokens is positive. The sink sets overlap by 40--70\% for four models, 31\% for InternVL3-1B, and 10\% for LLaVA-1.5-7B.

\begin{table*}[!tbp]
\centering
\footnotesize
\setlength{\tabcolsep}{3pt}
\begin{adjustbox}{max width=\textwidth}\begin{tabular}{@{}lcccccccccc@{}}
\toprule
 & \multicolumn{3}{c}{Sinks under the generic prompt} & \multicolumn{3}{c}{Sinks on the gray image} & Target & \multicolumn{2}{c}{Misses on gray sinks (\%)} & Overlap \\
\cmidrule(lr){2-4}\cmidrule(lr){5-7}\cmidrule(lr){8-8}\cmidrule(lr){9-10}
Model & $g$ & $t$ & change & $g$ & $t$ & change & change & Attn. & \method & (\%) \\
\midrule
\multicolumn{11}{@{}l}{\emph{COCO}} \\
SmolVLM-256M & 5.3 & 5.2 & $-0.12$ & 4.6 & 3.9 & $-0.58$ & $+2.67$ & 94 & 15 & 67 \\
SmolVLM-500M & 7.3 & 0.7 & $-6.26$ & 6.3 & 0.2 & $-5.78$ & $+5.26$ & 93 & 1 & 67 \\
InternVL3-1B & 6.1 & 5.5 & $-0.43$ & 1.6 & 1.5 & $+0.15$ & $+1.37$ & 96 & 6 & 31 \\
Qwen3-VL-2B & 4.5 & 2.4 & $-2.05$ & 3.4 & 1.6 & $-1.55$ & $+3.64$ & 89 & 0 & 70 \\
Qwen3-VL-8B & 4.9 & 2.5 & $-1.90$ & 2.8 & 1.1 & $-1.26$ & $+3.96$ & 97 & 0 & 40 \\
LLaVA-1.5-7B & 5.3 & 5.0 & $-0.28$ & 0.5 & 0.3 & $-0.27$ & $+3.73$ & 7 & 8 & 10 \\
\multicolumn{11}{@{}l}{\emph{V$^*$}} \\
SmolVLM-500M & 5.8 & 4.3 & $-1.32$ & 4.8 & 3.7 & $-0.97$ & $+0.98$ & 100 & 1 & 67 \\
InternVL3-1B & 6.0 & 5.4 & $-0.73$ & 2.2 & 2.2 & $-0.25$ & $+2.40$ & 100 & 3 & 31 \\
Qwen3-VL-2B & 4.8 & 2.5 & $-1.99$ & 2.9 & 1.7 & $-0.99$ & $+5.00$ & 91 & 3 & 60 \\
Qwen3-VL-8B & 5.0 & 2.3 & $-2.31$ & 2.8 & 1.1 & $-1.40$ & $+4.17$ & 92 & 0 & 40 \\
LLaVA-1.5-7B & 5.1 & 3.9 & $-1.07$ & 0.5 & 0.8 & $+0.29$ & $+3.38$ & 4 & 7 & 10 \\
\bottomrule
\end{tabular}\end{adjustbox}
\caption{\textbf{How the question reads the sinks} (bits, map-weighted heads, medians over held-out questions). $g$, $t$: advantage of the sink tokens' routing over the median visual token under the generic prompt and under the question; change: the sinks' log-ratio $d$ minus the median $d$ (close to $t-g$). Target: the same change for the target tokens. Misses on gray sinks: share of misses whose peak lies on a gray-image sink, for mean attention and \method. Overlap: share of gray-image sinks that are also generic-prompt sinks.}
\label{tab:s_sinklogit}
\end{table*}

\paragraph{Target size.}
\Cref{tab:s_strata} splits the held-out questions of every model into thirds by the share of visual tokens that the target covers. On COCO, \method' deficit against the label-tuned layer comes from the largest third; on RefCOCO no third is worse, and the smallest is better; on V$^*$ the deficit is largest for the smallest third. The last column gives, for the median question, the share of generic-prompt attention on the target divided by the share of question attention on it (head means, prior-weighted over layers): the generic prompt already puts about half as much attention on the largest objects as the question does, against a tenth on the smallest, which leaves less contrast at large objects; RefCOCO's largest targets show a similar ratio without a deficit, so this does not explain all of it.

\begin{table*}[!tbp]
\centering
\footnotesize
\setlength{\tabcolsep}{3pt}
\begin{adjustbox}{max width=\textwidth}\begin{tabular}{@{}llccccccc@{}}
\toprule
Dataset & Size third & Median size (\%) & Pairs & \method & ViCrop~\cite{zhang2025vicrop}, tuned$^\dagger$ & Attn. & $\Delta$ [95\% CI] & Generic / question \\
\midrule
COCO & small & 5.6 & 727 & 0.598 & 0.590 & 0.158 & $-0.001$ $[-0.034,{+}0.032]$ & 0.09 \\
 & medium & 14.6 & 809 & 0.726 & 0.708 & 0.182 & ${+}0.015$ $[-0.016,{+}0.046]$ & 0.21 \\
 & large & 34.4 & 822 & 0.830 & 0.877 & 0.316 & $-0.052$ $[-0.080,-0.026]$ & 0.49 \\
\addlinespace[2pt]
RefCOCO & small & 10.2 & 566 & 0.541 & 0.511 & 0.120 & ${+}0.037$ $[{+}0.003,{+}0.074]$ & 0.24 \\
 & medium & 20.4 & 542 & 0.714 & 0.697 & 0.155 & $-0.001$ $[-0.041,{+}0.035]$ & 0.37 \\
 & large & 36.1 & 602 & 0.762 & 0.757 & 0.271 & ${+}0.000$ $[-0.036,{+}0.033]$ & 0.55 \\
\addlinespace[2pt]
V$^*$ & small & 0.5 & 199 & 0.166 & 0.231 & 0.000 & $-0.067$ $[-0.113,-0.023]$ & -- \\
 & medium & 1.0 & 328 & 0.192 & 0.213 & 0.003 & $-0.021$ $[-0.053,{+}0.011]$ & -- \\
 & large & 2.0 & 428 & 0.245 & 0.271 & 0.002 & $-0.020$ $[-0.048,{+}0.009]$ & -- \\
\bottomrule
\end{tabular}\end{adjustbox}
\caption{\textbf{Localization by target size.} Thirds of each model's held-out questions by target size (share of visual tokens, median given), pooled over models (``pairs'': model--question pairs; on V$^*$ the thirds are uneven because many targets cover the same number of tokens). Pointing of \method, the label-tuned layer and mean attention; $\Delta$: \method minus the tuned layer, 95\% bootstrap interval over questions. Generic / question: median ratio of the target's share of generic-prompt and of question attention.}
\label{tab:s_strata}
\end{table*}

\paragraph{Comparison with a layer oracle.}
\Cref{tab:s_regret} compares the prior with an oracle that picks the best single layer of \method' own per-layer map $F_l$ on the report split itself. The regret is at most 0.031 on COCO and RefCOCO except for InternVL3-1B on COCO (0.115), and negative in five of the twelve pairs, where mixing layers beats every single layer. InternVL3-1B's map localizes in two depth ranges, layers 13--15 and 20--23 of 24; the bits favor the deeper one, which is the best on RefCOCO (0.71--0.72 at layers 20--22) but reaches only 0.45--0.49 on COCO, against 0.58 at layer 14; at the prior's top layer 27\% of its COCO peaks fall on the image border, against 15\% at layer 14 and 4\% on RefCOCO.

\begin{table*}[!tbp]
\centering
\footnotesize
\setlength{\tabcolsep}{3pt}
\begin{adjustbox}{max width=\textwidth}\begin{tabular}{@{}lccccccccccc@{}}
\toprule
 & & & \multicolumn{3}{c}{COCO} & \multicolumn{3}{c}{RefCOCO} & \multicolumn{3}{c}{V$^*$} \\
\cmidrule(lr){4-6}\cmidrule(lr){7-9}\cmidrule(l){10-12}
Model & Layers & Prior top 3 & Prior & Best layer & Regret & Prior & Best layer & Regret & Prior & Best layer & Regret \\
\midrule
SmolVLM-256M & 30 & 23, 21, 22 & 0.537 & 17 (0.542) & $+0.005$ & 0.460 & 21 (0.481) & $+0.021$ & -- & -- & -- \\
SmolVLM-500M & 32 & 20, 23, 27 & 0.718 & 20 (0.712) & $-0.005$ & 0.533 & 20 (0.509) & $-0.025$ & 0.042 & 29 (0.110) & $+0.068$ \\
InternVL3-1B & 24 & 20, 23, 22 & 0.463 & 14 (0.578) & $+0.115$ & 0.726 & 22 (0.719) & $-0.007$ & 0.042 & 15 (0.084) & $+0.042$ \\
Qwen3-VL-2B & 28 & 17, 20, 22 & 0.819 & 16 (0.802) & $-0.018$ & 0.786 & 17 (0.800) & $+0.014$ & 0.387 & 17 (0.377) & $-0.010$ \\
Qwen3-VL-8B & 36 & 21, 23, 24 & 0.891 & 23 (0.921) & $+0.031$ & 0.888 & 23 (0.902) & $+0.014$ & 0.372 & 24 (0.461) & $+0.089$ \\
LLaVA-1.5-7B & 32 & 14, 20, 19 & 0.908 & 14 (0.893) & $-0.015$ & 0.649 & 14 (0.667) & $+0.018$ & 0.209 & 14 (0.225) & $+0.016$ \\
\bottomrule
\end{tabular}\end{adjustbox}
\caption{\textbf{Regret of the prior}: held-out pointing of \method, the best single layer of its own map on the same held-out questions (an oracle; layer index and pointing), and the difference. Layers are counted from 0.}
\label{tab:s_regret}
\end{table*}

\paragraph{Whether-to-look bits.}
\Cref{tab:s_whether} replaces the where-to-look bits, in both the head weights and the prior, by the whether-to-look bits $[\mathrm{KL}_{\mathrm{bin}}(m^{xt}\Vert m^{xg})-\mathrm{KL}_{\mathrm{bin}}(m^{\varnothing t}\Vert m^{\varnothing g})]_+$ or by the total coarse-grained divergence of \cref{prop:chain}, corrected in the same way. Whether-to-look bits alone are worse in 17 of the 24 comparisons and never better; the total divergence is close to where-to-look bits alone except on LLaVA-1.5-7B, where it also changes one of the prior's top three layers and costs 0.056 in COCO pointing.

\begin{table*}[!tbp]
\centering
\footnotesize
\setlength{\tabcolsep}{2pt}
\begin{adjustbox}{max width=\textwidth}\begin{tabular}{@{}lcccccc@{}}
\toprule
 & \multicolumn{3}{c}{COCO} & \multicolumn{3}{c}{RefCOCO} \\
\cmidrule(lr){2-4}\cmidrule(l){5-7}
Model & where (ours) & whether & whether $+$ where & where (ours) & whether & whether $+$ where \\
\midrule
SmolVLM-256M & 0.537 / 0.762 & 0.532 / 0.749$^*$ & 0.532 / 0.755$^*$ & 0.460 / 0.638 & 0.470 / 0.637 & 0.481 / 0.638$^*$ \\
SmolVLM-500M & 0.718 / 0.854 & 0.644$^*$ / 0.843$^*$ & 0.723 / 0.857$^*$ & 0.533 / 0.680 & 0.512 / 0.660$^*$ & 0.530 / 0.676$^*$ \\
InternVL3-1B & 0.463 / 0.755 & 0.356$^*$ / 0.707$^*$ & 0.463 / 0.754 & 0.726 / 0.793 & 0.614$^*$ / 0.743$^*$ & 0.730 / 0.793 \\
Qwen3-VL-2B & 0.819 / 0.898 & 0.735$^*$ / 0.870$^*$ & 0.819 / 0.897$^*$ & 0.786 / 0.812 & 0.775 / 0.803$^*$ & 0.789 / 0.810$^*$ \\
Qwen3-VL-8B & 0.891 / 0.913 & 0.903 / 0.904$^*$ & 0.893 / 0.913 & 0.888 / 0.852 & 0.870 / 0.844$^*$ & 0.884 / 0.854$^*$ \\
LLaVA-1.5-7B & 0.908 / 0.877 & 0.517$^*$ / 0.703$^*$ & 0.852$^*$ / 0.841$^*$ & 0.649 / 0.754 & 0.516$^*$ / 0.686$^*$ & 0.646 / 0.740$^*$ \\
\bottomrule
\end{tabular}\end{adjustbox}
\caption{\textbf{Which bits}: pointing / AUROC on held-out data with head weights and prior from where-to-look bits (\method), whether-to-look bits, or both. $^*$Significantly different from \method ($p<0.05$, uncorrected).}
\label{tab:s_whether}
\end{table*}

\paragraph{Controls for both faithfulness axes.}
\Cref{tab:s_faith_ranks} adds the flat-first control, the reversed codec-bit order, to the seven maps of \cref{fig:faith}. Its curves are those of the codec-bit order read at complementary fractions, so all eight maps are scored on the stored fractions that reversal maps onto themselves, $\{0,0.3,0.5,0.7,1\}$, instead of all eight fractions; ranks therefore differ from \cref{fig:faith} by up to 0.3. Each axis favors one end of the texture order: the codec-bit map ranks 5.1 on the token axis and last on the bit axis, the flat-first map last on the token axis and 5.9 on the bit axis, above random. \method is significantly better than the flat-first control in all ten settings of each axis (18 of 20 after Holm correction). \Cref{tab:s_codec_corr} gives the per-question Spearman correlation of each map with the codec-bit map: \method' maps are nearly uncorrelated with texture ($-0.01$ to $0.18$), whereas Chefer relevance and mean attention lean toward flat regions (negative correlations in five and four models), which the bit axis would reward, yet they rank below \method on it.

\begin{table}[!tbp]
\centering
\footnotesize
\begin{adjustbox}{max width=\linewidth}\begin{tabular}{@{}lcc@{}}
\toprule
Map & Token axis & Bit axis \\
\midrule
\method & 1.1 & 1.1 \\
ViCrop~\cite{zhang2025vicrop}, tuned$^\dagger$ & 2.6 & 2.5 \\
ViCrop, all layers & 3.1 & 3.6 \\
CARVE~\cite{ge2025carve} & 4.2 & 3.5 \\
Chefer relevance~\cite{chefer2021generic} & 4.9 & 4.4 \\
Codec bits (texture first) & 5.1 & 8.0 \\
Reversed codec bits (flat first) & 8.0 & 5.9 \\
Random & 7.0 & 7.0 \\
\bottomrule
\end{tabular}\end{adjustbox}
\caption{\textbf{Mean rank of eight maps} by insertion$-$deletion AUC over the ten COCO settings of each axis, with both texture controls, on the fractions $\{0,0.3,0.5,0.7,1\}$. $^\dagger$Layer tuned with labels.}
\label{tab:s_faith_ranks}
\end{table}
\begin{table*}[!tbp]
\centering
\footnotesize
\setlength{\tabcolsep}{2pt}
\begin{adjustbox}{max width=\textwidth}\begin{tabular}{@{}lccccccc@{}}
\toprule
Model & \method & ViCrop~\cite{zhang2025vicrop}, tuned$^\dagger$ & ViCrop, all layers & CARVE~\cite{ge2025carve} & Chefer relevance~\cite{chefer2021generic} & Attention (mean) & KL-weighted map \\
\midrule
SmolVLM-256M & $+0.09$ & $+0.09$ & $+0.18$ & $+0.04$ & $-0.05$ & $-0.02$ & $+0.04$ \\
SmolVLM-500M & $+0.14$ & $+0.16$ & $+0.18$ & $+0.06$ & $-0.04$ & $-0.05$ & $+0.04$ \\
InternVL3-1B & $+0.10$ & $-0.00$ & $+0.17$ & $-0.19$ & $-0.21$ & $-0.20$ & $-0.10$ \\
Qwen3-VL-2B & $+0.15$ & $+0.13$ & $+0.18$ & $-0.08$ & $-0.16$ & $-0.16$ & $-0.10$ \\
Qwen3-VL-8B & $+0.18$ & $+0.23$ & $+0.25$ & $-0.10$ & $-0.16$ & -- & -- \\
LLaVA-1.5-7B & $-0.01$ & $-0.06$ & $-0.06$ & $+0.00$ & $+0.03$ & -- & -- \\
\bottomrule
\end{tabular}\end{adjustbox}
\caption{\textbf{Correlation of each map with the codec-bit map} (Spearman over the visual tokens, mean over held-out COCO questions). Positive: the map favors textured tokens. --: map not in the faithfulness study of that model. $^\dagger$Layer tuned with labels.}
\label{tab:s_codec_corr}
\end{table*}

\section{Transfer, sensitivity and robustness}
\label{sec:s_transfer}
Transfer uses the fixed estimator, weighting rule, and hyperparameters. The sensitivity studies vary one input choice at a time.

\paragraph{Settings.}
Janus-Pro-1B and -7B~\cite{chen2025januspro} (Hugging Face checkpoints \texttt{deepseek-community/Janus-Pro-1B/7B}) encode a $384\times384$ image with SigLIP-L into a $24\times24$ grid of 576 visual tokens; their language models have 24 layers of 16 heads and 30 layers of 32 heads, and the chat template places the image before the text, as in every other model.
For Qwen3-VL at 896\,px, each image is resized to one $896\times896$ tile ($28\times28=784$ tokens of 32\,px). At native resolution the aspect ratio is kept and the image is resized to a multiple of 32\,px with at most 1280 tokens and no upscaling: COCO images get 265 tokens on average, V$^*$ images 1248--1280. The processor's pixel limits are set per call so that it does not rescale the frame, and every dump checks the token grid.
At each resolution the prior is re-estimated from the unlabeled COCO calibration questions and ViCrop's layer is re-tuned with their labels (and used for V$^*$); as a check, the 448-px prior is also applied unchanged. As in the main study, ViCrop's tuned layer for RefCOCO is tuned on the RefCOCO expressions with odd image ids (layers 13 and 23 for Janus-Pro-1B and -7B) and for COCO and V$^*$ on COCO (layers 17 and 16).

\paragraph{Localization across families and resolutions.}
\Cref{tab:s_transfer_coco,tab:s_transfer_vstar,tab:s_transfer_refcoco} give every map in every new setting and \cref{tab:s_transfer_prior} the prior and the sinks.
Over the 14 new model--dataset pairs, \method is better than the label-tuned layer in AUROC in 11 (11 after Holm correction) and worse in 1 (1), Janus-Pro-1B on RefCOCO; in pointing it is better in 3 (1) and worse in 2 (0), both for Qwen3-VL-8B at 896\,px. Against ViCrop over all layers it is better in AUROC in all 14 and in pointing in 4 (2), and worse in pointing in 2 (1): Qwen3-VL-2B at 896\,px on COCO and Qwen3-VL-8B at native resolution on V$^*$. Against CARVE it is better in pointing in 13 (12) and in AUROC in 14, and against Chefer relevance, mean attention and the KL-weighted map in all 14 on both metrics.
In Janus-Pro, mean attention peaks at one border token (row 18 or 19, column 0 or 23) for 55--65\% of the COCO questions and hits the V$^*$ target in 4 (1B) and 1 (7B) of 191 questions; \method puts at most 2.1\% of its peaks on any one position.
For Qwen3-VL, the top three layers of the prior are the same at all three resolutions (layers 17, 20, 22 of 28 for 2B; 21, 23, 24 of 36 for 8B), the bits per layer correlate with those at 448\,px with Spearman $\rho=0.98$--$0.99$, and the 448-px prior gives held-out COCO pointing within 0.023 and AUROC within 0.002 of the prior estimated at the same resolution. The layer that ViCrop tunes on COCO at 896\,px for Qwen3-VL-2B (layer 18) points at the COCO object for 0.761 of the questions but at the V$^*$ target for only 0.225.
At higher resolution the sink of mean attention weakens or moves (for Qwen3-VL-8B at 896\,px the top-left token holds 12\% of the COCO peaks instead of 89\%, but 88\% again at native resolution; for Qwen3-VL-2B the most frequent position moves to other border tokens), yet mean attention never peaks on a V$^*$ target, and with sinks defined as in \cref{sec:sinks} (the 5\% of tokens with the highest generic-prompt attention), 94--100\% of its misses land on a sink at both higher resolutions and 98--100\% in Janus-Pro, against 0--3\% for \method.

\begin{table*}[!tbp]
\centering
\footnotesize
\setlength{\tabcolsep}{3pt}
\begin{adjustbox}{max width=\textwidth}\begin{tabular}{@{}lcccccccccccc@{}}
\toprule
 & \multicolumn{2}{c}{Janus-Pro-1B, 384\,px} & \multicolumn{2}{c}{Janus-Pro-7B, 384\,px} & \multicolumn{2}{c}{Qwen3-VL-2B, 896\,px} & \multicolumn{2}{c}{Qwen3-VL-2B, native} & \multicolumn{2}{c}{Qwen3-VL-8B, 896\,px} & \multicolumn{2}{c}{Qwen3-VL-8B, native} \\
\cmidrule(lr){2-3}\cmidrule(lr){4-5}\cmidrule(lr){6-7}\cmidrule(lr){8-9}\cmidrule(lr){10-11}\cmidrule(lr){12-13}
Method & Pt. & AUROC & Pt. & AUROC & Pt. & AUROC & Pt. & AUROC & Pt. & AUROC & Pt. & AUROC \\
\midrule
\method (ours) & 0.835 & 0.849 & 0.840 & 0.855 & 0.740 & 0.889 & 0.786 & 0.891 & 0.880 & 0.908 & 0.896 & 0.912 \\
ViCrop~\cite{zhang2025vicrop}, tuned layer$^\dagger$ & 0.776$^*$ & 0.800$^*$ & 0.819 & 0.835$^*$ & 0.761 & 0.806$^*$ & 0.789 & 0.885$^*$ & 0.929$^*$ & 0.892$^*$ & 0.896 & 0.898$^*$ \\
ViCrop, layer band & 0.786$^*$ & 0.839$^*$ & 0.796$^*$ & 0.829$^*$ & 0.835$^*$ & 0.873$^*$ & 0.794 & 0.870$^*$ & 0.898 & 0.882$^*$ & 0.883 & 0.887$^*$ \\
ViCrop, all layers & 0.799 & 0.838$^*$ & 0.796$^*$ & 0.820$^*$ & 0.819$^*$ & 0.868$^*$ & 0.784 & 0.863$^*$ & 0.903 & 0.879$^*$ & 0.868 & 0.883$^*$ \\
CARVE~\cite{ge2025carve} (re-impl.) & 0.654$^*$ & 0.775$^*$ & 0.392$^*$ & 0.699$^*$ & 0.448$^*$ & 0.690$^*$ & 0.481$^*$ & 0.722$^*$ & 0.344$^*$ & 0.683$^*$ & 0.593$^*$ & 0.742$^*$ \\
Chefer relevance~\cite{chefer2021generic} & 0.252$^*$ & 0.732$^*$ & 0.338$^*$ & 0.683$^*$ & 0.249$^*$ & 0.623$^*$ & 0.318$^*$ & 0.643$^*$ & 0.117$^*$ & 0.618$^*$ & 0.135$^*$ & 0.643$^*$ \\
\bottomrule
\end{tabular}\end{adjustbox}
\caption{\textbf{Transfer: held-out COCO} (393 questions with even image ids) for the new family and the higher resolutions. Pointing / token AUROC; $^*$: paired two-sided Wilcoxon $p<0.05$ against \method (uncorrected). $^\dagger$Layer tuned with labels on the COCO calibration split at the same resolution.}
\label{tab:s_transfer_coco}
\end{table*}

\begin{table*}[!tbp]
\centering
\footnotesize
\setlength{\tabcolsep}{3pt}
\begin{adjustbox}{max width=\textwidth}\begin{tabular}{@{}lcccccccccccc@{}}
\toprule
 & \multicolumn{2}{c}{Janus-Pro-1B, 384\,px} & \multicolumn{2}{c}{Janus-Pro-7B, 384\,px} & \multicolumn{2}{c}{Qwen3-VL-2B, 896\,px} & \multicolumn{2}{c}{Qwen3-VL-2B, native} & \multicolumn{2}{c}{Qwen3-VL-8B, 896\,px} & \multicolumn{2}{c}{Qwen3-VL-8B, native} \\
\cmidrule(lr){2-3}\cmidrule(lr){4-5}\cmidrule(lr){6-7}\cmidrule(lr){8-9}\cmidrule(lr){10-11}\cmidrule(lr){12-13}
Method & Pt. & AUROC & Pt. & AUROC & Pt. & AUROC & Pt. & AUROC & Pt. & AUROC & Pt. & AUROC \\
\midrule
\method (ours) & 0.277 & 0.837 & 0.257 & 0.819 & 0.508 & 0.948 & 0.524 & 0.962 & 0.398 & 0.931 & 0.450 & 0.950 \\
ViCrop~\cite{zhang2025vicrop}, tuned layer$^\dagger$ & 0.215$^*$ & 0.704$^*$ & 0.272 & 0.806$^*$ & 0.225$^*$ & 0.875$^*$ & 0.492 & 0.958$^*$ & 0.455$^*$ & 0.925 & 0.482 & 0.953 \\
ViCrop, layer band & 0.309 & 0.781$^*$ & 0.283 & 0.788$^*$ & 0.361$^*$ & 0.930$^*$ & 0.372$^*$ & 0.945$^*$ & 0.429 & 0.927$^*$ & 0.503$^*$ & 0.947$^*$ \\
ViCrop, all layers & 0.267 & 0.749$^*$ & 0.267 & 0.774$^*$ & 0.361$^*$ & 0.931$^*$ & 0.372$^*$ & 0.945$^*$ & 0.429 & 0.930$^*$ & 0.508$^*$ & 0.947$^*$ \\
CARVE~\cite{ge2025carve} (re-impl.) & 0.220$^*$ & 0.758$^*$ & 0.079$^*$ & 0.754$^*$ & 0.356$^*$ & 0.887$^*$ & 0.398$^*$ & 0.909$^*$ & 0.037$^*$ & 0.802$^*$ & 0.010$^*$ & 0.853$^*$ \\
Chefer relevance~\cite{chefer2021generic} & 0.068$^*$ & 0.745$^*$ & 0.021$^*$ & 0.766$^*$ & 0.073$^*$ & 0.871$^*$ & 0.079$^*$ & 0.904$^*$ & 0.000$^*$ & 0.853$^*$ & 0.005$^*$ & 0.895$^*$ \\
\bottomrule
\end{tabular}\end{adjustbox}
\caption{\textbf{Transfer: V$^*$ Bench} (191 questions). Same conventions as \cref{tab:s_transfer_coco}; the tuned layer is the one tuned on COCO. Chance pointing (mean share of target tokens): 0.007 for Janus-Pro, 0.006 at 896\,px and 0.005 at native resolution.}
\label{tab:s_transfer_vstar}
\end{table*}

\begin{table}[!tbp]
\centering
\footnotesize
\setlength{\tabcolsep}{3pt}
\begin{adjustbox}{max width=\linewidth}\begin{tabular}{@{}lcccc@{}}
\toprule
 & \multicolumn{2}{c}{Janus-Pro-1B} & \multicolumn{2}{c}{Janus-Pro-7B} \\
\cmidrule(lr){2-3}\cmidrule(lr){4-5}
Method & Pt. & AUROC & Pt. & AUROC \\
\midrule
\method (ours) & 0.653 & 0.725 & 0.667 & 0.731 \\
ViCrop~\cite{zhang2025vicrop}, tuned layer$^\dagger$ & 0.695 & 0.730$^*$ & 0.625 & 0.688$^*$ \\
ViCrop, layer band & 0.635 & 0.716$^*$ & 0.600$^*$ & 0.710$^*$ \\
ViCrop, all layers & 0.625 & 0.701$^*$ & 0.604$^*$ & 0.705$^*$ \\
CARVE~\cite{ge2025carve} (re-impl.) & 0.628 & 0.636$^*$ & 0.488$^*$ & 0.619$^*$ \\
Chefer relevance~\cite{chefer2021generic} & 0.116$^*$ & 0.575$^*$ & 0.151$^*$ & 0.585$^*$ \\
Attention (mean) & 0.147$^*$ & 0.593$^*$ & 0.119$^*$ & 0.587$^*$ \\
KL-weighted map & 0.516$^*$ & 0.631$^*$ & 0.221$^*$ & 0.624$^*$ \\
Random & 0.211$^*$ & 0.499$^*$ & 0.211$^*$ & 0.499$^*$ \\
\bottomrule
\end{tabular}\end{adjustbox}
\caption{\textbf{Transfer: RefCOCO} (285 held-out expressions) for Janus-Pro, with the prior from COCO. Same conventions as \cref{tab:s_transfer_coco}; the tuned layer is tuned on the 315 RefCOCO expressions with odd image ids.}
\label{tab:s_transfer_refcoco}
\end{table}

\begin{table*}[!tbp]
\centering
\footnotesize
\setlength{\tabcolsep}{4pt}
\begin{adjustbox}{max width=\textwidth}\begin{tabular}{@{}llccccc@{}}
\toprule
Model & Input & Top-3 layers & $\rho$ vs 448\,px & 448\,px prior: Pt. / AUROC & Own prior: Pt. / AUROC & Attn. peak at one token (COCO / V$^*$) \\
\midrule
Qwen3-VL-2B & 448\,px & 17, 20, 22 & -- & -- & 0.819 / 0.898 & 45\% / 46\% \\
 & 896\,px & 17, 20, 22 & 0.98 & 0.763 / 0.890 & 0.740 / 0.889 & 38\% / 87\% \\
 & native & 17, 20, 22 & 0.98 & 0.779 / 0.892 & 0.786 / 0.891 & 16\% / 63\% \\
Qwen3-VL-8B & 448\,px & 21, 23, 24 & -- & -- & 0.891 / 0.913 & 89\% / 92\% \\
 & 896\,px & 21, 23, 24 & 0.99 & 0.885 / 0.909 & 0.880 / 0.908 & 12\% / 9\% \\
 & native & 21, 23, 24 & 0.99 & 0.898 / 0.912 & 0.896 / 0.912 & 88\% / 3\% \\
Janus-Pro-1B & 384\,px & 13, 16, 22 & -- & -- & 0.835 / 0.849 & 65\% / 90\% \\
Janus-Pro-7B & 384\,px & 16, 17, 20 & -- & -- & 0.840 / 0.855 & 55\% / 32\% \\
\bottomrule
\end{tabular}\end{adjustbox}
\caption{\textbf{The prior across resolutions and families.} Top-3 layers of the prior; Spearman $\rho$ of the bits per layer with those at 448\,px; held-out COCO pointing / AUROC with the 448-px prior applied to the higher-resolution maps and with the prior estimated at the same resolution; share of questions whose mean-attention peak lies at its most frequent position (COCO / V$^*$).}
\label{tab:s_transfer_prior}
\end{table*}

\begin{table*}[!t]
\centering
\footnotesize
\setlength{\tabcolsep}{4pt}
\begin{adjustbox}{max width=\textwidth}\begin{tabular}{@{}lcccccccc@{}}
\toprule
 & \multicolumn{4}{c}{Qwen3-VL-2B} & \multicolumn{4}{c}{Qwen3-VL-8B} \\
\cmidrule(lr){2-5}\cmidrule(l){6-9}
 & \multicolumn{2}{c}{896\,px} & \multicolumn{2}{c}{native} & \multicolumn{2}{c}{896\,px} & \multicolumn{2}{c}{native} \\
\cmidrule(lr){2-3}\cmidrule(lr){4-5}\cmidrule(lr){6-7}\cmidrule(l){8-9}
Window & 1/2 & 1/3 & 1/2 & 1/3 & 1/2 & 1/3 & 1/2 & 1/3 \\
\midrule
No zoom & 63.9 & 63.9 & 69.1 & 69.1$^*$ & 65.4$^*$ & 65.4$^*$ & 72.3$^*$ & 72.3$^*$ \\
Center crop & 58.6 & 57.6$^*$ & 66.5$^*$ & 64.4$^*$ & 66.0$^*$ & 64.4$^*$ & 74.3$^*$ & 73.3$^*$ \\
Attention (mean) & 63.9 & 57.1$^*$ & 70.7 & 64.4$^*$ & 63.9$^*$ & 62.8$^*$ & 72.3$^*$ & 69.1$^*$ \\
Chefer relevance~\cite{chefer2021generic} & 61.8 & 62.8$^*$ & 69.6 & 68.6$^*$ & 64.4$^*$ & 62.8$^*$ & 73.3$^*$ & 69.1$^*$ \\
CARVE~\cite{ge2025carve} & 66.0 & 70.2 & 69.1$^*$ & 72.8$^*$ & 68.6 & 67.0$^*$ & 74.9$^*$ & 72.8$^*$ \\
ViCrop~\cite{zhang2025vicrop}, all layers & 66.0 & 72.8 & 73.3 & 79.1 & 73.8 & 78.5 & 82.2 & 82.7 \\
ViCrop, re-tuned layer$^\dagger$ & 62.8 & 67.0 & 74.3 & 78.5 & 73.3 & 79.6 & 82.7 & 84.3 \\
\method, 448-px prior & 64.9 & 72.3 & 74.3 & 77.5 & 72.3 & 77.0 & 84.3 & 84.3 \\
\textbf{\method} & 65.4 & 70.7 & 74.3 & 79.6 & 72.3 & 77.5 & 84.8 & 84.8 \\
\method, bits window & 69.6 & 71.7 & 75.4 & 79.6 & 69.1 & 71.7$^*$ & 77.5$^*$ & 76.4$^*$ \\
\midrule
Attention: corner crops (\%) & 99 & 99 & 100 & 99 & 97 & 62 & 91 & 67 \\
Attention: target in crop (\%) & 39 & 7 & 43 & 13 & 8 & 1 & 8 & 2 \\
\method: target in crop (\%) & 74 & 60 & 77 & 66 & 72 & 57 & 77 & 58 \\
\bottomrule
\end{tabular}\end{adjustbox}
\caption{\textbf{V$^*$ zoom-in at higher resolution under forced choice} (191 questions; accuracy in \%). $^*$: $p<0.05$ against \method (exact McNemar, uncorrected). Bottom: share of mean-attention crops that start at an image corner, and share of crops containing every target box. $^\dagger$Tuned with COCO labels at the same resolution.}
\label{tab:s_zoom_highres}
\end{table*}
\begin{table*}[!t]
\centering
\footnotesize
\setlength{\tabcolsep}{2.4pt}
\begin{adjustbox}{max width=\textwidth}\begin{tabular}{@{}lcccccc@{}}
\toprule
 & \multicolumn{3}{c}{Primary rule} & \multicolumn{3}{c}{Bits window} \\
\cmidrule(lr){2-4}\cmidrule(l){5-7}
\method $-$ & diff. & 95\% CI & $p$ & diff. & 95\% CI & $p$ \\
\midrule
ViCrop~\cite{zhang2025vicrop}, re-tuned layer$^\dagger$ & ${+}0.9$ & $[-0.9,{+}2.6]$ & 0.375 & $-1.4$ & $[-3.4,{+}0.5]$ & 0.164 \\
ViCrop, all layers & ${+}0.1$ & $[-1.6,{+}2.0]$ & 0.944 & $-2.2$ & $[-4.2,-0.1]$ & 0.044 \\
CARVE~\cite{ge2025carve} & ${+}6.0$ & $[{+}3.5,{+}8.6]$ & $<$0.001 & ${+}3.7$ & $[{+}1.8,{+}5.7]$ & $<$0.001 \\
Chefer relevance~\cite{chefer2021generic} & ${+}9.6$ & $[{+}6.3,{+}12.8]$ & $<$0.001 & ${+}7.3$ & $[{+}4.6,{+}9.9]$ & $<$0.001 \\
Attention (mean) & ${+}10.7$ & $[{+}7.3,{+}14.2]$ & $<$0.001 & ${+}8.4$ & $[{+}5.5,{+}11.4]$ & $<$0.001 \\
Center crop & ${+}10.5$ & $[{+}6.7,{+}14.5]$ & $<$0.001 & ${+}8.2$ & $[{+}5.0,{+}11.5]$ & $<$0.001 \\
No zoom & ${+}8.5$ & $[{+}4.7,{+}12.4]$ & $<$0.001 & ${+}6.2$ & $[{+}2.7,{+}9.8]$ & $<$0.001 \\
\method, 448-px prior & ${+}0.3$ & $[-0.2,{+}0.9]$ & 0.337 & $-2.0$ & $[-3.9,-0.1]$ & 0.047 \\
\method, primary rule & \multicolumn{3}{c}{--} & $-2.3$ & $[-4.1,-0.5]$ & 0.020 \\
\bottomrule
\end{tabular}\end{adjustbox}
\caption{\textbf{Pooled high-resolution zoom-in under forced choice} (eight settings, question clusters; conventions as in \cref{tab:s_vstar_fc_pooled}).}
\label{tab:s_zoom_highres_pooled}
\end{table*}
\begin{table*}[!t]
\centering
\footnotesize
\setlength{\tabcolsep}{2.4pt}
\begin{adjustbox}{max width=\textwidth}\begin{tabular}{@{}lcccccc@{}}
\toprule
 & \multicolumn{3}{c}{as scored} & \multicolumn{3}{c}{no letter at chance} \\
\cmidrule(lr){2-4}\cmidrule(l){5-7}
\method $-$ & diff. & 95\% CI & $p$ & diff. & 95\% CI & $p$ \\
\midrule
ViCrop~\cite{zhang2025vicrop}, re-tuned layer$^\dagger$ & ${+}0.9$ & $[-1.0,{+}2.8]$ & 0.378 & ${+}0.7$ & $[-1.0,{+}2.4]$ & 0.411 \\
ViCrop, all layers & $-0.5$ & $[-2.4,{+}1.5]$ & 0.644 & $-0.4$ & $[-2.1,{+}1.3]$ & 0.660 \\
CARVE~\cite{ge2025carve} & ${+}5.9$ & $[{+}3.2,{+}8.6]$ & $<$0.001 & ${+}5.1$ & $[{+}2.7,{+}7.5]$ & $<$0.001 \\
Chefer relevance~\cite{chefer2021generic} & ${+}11.5$ & $[{+}8.3,{+}14.8]$ & $<$0.001 & ${+}9.0$ & $[{+}6.3,{+}11.9]$ & $<$0.001 \\
Attention (mean) & ${+}12.9$ & $[{+}9.5,{+}16.4]$ & $<$0.001 & ${+}10.6$ & $[{+}7.5,{+}13.8]$ & $<$0.001 \\
Center crop & ${+}12.7$ & $[{+}8.8,{+}16.7]$ & $<$0.001 & ${+}10.5$ & $[{+}6.9,{+}14.1]$ & $<$0.001 \\
No zoom & ${+}2.7$ & $[-1.4,{+}6.8]$ & 0.203 & ${+}6.6$ & $[{+}2.8,{+}10.4]$ & $<$0.001 \\
\method, 448-px prior & ${+}0.5$ & $[{+}0.1,{+}1.0]$ & 0.038 & ${+}0.4$ & $[{+}0.0,{+}0.9]$ & 0.047 \\
\method, bits window & ${+}3.4$ & $[{+}1.4,{+}5.5]$ & 0.002 & ${+}3.0$ & $[{+}1.0,{+}4.9]$ & 0.003 \\
\bottomrule
\end{tabular}\end{adjustbox}
\caption{\textbf{Pooled high-resolution zoom-in with generated answers} (eight settings, question clusters; conventions as in \cref{tab:s_zoom_vstar_q}).}
\label{tab:s_zoom_highres_gen}
\end{table*}

\paragraph{Zoom-in at 896\,px and native resolution.}
We repeat the V$^*$ zoom-in study on Qwen3-VL-2B and -8B at two higher resolutions, recalibrating the prior and retuning ViCrop's layer at each resolution (\cref{tab:s_zoom_highres}). Under forced choice, no-zoom accuracy rises from 48--49\% at 448\,px to 64--72\%. Mean attention sends 62--100\% of its crops to corners and does not help on average ($-2.2$ points, 95\% CI $[-5.0,0.7]$). \method' crops contain the target in 57--77\% of questions.

Pooled over the eight model--resolution--window settings (\cref{tab:s_zoom_highres_pooled}), the primary rule gains 8.5 points over no zoom ($[4.7,12.4]$) and improves over CARVE, Chefer relevance, mean attention, and a center crop; its intervals against the retuned layer and all-layer ViCrop include zero. Reusing the 448-px prior instead of recalibrating costs 0.3 points ($[-0.2,0.9]$). The bits-window rule trails the primary rule by 2.3 points ($p=0.020$), with the whole gap on Qwen3-VL-8B. With generated answers (\cref{tab:s_zoom_highres_gen}), the gain over no zoom is 2.7 points as scored and 6.6 with chance credit.

\paragraph{Sensitivity to the generic prompt and the gray value.}
We compare three additional generic prompts and gray levels of 64 and 192 on held-out COCO (\cref{tab:s_sensitivity}). Each variant recalibrates the \method prior without labels and retunes ViCrop's layer with labels. Changing the gray value changes \method pointing by at most 0.0102 and AUROC by at most 0.0005. The ``Describe this image'' and ``What do you see in this image?'' variants change \method pointing by $-0.031$ to $+0.059$. The caption prompt reduces pointing by 0.059 on InternVL3-1B and 0.191 on Qwen3-VL-2B.

To quantify how much the reference already attends to the target, we divide its mean attention on the object by the object's share of visual tokens. A ratio of 1 corresponds to uniform routing. The caption prompt raises this ratio from 0.52 to 0.71 on InternVL3-1B and from 0.48 to 0.72 on Qwen3-VL-2B. Across ten prompt variants, the change in target attention correlates with the change in \method pointing ($\rho=-0.86$, $p=0.002$) and AUROC ($\rho=-0.82$, $p=0.004$). Centering the pointing changes within model gives weaker evidence ($\rho=-0.56$, $p=0.12$ over nine variants across three models).

\paragraph{A label-free check of the generic prompt.}
For each reference, we average $\sum_{l,h}w_{lh}$ over the unlabeled COCO calibration questions (\cref{tab:s_promptbits}). The caption prompt yields the fewest corrected bits in all three models tested, 3--17\% below the default. Across ten prompt variants, the change in bits correlates with the change in pointing ($\rho=0.76$, $p=0.011$). In three models, ``Describe this image.'' yields more bits than the default, with pointing changes from $-0.031$ to $+0.020$. This check was added after the protocol was fixed.

\paragraph{Codec robustness of the bit axis.}
We recompute per-cell bit estimates for \texttt{mbt2018-mean} at qualities 2, 4, and 6 and \texttt{bmshj2018-hyperprior}~\cite{balle2018hyperprior} at quality 4. The stored perturbation curves keep their original deletion and insertion orders, including the texture control ranked by quality-4 bits. Recomputing the quality-4 axis reproduces 96\% of stored curves exactly, with a mean absolute AUC difference of $1.9\times10^{-5}$ over 49{,}616 curves. Tied scores account for the discrepancies. Mean ranks change by at most 0.1, and the texture control ranks last under all four codec settings (\cref{tab:s_codec}).

\paragraph{Faithfulness of maps that multiply by attention.}
On the 393 held-out COCO questions of the four smaller models we ran deletion and insertion for mean attention, the KL-weighted map and the attention under the generic prompt, which ignores the question. The probability curves of the codec-bit control, run again alongside these maps, match those of the main study (maximum difference $3.7\times10^{-6}$ in deletion AUC, same first answer token for every question), so the new curves are comparable with the stored ones. \Cref{tab:s_faith_foils} ranks the ten maps over the 16 settings (four models, gray and blur, two axes): \method is significantly better than all three new maps in all 16 settings, also after Holm correction over the 16, and generic-prompt attention ties with random.

\paragraph{Faithfulness on TextVQA.}
On a 200-question subset of the TextVQA zoom-in sample for SmolVLM-500M, InternVL3-1B, and Qwen3-VL-2B, \cref{tab:s_faith_textvqa} ranks the maps over 12 settings. Here maps that keep attention mass lead: CARVE is significantly better than \method in 10 of 12 settings (8 after Holm correction), mean attention in 5 (4) and Chefer relevance in 6 (5), while \method is significantly better than the label-tuned layer in 5 settings (4) and never worse. Two measurements distinguish the TextVQA setting from COCO. On TextVQA the question changes routing more (prior-weighted bits 19--22 against 5--15 in the three models) and its routing overlaps the generic prompt's less (0.21--0.29 against 0.28--0.55, histogram intersection at the prior's layers). These measurements alone do not explain why attention-weighted maps perform better in this experiment.

\begin{table}[H]
\centering
\footnotesize
\setlength{\tabcolsep}{3pt}
\begin{adjustbox}{max width=\linewidth}\begin{tabular}{@{}llccccc@{}}
\toprule
 & & \multicolumn{2}{c}{\method} & \multicolumn{2}{c}{ViCrop~\cite{zhang2025vicrop}, tuned$^\dagger$} & \\
\cmidrule(lr){3-4}\cmidrule(lr){5-6}
Model & Variant & Pt. & AUROC & Pt. & AUROC & $g$ on obj. \\
\midrule
SmolVLM-500M & default & 0.718 & 0.854 & 0.687 & 0.827 & 0.71 \\
 & describe & $-0.031$$^*$ & $-0.011$$^*$ & $-0.015$ & $-0.012$$^*$ & 0.73 \\
 & see & $-0.031$ & $-0.028$$^*$ & $-0.005$ & $-0.029$$^*$ & 0.91 \\
 & caption & $-0.018$ & $-0.029$$^*$ & $-0.008$ & $-0.028$$^*$ & 0.96 \\
 & gray 64 & $0.0000$ & ${+}0.0002$$^*$ & $0.0000$ & $0.0000$ & 0.71 \\
 & gray 192 & $0.0000$ & ${+}0.0001$$^*$ & $0.0000$ & $0.0000$ & 0.71 \\
\midrule
InternVL3-1B & default & 0.463 & 0.755 & 0.588 & 0.738 & 0.52 \\
 & describe & ${+}0.020$ & ${+}0.021$$^*$ & ${+}0.041$ & ${+}0.048$$^*$ & 0.45 \\
 & see & ${+}0.059$$^*$ & $-0.007$$^*$ & ${+}0.005$ & ${+}0.033$$^*$ & 0.49 \\
 & caption & $-0.059$$^*$ & $-0.105$$^*$ & ${+}0.038$ & ${+}0.035$$^*$ & 0.71 \\
 & gray 64 & ${+}0.0025$ & $-0.0004$$^*$ & $0.0000$ & $0.0000$ & 0.52 \\
 & gray 192 & $-0.0025$ & $-0.0003$$^*$ & $0.0000$ & $0.0000$ & 0.52 \\
\midrule
Qwen3-VL-2B & default & 0.819 & 0.898 & 0.827 & 0.890 & 0.48 \\
 & describe & ${+}0.008$ & $-0.004$$^*$ & $-0.087$$^*$ & $-0.055$$^*$ & 0.47 \\
 & see & $-0.010$ & $-0.004$$^*$ & $-0.023$ & $-0.045$$^*$ & 0.46 \\
 & caption & $-0.191$$^*$ & $-0.083$$^*$ & $-0.069$$^*$ & $-0.040$$^*$ & 0.72 \\
 & gray 64 & $-0.0025$ & $-0.0004$ & $0.0000$ & $0.0000$ & 0.48 \\
 & gray 192 & ${+}0.0102$ & $-0.0003$$^*$ & $0.0000$ & $0.0000$ & 0.48 \\
\midrule
Qwen3-VL-8B & default & 0.891 & 0.913 & 0.885 & 0.897 & 0.52 \\
 & describe & $-0.003$ & $0.000$ & ${+}0.020$ & $-0.010$$^*$ & 0.53 \\
 & gray 64 & $0.0000$ & $-0.0003$$^*$ & $0.0000$ & $0.0000$ & 0.52 \\
\bottomrule
\end{tabular}\end{adjustbox}
\caption{\textbf{Sensitivity to the generic prompt and the gray value} (held-out COCO). Default: ``Write a general description of the image.'' and gray 124, with pointing and AUROC; other rows give the change against the default, paired over the same questions ($^*$: Wilcoxon $p<0.05$). Prompts: describe, ``Describe this image.''; see, ``What do you see in this image?''; caption, ``Provide a short caption for the image.''. \method re-estimates its prior without labels and ViCrop re-tunes its layer with labels for each variant. $g$ on obj.: generic-prompt attention on the queried object relative to its area (1 = uniform).}
\label{tab:s_sensitivity}
\end{table}

\begin{table}[H]
\centering
\footnotesize
\setlength{\tabcolsep}{2pt}
\begin{adjustbox}{max width=\linewidth}\begin{tabular}{@{}llcccc@{}}
\toprule
Model & Reference & Bits & Change (\%) & $g$ on obj. & $\Delta$ Pt. \\
\midrule
SmolVLM-500M & default & 53.8 & $+0$ & 0.71 & $+0.000$ \\
 & describe & 56.4 & $+5$ & 0.73 & $-0.031$ \\
 & see & 47.5 & $-12$ & 0.91 & $-0.031$ \\
 & caption & 44.7 & $-17$ & 0.96 & $-0.018$ \\
 & gray 64 & 55.0 & $+2$ & 0.71 & $+0.000$ \\
 & gray 192 & 54.5 & $+1$ & 0.71 & $+0.000$ \\
\addlinespace[2pt]
InternVL3-1B & default & 46.7 & $+0$ & 0.52 & $+0.000$ \\
 & describe & 53.0 & $+13$ & 0.45 & $+0.020$ \\
 & see & 51.9 & $+11$ & 0.49 & $+0.059$ \\
 & caption & 45.3 & $-3$ & 0.71 & $-0.059$ \\
 & gray 64 & 47.6 & $+2$ & 0.52 & $+0.003$ \\
 & gray 192 & 47.2 & $+1$ & 0.52 & $-0.003$ \\
\addlinespace[2pt]
Qwen3-VL-2B & default & 39.0 & $+0$ & 0.48 & $+0.000$ \\
 & describe & 43.2 & $+11$ & 0.47 & $+0.008$ \\
 & see & 41.4 & $+6$ & 0.46 & $-0.010$ \\
 & caption & 35.3 & $-10$ & 0.72 & $-0.191$ \\
 & gray 64 & 38.0 & $-3$ & 0.48 & $-0.003$ \\
 & gray 192 & 43.4 & $+11$ & 0.48 & $+0.010$ \\
\addlinespace[2pt]
Qwen3-VL-8B & default & 86.2 & $+0$ & 0.52 & $+0.000$ \\
 & describe & 85.6 & $-1$ & 0.53 & $-0.003$ \\
 & gray 64 & 86.5 & $+0$ & 0.52 & $+0.000$ \\
\bottomrule
\end{tabular}\end{adjustbox}
\caption{\textbf{Bits as a label-free check of the reference.} Bits are the mean of $\sum_{l,h}w_{lh}$ over unlabeled COCO calibration questions. Prompts, object attention, and changes in pointing use the conventions of \cref{tab:s_sensitivity}.}
\label{tab:s_promptbits}
\end{table}

\begin{table}[H]
\centering
\footnotesize
\setlength{\tabcolsep}{2.6pt}
\begin{adjustbox}{max width=\linewidth}\begin{tabular}{@{}lcccc@{}}
\toprule
Map & mbt~\cite{minnen2018joint}, $q2$ & mbt, $q4^\S$ & mbt, $q6$ & bmshj~\cite{balle2018hyperprior}, $q4$ \\
Bits per pixel & 0.21 & 0.47 & 0.95 & 0.49 \\
$\rho$ with $q4$ bits & 0.95 & -- & 0.97 & 0.99 \\
\midrule
\method (ours) & 1.1 & 1.1 & 1.0 & 1.0 \\
ViCrop~\cite{zhang2025vicrop}, tuned layer$^\dagger$ & 2.7 & 2.7 & 2.8 & 2.8 \\
ViCrop, all layers & 3.6 & 3.6 & 3.5 & 3.5 \\
CARVE~\cite{ge2025carve} & 3.5 & 3.4 & 3.4 & 3.4 \\
Chefer relevance~\cite{chefer2021generic} & 4.1 & 4.2 & 4.3 & 4.3 \\
Random & 6.0 & 6.0 & 6.0 & 6.0 \\
Codec bits (texture control) & 7.0 & 7.0 & 7.0 & 7.0 \\
\bottomrule
\end{tabular}\end{adjustbox}
\caption{\textbf{The bit axis under four codecs.} Mean rank by insertion$-$deletion AUC on the bit axis over the 10 settings of \cref{fig:faith}. mbt: \texttt{mbt2018-mean}; bmshj: \texttt{bmshj2018-hyperprior}; $q$: CompressAI quality; $^\S$the paper's codec. First rows: mean bits per pixel of the held-out images and Spearman correlation of the token bits with those of the paper's codec.}
\label{tab:s_codec}
\end{table}

\begin{table}[H]
\centering
\footnotesize
\setlength{\tabcolsep}{2.6pt}
\begin{adjustbox}{max width=\linewidth}\input{tab_supp/faith_foils}\end{adjustbox}
\caption{\textbf{COCO faithfulness with maps that multiply by attention} (four smaller models, 16 settings). Mean rank of ten maps by insertion$-$deletion AUC on each axis, and in how many settings \method is significantly better or worse (Wilcoxon $p<0.05$; in parentheses after Holm correction over the 16). $^\ddagger$Maps that multiply by attention, run for this comparison.}
\label{tab:s_faith_foils}
\end{table}

\begin{table}[H]
\centering
\footnotesize
\setlength{\tabcolsep}{2.6pt}
\begin{adjustbox}{max width=\linewidth}\input{tab_supp/faith_textvqa}\end{adjustbox}
\caption{\textbf{TextVQA faithfulness} (three models, 200 questions each, 12 settings). Same conventions as \cref{tab:s_faith_foils}.}
\label{tab:s_faith_textvqa}
\end{table}

\end{document}